\pdfoutput=1

\documentclass[11pt]{article}

\usepackage{ACL2023}

\usepackage{times}
\usepackage{latexsym}

\usepackage[T1]{fontenc}

\usepackage[utf8]{inputenc}

\usepackage{microtype}

\usepackage{inconsolata}

\usepackage{booktabs}
\usepackage{microtype}
\usepackage{amsfonts}
\usepackage{amsmath}
\usepackage{amssymb}
\usepackage{amsthm}
\usepackage{booktabs}
\usepackage{dsfont}
\usepackage{paralist}
\usepackage{times}
\usepackage{xspace}
\usepackage{multirow}
\usepackage{graphicx}
\usepackage{float} 
\usepackage{lipsum}
\usepackage{appendix}
\usepackage{enumitem}
\usepackage{pifont}
\usepackage{makecell}
\usepackage{caption}
\usepackage{latexsym}
\usepackage{cleveref}
\usepackage{subcaption}
\usepackage{placeins}
\usepackage{upgreek}

\newcommand{\ra}[1]{\renewcommand{\arraystretch}{#1}}

\definecolor{c_255_197_197}{RGB}{255,197,197}
\definecolor{c_255_74_74}{RGB}{255,74,74}

\title{Neighboring Words Affect Human Interpretation of Saliency Explanations}

  \author{Alon Jacovi$^1$\thanks{\; Both authors contributed equally to this research.} \;\;\;\; Hendrik Schuff$^{2,3*}$ \\ \textbf{Heike Adel$^2$  \;\;\;\; Ngoc Thang Vu$^3$ \;\;\;\; Yoav Goldberg$^{1,4}$}\\
$^1$Bar Ilan University  \quad $^2$Bosch Center for Artificial Intelligence \\
$^3$University of Stuttgart \quad $^4$Allen Institute for Artificial Intelligence  \\
  {\normalsize\tt  alonjacovi@gmail.com}\quad
  {\normalsize\tt  \{hendrik.schuff,heike.adel\}@de.bosch.com}\\
  {\normalsize\tt  thang.vu@ims.uni-stuttgart.de}\quad
  {\normalsize\tt  yoav.goldberg@gmail.com}
  }

\begin{document}
\maketitle
\begin{abstract}

Word-level saliency explanations (``heat maps over words'') are often used to communicate feature-attribution in text-based models. Recent studies found that superficial  factors such as word length can distort human interpretation of the communicated saliency scores. We conduct a user study to investigate how the marking of a word's \textit{neighboring words} affect the explainee's perception of the word's importance in the context of a saliency explanation.
We find that neighboring words have significant effects on the word's importance rating.
Concretely, we identify that the influence changes based on neighboring direction (left vs. right) and 
a-priori linguistic and computational measures of phrases and collocations (vs. unrelated neighboring words).
Our results question whether text-based saliency explanations should be continued to be communicated at word level, and inform future research on alternative saliency explanation methods.
\end{abstract}

\section{Introduction}

\begin{figure}[t]
\centering
\includegraphics[width=0.99\linewidth]{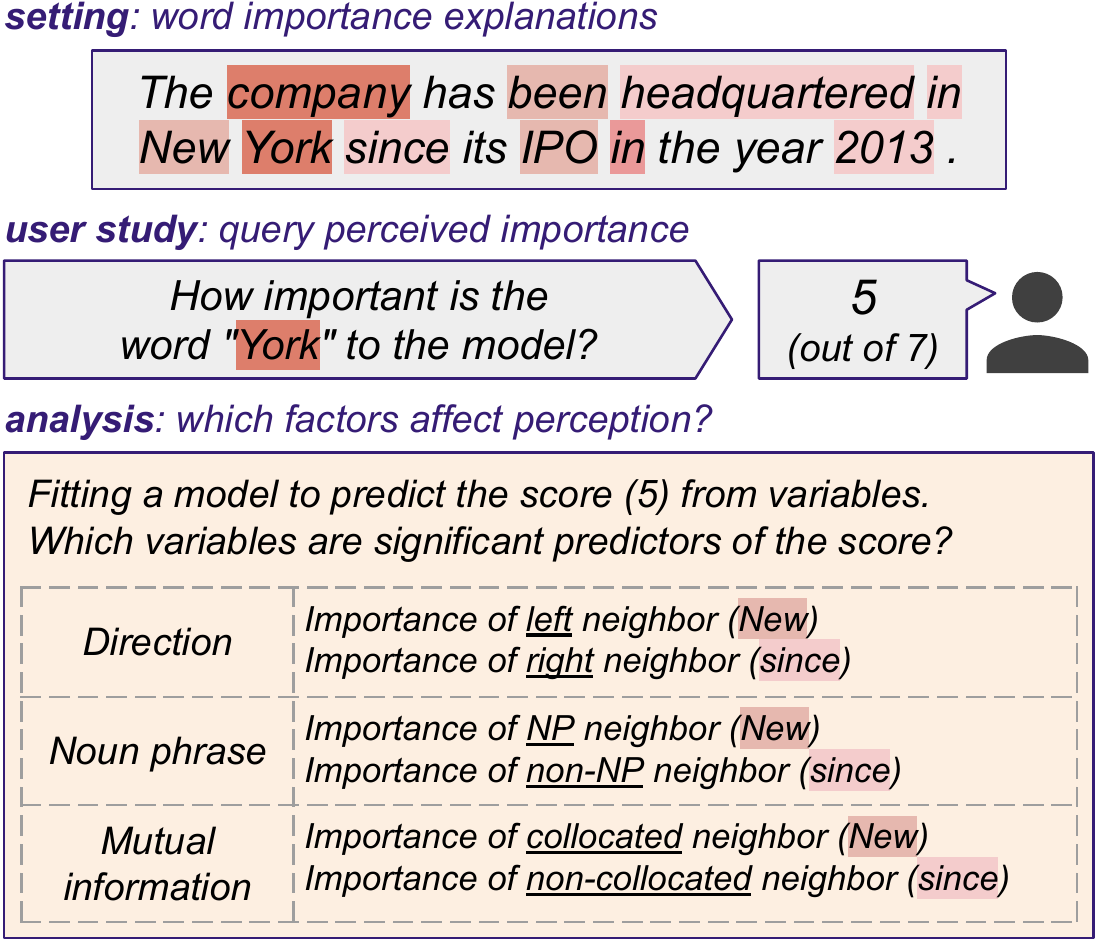}
\caption{Illustration of the user study. We ask laypeople to rate the perceived importance of words following a word-importance explanation (\textit{grey}). Then we analyze the effect of the importance of neighboring words on this interpretation, conditioned on the relationship between the words across various measures (\textit{orange}).
}
\label{fig:intro-example}
\end{figure}
In the context of explainability methods that assign importance scores to individual words, we are interested in characterizing the effect of \textit{phrase-level} features on the perceived importance of a particular word: Text is naturally constructed and comprehended in various levels of granularity that go beyond the word-level \cite{Chomsky+1957,lexicalchunks}.
For example (\Cref{fig:intro-example}), the role of the word ``\textit{York}'' is contextualized by the phrase ``\textit{New York}'' that contains it. 
Given an explanation that attributes importance to ``\textit{New}'' and ``\textit{York}'' separately, what is the effect of the importance score of ``\textit{New}'' on the explainee's understanding of the importance ``\textit{York}''? Our study investigates this question.

Current feature-attribution explanations in NLP mostly operate at word-level or subword-level \cite{DBLP:journals/csur/MadsenRC23,arras__2017,ribeiro16-lime,diogo2019-survey}.
Previous work investigated the effect of word and sentence-level features on subjective interpretations of saliency explanations on text \cite{DBLP:conf/fat/SchuffJAGV22}---finding that features such as word length and
frequency
bias users' perception of explanations (e.g., users may assign higher importance to longer words).

It is not trivial for an explanation of an AI system to successfully communicate the intended information to the explainee \cite{miller19-social,dinu_challenging_2020,DBLP:conf/nips/FelCCCVS21,DBLP:journals/corr/abs-2112-09669}. In the case of \textit{feature-attribution} explanations \cite{DBLP:journals/jair/BurkartH21,PMID:33079674}, which commonly appear in NLP as explanations based on word importance \cite{DBLP:journals/csur/MadsenRC23,DBLP:conf/ijcnlp/DanilevskyQAKKS20}, we must understand how the explainee interprets the role of the attributed inputs on model outputs \cite{DBLP:conf/nips/NguyenKN21,DBLP:conf/aaai/ZhouBRS22}.
Research shows that it is often an error to assume that explainees will interpret explanations ``as intended'' \cite{DBLP:conf/acl/GonzalezRS21,ehsan2021-ai-experts-explanation-perception}.

\begin{figure*}[t]
\centering
\fbox{\includegraphics[width=0.75\linewidth]{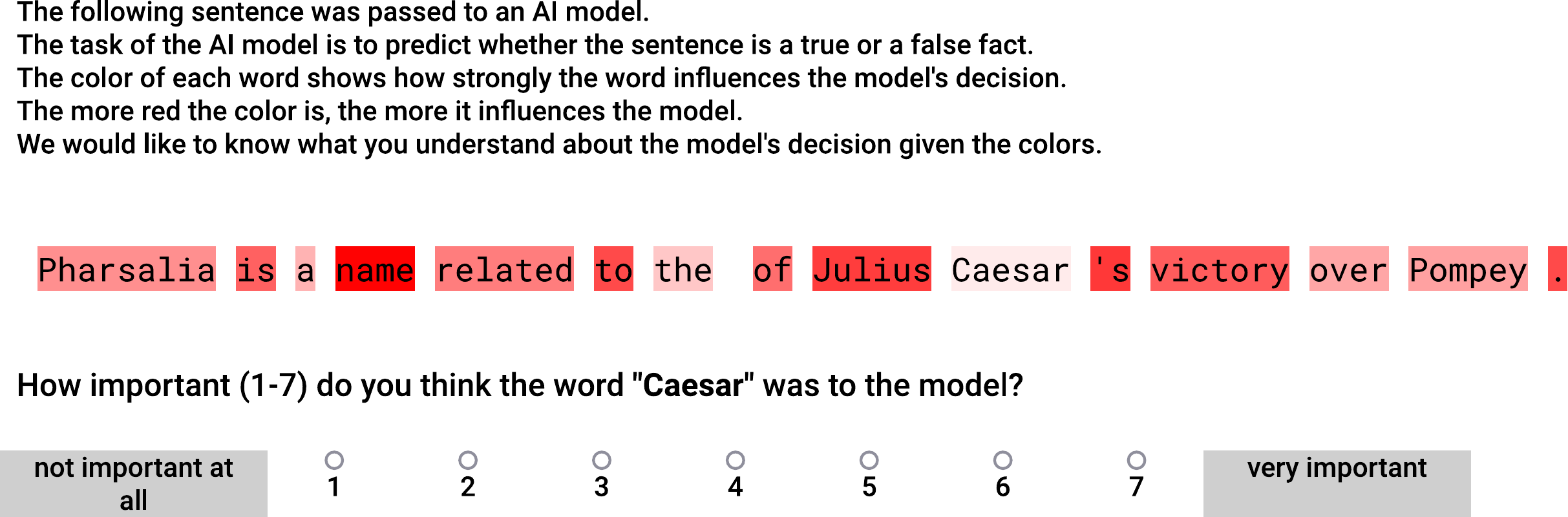}}
\caption{Screenshot of the rating interface.}
\label{fig:gui}
\end{figure*}

The study involves two phases (\Cref{fig:intro-example}).
First, we collect subjective self-reported ratings of importance by laypeople, in a setting of color-coded word importance explanations of a fact-checking NLP model (\Cref{sec:specification}, \Cref{fig:gui}). 
Then, we fit a statistical model to map the importance of \textit{neighboring words} to the word's rating, conditioned on various a-priori measures of bigram constructs, such as the words'
syntactic relation or the degree to which they collocate in a corpus \cite{collocations-survey}. 

We observe significant effects (\Cref{sec:analysis}) for: 1. left-adjacency vs. right-adjacency; 2. the difference in importance between the two words; 3. the phrase relationship between the words (common phrase vs. no relation).
We then deduce likely causes for these effects from relevant literature (\Cref{sec:relation_psychology}).
We are also able to reproduce results by \citet{DBLP:conf/fat/SchuffJAGV22} in a different English language domain (\Cref{sec:reproduction}).
We release the collected data and analysis code.\footnote{\url{https://github.com/boschresearch/human-interpretation-saliency}.}

We conclude that laypeople interpretation of word importance explanations in English \textbf{can be biased via neighboring words' importance}, likely moderated by reading direction and phrase units of language.
Future work on feature-attribution should investigate more effective methods of communicating information \cite{mosca-etal-2022-grammarshap,DBLP:journals/corr/abs-2210-13270}, and implementations of such explanations should take care not to assume that human users interpret word-level importance objectively.

\section{Study Specification} \label{sec:specification}
Our analysis has two phases: Collecting subjective interpretations of word-importances from laypeople, and testing for significant influence in various properties on the collected ratings---in particular, properties of \textit{adjacent words} to the rated word.

\subsection{Collecting Perceived Importance} \label{subsec:step1}

We ask laypeople to rate the importance of a word within a feature-importance explanation (\Cref{fig:gui}).
The setting is based on \citet{DBLP:conf/fat/SchuffJAGV22}, with the main difference in the text domain.
We use the Amazon Mechanical Turk crowd-sourcing platform to recruit a total of 100 participants.\footnote{We select English-speaking raters from English-speaking countries and analyze responses from 64 participants for our first and 36 participants for our second experiment. Details are provided in \Cref{app:user-study-details}.}


\noindent\textbf{Explanations.} We use color-coding visualization of word importance explanations as the more common format in the literature \cite[e.g.,][]{arras__2017,wang_gradient-based_2020,tenney2020language,DBLP:journals/corr/abs-2112-09669}.
We use importance values from two sources: Randomized, and SHAP-values\footnote{As the largest observed SHAP value in our data is 0.405, we normalize all SHAP values with $0.405^{-1}$ to cover the full color range.} \cite{NIPS2017_7062} for \verb+facebook/bart-large-mnli+\footnote{\url{https://huggingface.co/facebook/bart-large-mnli}} \cite{DBLP:conf/emnlp/YinHR19,lewis-etal-2020-bart} as a fact-checking model.
\\[0.5em] 
\noindent\textbf{Task.}  We communicate to the participants that the model is performing a plausible task of deciding whether the given sentence is fact or non-fact \cite{designs5030042}. The source texts are a sample of 150 Wikipedia sentences,\footnote{\label{footnote:wikisent}From the \textit{Wikipedia Sentences} collection, see \url{kaggle.com/datasets/mikeortman/wikipedia-sentences}.}
in order to select text in a domain that has a high natural rate of multi-word chunks. 
\\[0.5em] 
\noindent\textbf{Procedure.} We ask the explainee: ``How important (1-7) do you think the word [...] was to the model?'' and receive a point-scale answer with an optional comment field. This repeats for one randomly-sampled word in each of the 150 sentences.

\subsection{Measuring Neighbor Effects} \label{subsec:step2}

\begin{table}[t]
\ra{1.2}
\centering
\resizebox{0.99\linewidth}{!}{
\begin{tabular}{>{\raggedright}p{1.8cm}>{\raggedright}p{3.3cm}p{3.7cm}}
  \toprule 
\textbf{Measure} & \textbf{Examples} & \textbf{Description} \\ 
  \midrule 
\textit{First-order constituent} & highly developed, more than, such as & Smallest multi-word constituent sub-trees in the constituency tree. \\
\textit{Noun phrase} & tokyo marathon, ski racer, the UK & Multi-word noun phrase in the constituency tree. \\
\midrule
\textit{Frequency} & the United, the family, a species & Raw, unnormalized frequency. \\
\textit{Poisson Stirling} &  an American, such as, a species & Poisson Stirling bigram score. \\
\textit{$\upvarphi^2$} & Massar Egbari, ice hockey, Udo Dirkschneider & Square of the Pearson correlation coefficient. \\ 
   \bottomrule
\end{tabular} }
\caption{Illustrative subset of our phrase measures. 
}
\label{tab:chunk-measures}
\end{table}

\begin{figure*}[h!]
    \centering
    \begin{subfigure}[t]{.24\textwidth}
        \centering
    \includegraphics[width=\textwidth]{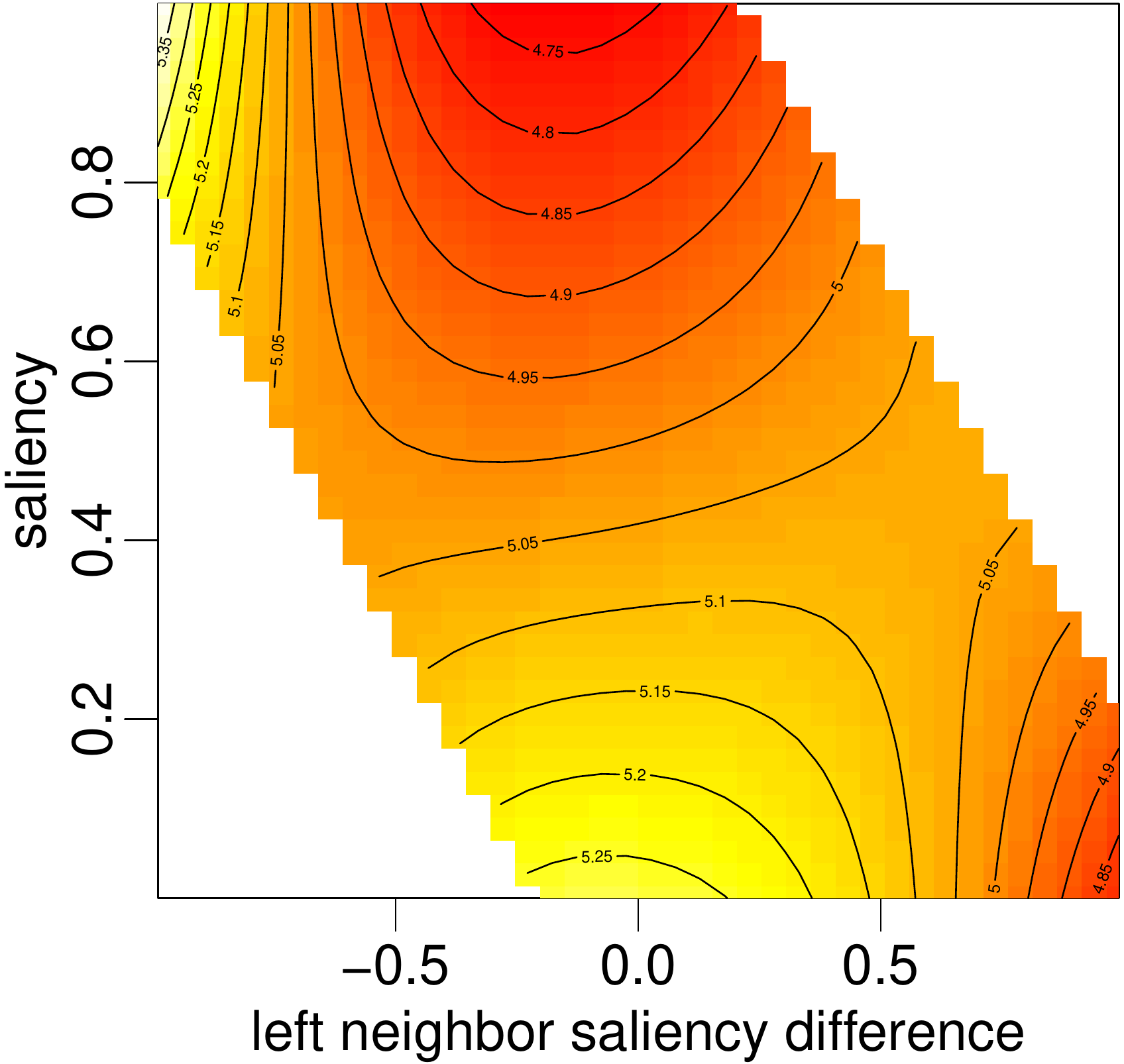}
    \caption{Left, no chunk. ($\ast$)}\label{fig:left_neighbor_saliency_difference_saliency_no_lexical_chunk}
    \end{subfigure}%
    \hfill
    \begin{subfigure}[t]{.24\textwidth}
        \centering
    \includegraphics[width=\textwidth]{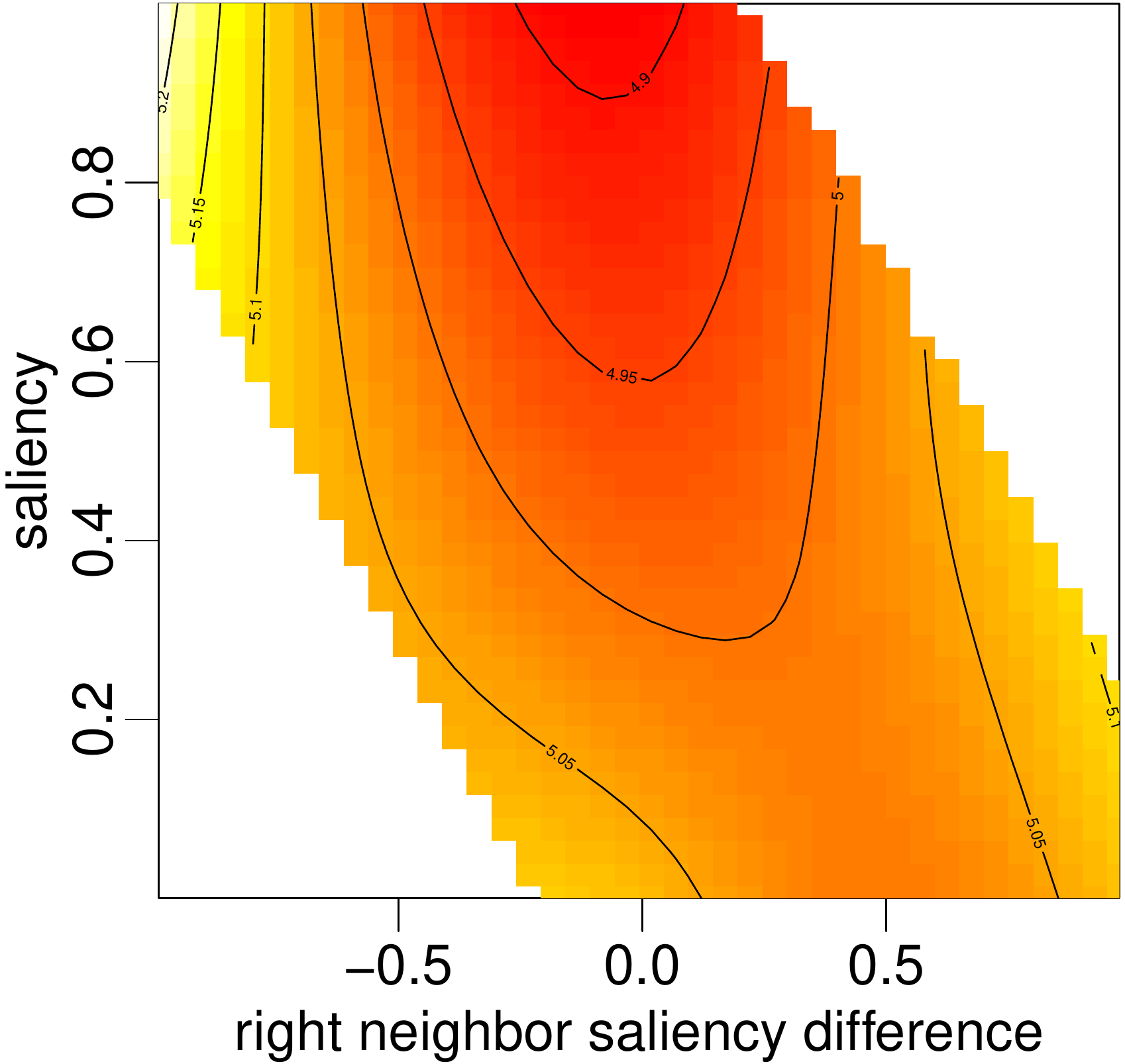}
    \caption{Right, no chunk. ($\ast$)}\label{fig:right_neighbor_saliency_difference_saliency_no_lexical_chunk}
    \end{subfigure}%
        \hfill
    \begin{subfigure}[t]{.24\textwidth}
        \centering
    \includegraphics[width=\textwidth]{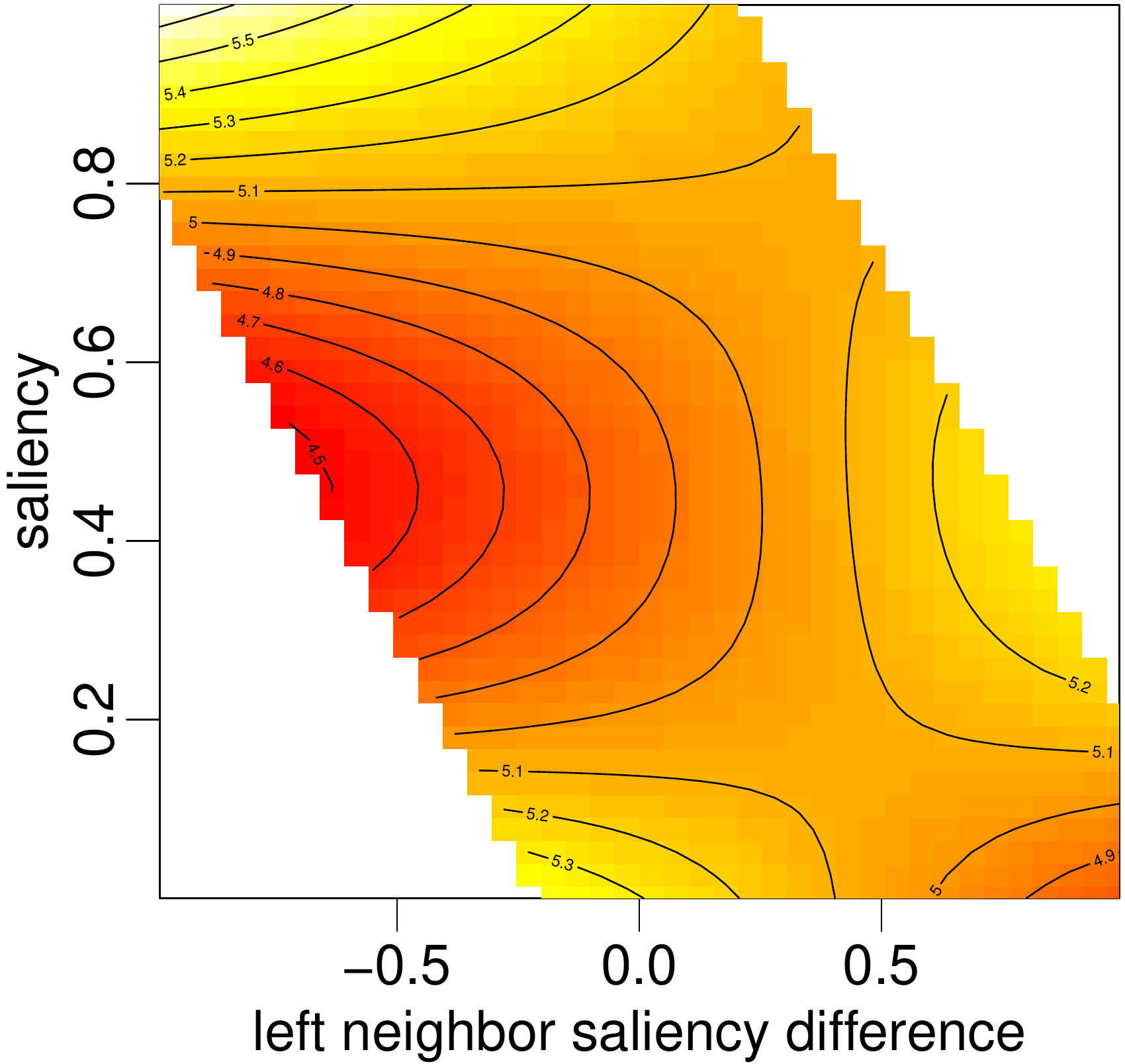}
    \caption{Left, chunk. ($\ast$)}\label{fig:left_neighbor_saliency_difference_saliency_lexical_chunk}
    \end{subfigure}%
    \hfill
    \begin{subfigure}[t]{.24\textwidth}
        \centering
    \includegraphics[width=\textwidth]{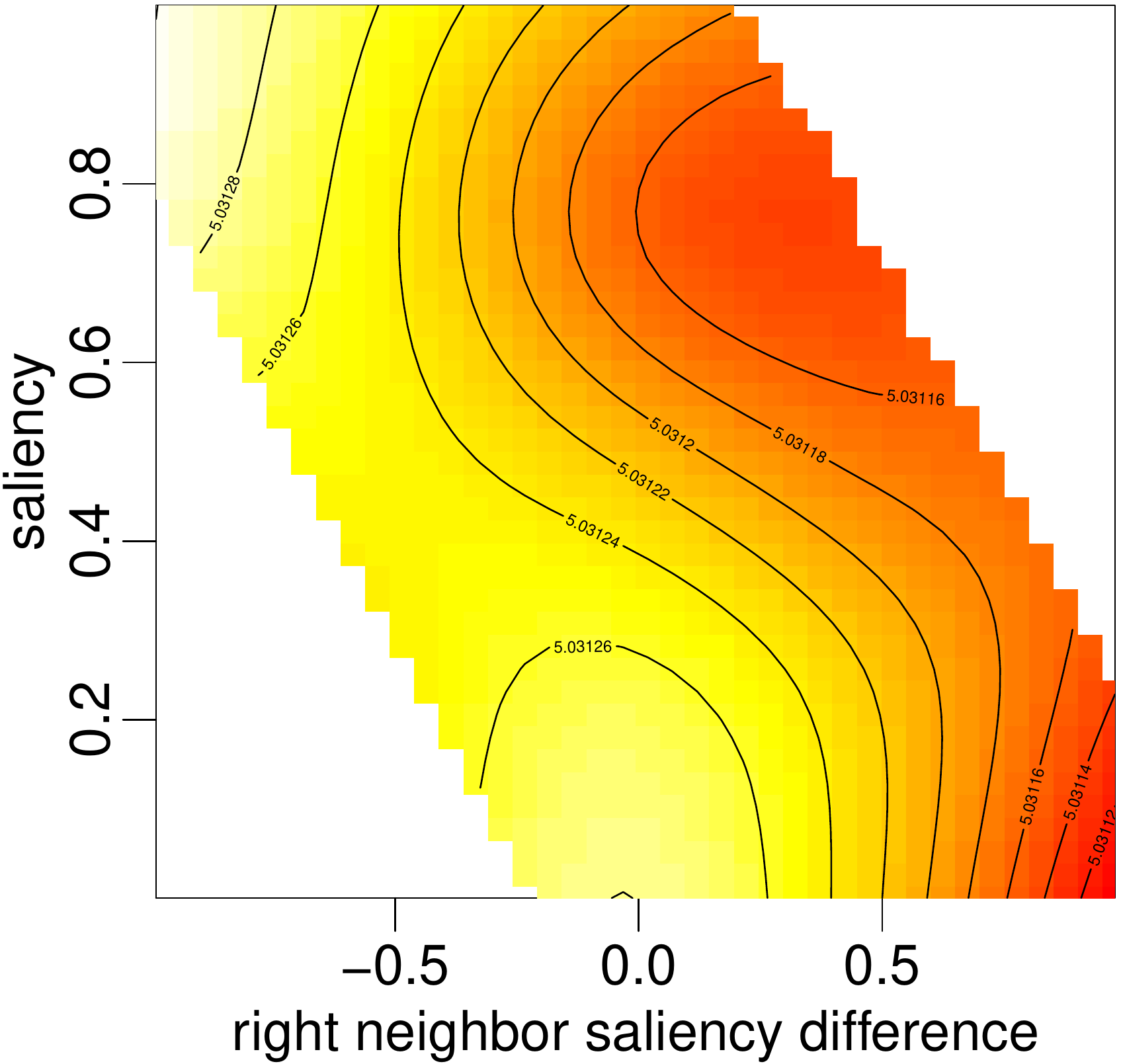}
    \caption{Right, chunk.}\label{fig:right_neighbor_saliency_difference_saliency_lexical_chunk}
    \end{subfigure}
    \caption{Left and right neighbours. ($\ast$) marks statistically significant smooths. Colors are normalized per figure.}\label{fig:left_and_right_neighbours}
\end{figure*}

Ideally, the importance ratings of a word will be explained entirely by its saliency strength. However, previous work showed that this is not the case. Here, we are interested in whether and how much the participants' answers can be explained by properties of neighboring words, \textit{beyond} what can be explained by the rated word's saliency alone. 
\\[0.5em] 
\noindent\textbf{Modeling.}
We analyze the collected ratings using an ordinal generalized additive mixed model (GAMM).\footnote{Introductory description in \Cref{app:technical-details}.}
Its key properties are that it models the ordinal response variable (i.e., the importance ratings in our setting) on a continuous latent scale as a sum of smooth functions of covariates, while also accounting for random effects.\footnote{Random effects allow to control for, e.g., systematic differences in individual participants' rating behaviour, such as a specific participant with a tendency to give overall higher ratings than other participants.}
\\[0.5em] 
\noindent\textbf{Precedent model terms.}
We include all covariates tested by \citet{DBLP:conf/fat/SchuffJAGV22}, including the rated word's saliency, word length, and so on, in order to control for them when testing our new phrase-level variables.
We follow \citeauthor{DBLP:conf/fat/SchuffJAGV22}'s controls for all precedent main and random effects.\footnote{We exclude the pairwise interactions from their modeling, due to increased  stability without losing expressiveness.}
\\[0.5em] 
\noindent\textbf{Novel neighbor terms.}
The following variables dictate our added model terms as the basis for the analysis: Left or right adjacency; rated word's saliency (color intensity); saliency difference between the two words; and whether the  words hold a weak or strong relationship.
We include four new bivariate smooth term (\Cref{fig:left_and_right_neighbours}) based on the interactions of the above variables. 

We refer to a bigram with a strong relationship as a chunk.
To arrive at a reliable measure for chunks, we methodically test various measures of bigram  relationships, in two different categories (\Cref{tab:chunk-measures}): \textit{syntactic}, via dependency parsing, and \textit{statistical}, via word collocation in a corpus.
Following \citet{Frantzi2000AutomaticRO}, we use both syntactic and statistical measures together, as first-order constituents among the 0.875 percentile for \textit{$\upvarphi^2$} collocations (our observations are robust to choices of statistical measure and percentile; see \Cref{app:robustness}).

\section{Reproducing Prior Results} \label{sec:reproduction}
Our study is similar to the experiments of \citet{DBLP:conf/fat/SchuffJAGV22} who investigate the effects of word-level and sentence-level features on importance perception.
Thus, it is well-positioned to attempt a reproduction of prior observations, to confirm whether they persist in a different language domain: Medium-form Wikipedia texts vs. short-form restaurant reviews in \citeauthor{DBLP:conf/fat/SchuffJAGV22}, and SHAP-values vs. Integrated-Gradients \cite{DBLP:conf/icml/SundararajanTY17}.

The result is positive: We reproduce the previously reported
significant effects of \textit{word length}, \textit{display index} (i.e., the position of the rated instance within the 150 sentences), \textit{capitalization}, and \textit{dependency relation} for randomized explanations as well as SHAP-value explanations (details in \Cref{app:user-study-details}).
This result reinforces prior observations that human users are at significant risk of biased perception of saliency explanations despite an objective visualization interface.

\section{Neighbor Effects Analysis} \label{sec:analysis}
In the following, we present our results for our two experiments using (a) random saliency values and (b) SHAP values.

\begin{table}[t]
\centering
\resizebox{0.99\linewidth}{!}{
\begin{tabular}{p{4.85cm}rrrr}
  \toprule
\textbf{Term} & \textbf{(e)df} & \textbf{Ref.df} & \textbf{F} & $p$ \\ 
  \midrule
s(saliency) & 11.22 & 19.00 & 580.89 & \textbf{$<$0.0001} \\ 
  s(display index) & 3.04 & 9.00 & 22.02 & \textbf{$<$0.0001} \\ 
  s(word length) & 1.64 & 9.00 & 16.44 & \textbf{$<$0.0001} \\ 
  s(sentence length) & 0.00 & 4.00 & 0.00 & 0.425 \\ 
  s(relative word frequency) & 0.00 & 9.00 & 0.00 & 0.844 \\ 
  s(normalized saliency rank) & 0.59 & 9.00 & 0.37 & 0.115 \\ 
  s(word position) & 0.58 & 9.00 & 0.18 & 0.177 \\ 
  te(left diff.,saliency): no chunk & 3.12 & 24.00 & 1.50 & \textbf{0.002} \\ 
  te(left diff.,saliency): chunk & 2.24 & 24.00 & 0.51 & \textbf{0.038} \\ 
  te(right diff.,saliency): no chunk & 2.43 & 24.00 & 0.47 & \textbf{0.049} \\ 
  te(right diff.,saliency): chunk & 0.00 & 24.00 & 0.00 & 0.578 \\ 
  \midrule
  capitalization &2.00  & & 3.15 & \textbf{0.042} \\ 
  dependency relation & 35.00  & & 2.92 & \textbf{$<$0.0001} \\
   \bottomrule
\end{tabular}}
\caption{(Effective) degrees of freedom, reference degrees of freedom and Wald test statistics for the univariate smooth terms (top) and parametric terms (bottom) for our randomized saliency experiment.
} 
\label{tab:stats_main}
\end{table}

\subsection{Randomized Explanations}
Regarding our additionally introduced neighbor terms, \Cref{fig:left_and_right_neighbours} shows the estimates for the four described functions (left/right $\times$ chunk/no chunk).
\Cref{tab:stats_main} lists all smooth and parametric terms along with Wald test results \cite{wood2013p,wood2013simple}.
\Cref{app:user-study-details} includes additional results.

\begin{figure*}[h!]
    \centering
    \begin{subfigure}[t]{.25\textwidth}
        \centering
        \includegraphics[width=\textwidth]{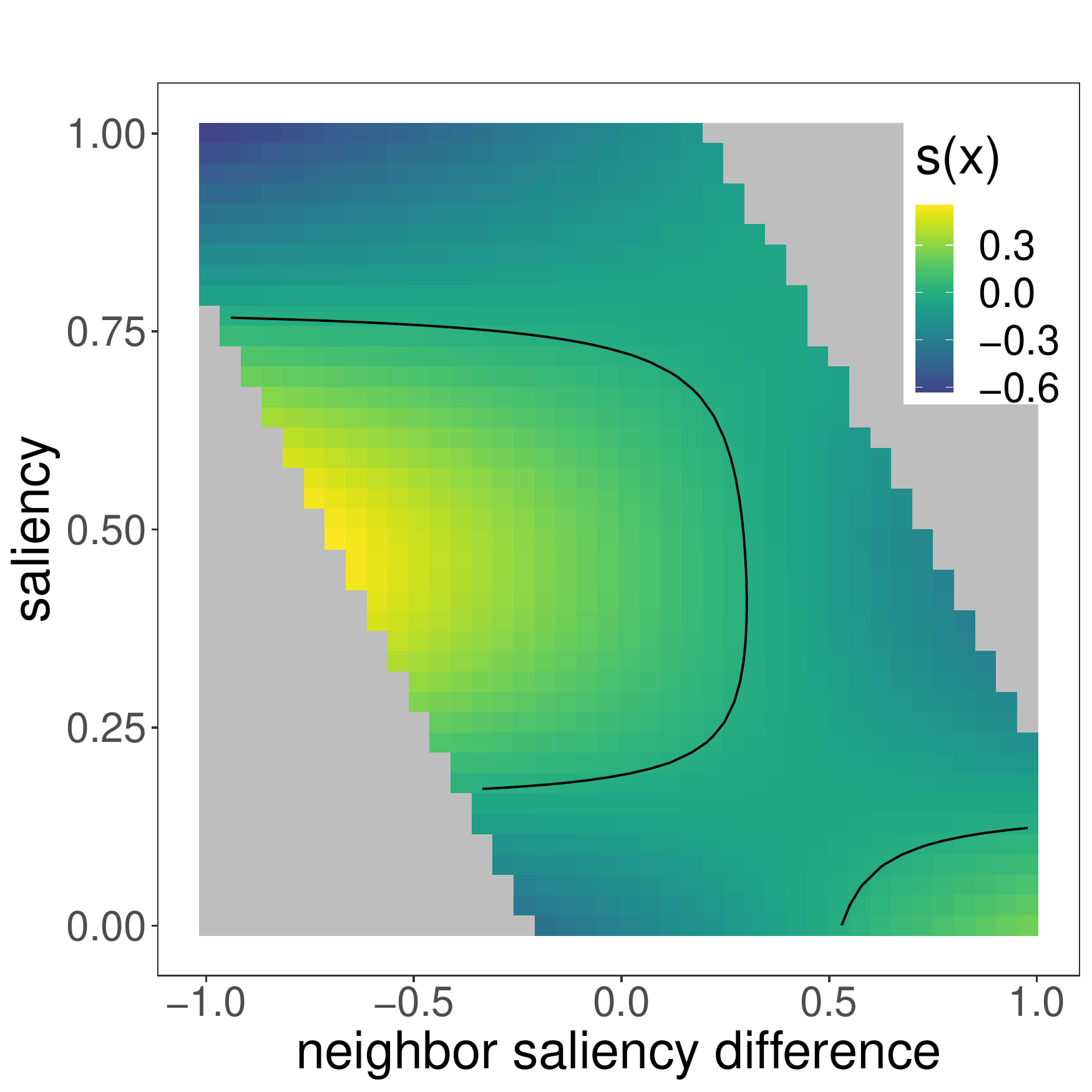}
        \caption{Right/left difference for chunks (contour marks 0).}\label{fig:main_diff_plot_chunk_val}
    \end{subfigure}%
    \hspace{1cm}
    \begin{subfigure}[t]{.25\textwidth}
        \centering
        \includegraphics[width=\textwidth]{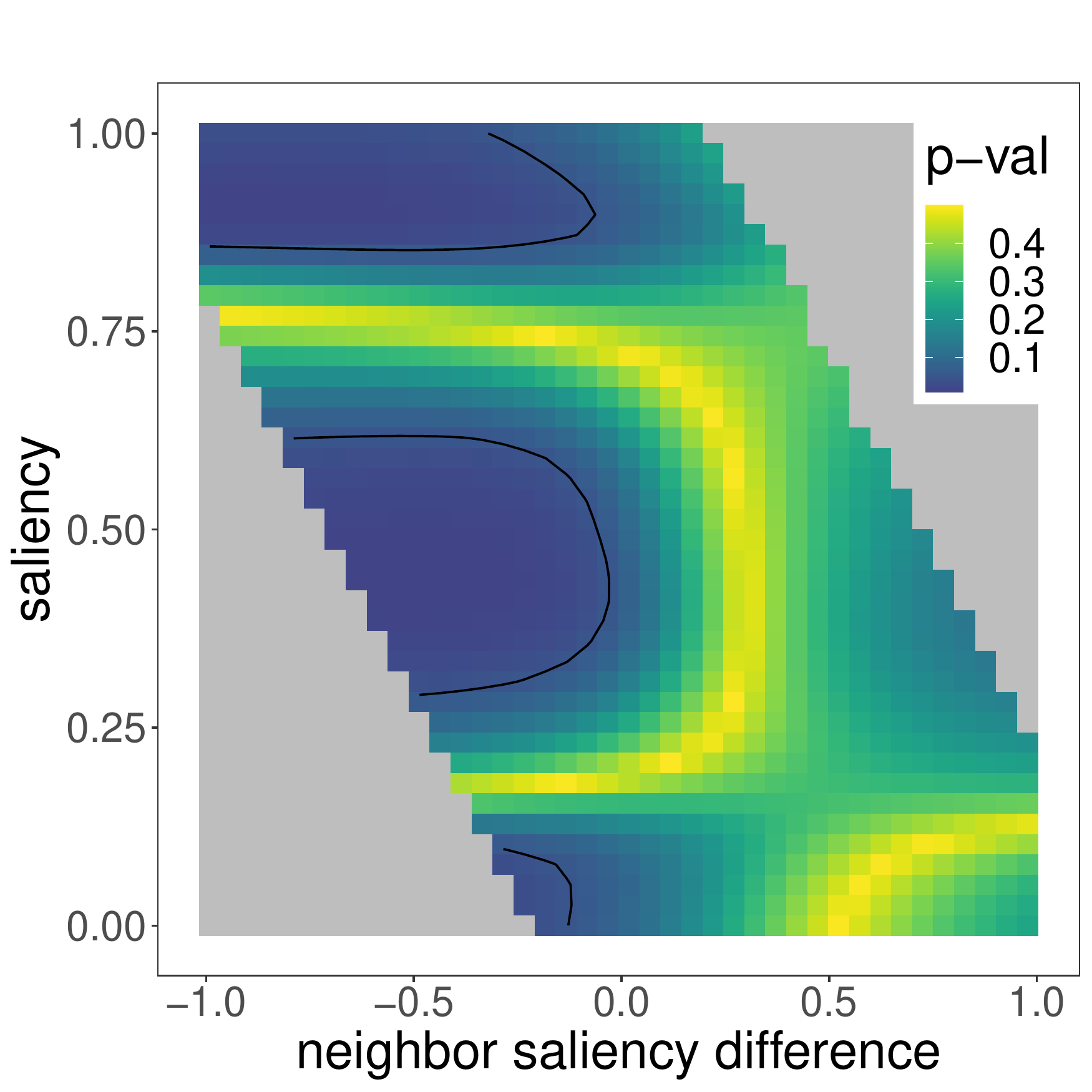}
        \caption{Right/left difference (chunk) $p$ values (contour marks 0.05).}\label{fig:main_diff_plot_chunk_p}
    \end{subfigure}%
    \hspace{1cm}
    \begin{subfigure}[t]{.25\textwidth}
        \centering
        \includegraphics[width=\textwidth]{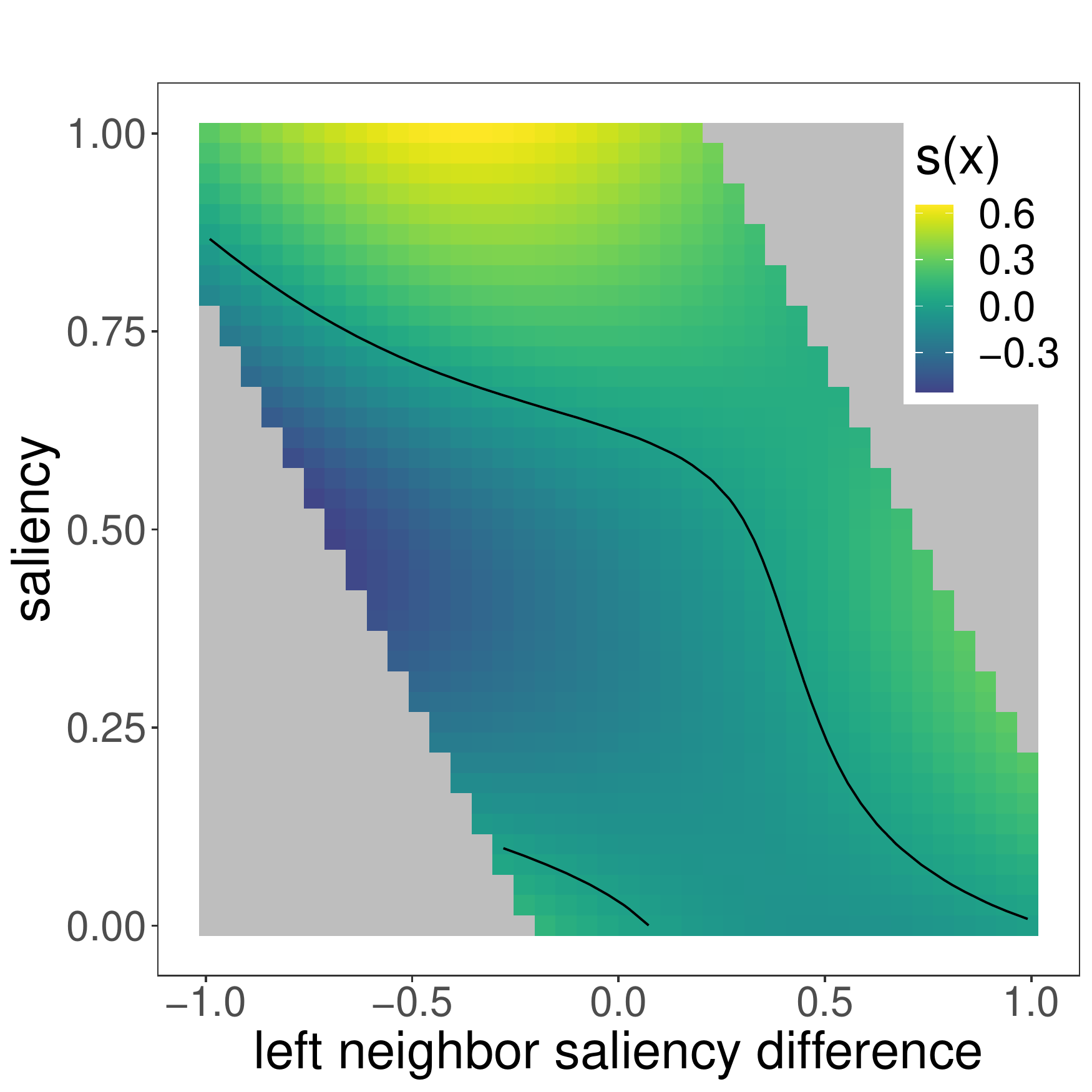}
        \caption{Chunk/no chunk difference for left neighbor (contour marks 0).}\label{fig:main_diff_plot_left_val}
    \end{subfigure}

    \begin{subfigure}[t]{.25\textwidth}
        \centering
        \includegraphics[width=\textwidth]{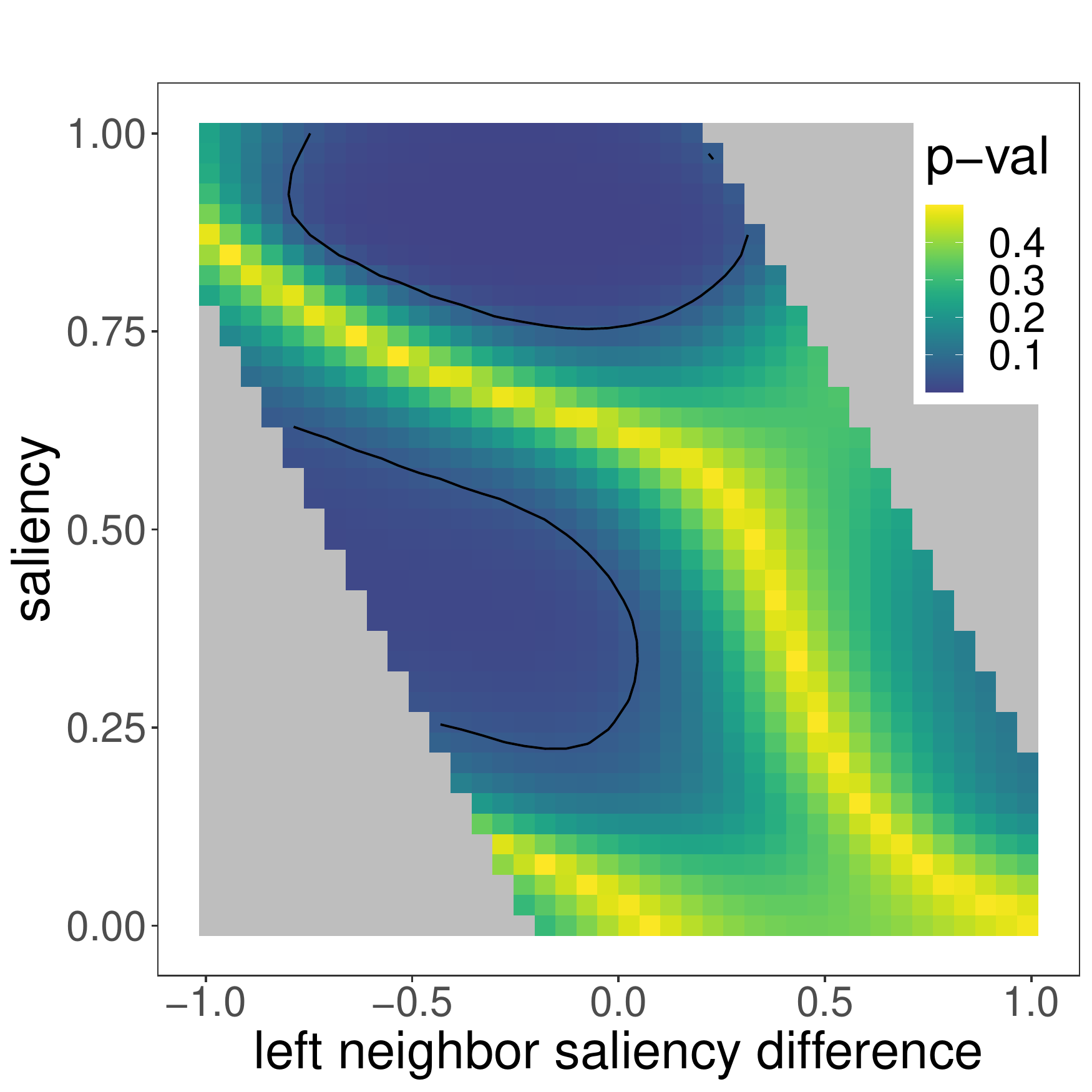}
        \caption{Chunk/no chunk difference (left) $p$ values (contour marks 0.05).}\label{fig:main_diff_plot_left_p}
    \end{subfigure}%
    \hspace{1cm}
    \begin{subfigure}[t]{.25\textwidth}
        \centering
        \includegraphics[width=\textwidth]{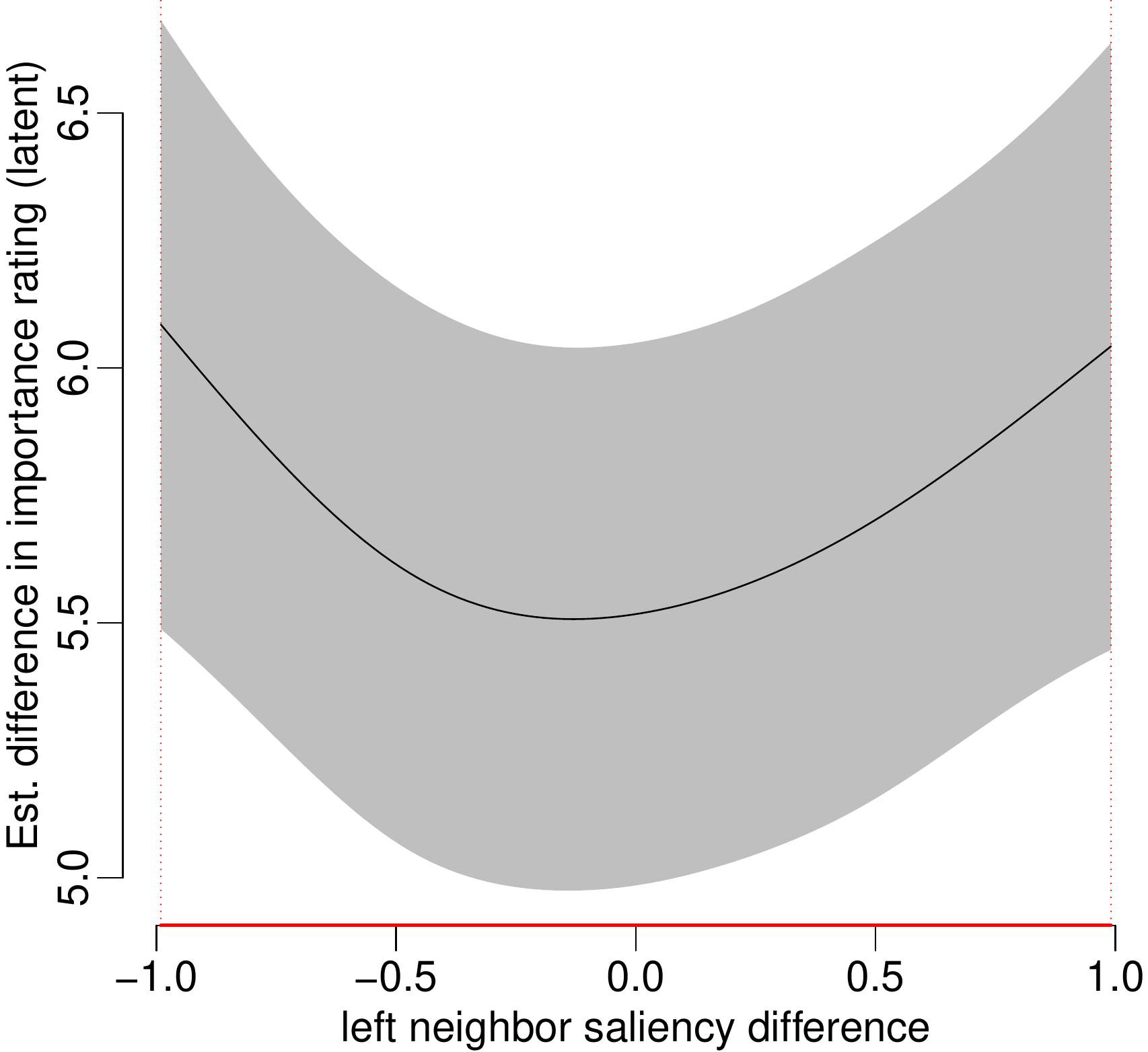}
        \caption{Difference between rated saliency and left neighbor saliency.
        }\label{fig:main_diff_saliency_levels_left}
    \end{subfigure}%
    \caption{Difference plots. Contour refers to the contour line. Red x-axis in (e) marks significant differences.}\label{fig:difference_plots_main_paper}
\end{figure*}

\paragraph{Asymmetric influence.}
\Cref{fig:left_neighbor_saliency_difference_saliency_no_lexical_chunk} vs.  \Cref{fig:right_neighbor_saliency_difference_saliency_no_lexical_chunk} and \Cref{fig:left_neighbor_saliency_difference_saliency_lexical_chunk} vs.  \Cref{fig:right_neighbor_saliency_difference_saliency_lexical_chunk} reveal qualitative differences between left and right neighbor's influences.
We quantitatively confirm these differences by calculating areas of significant differences \cite{fasiolo2020scalable,https://doi.org/10.1111/j.1467-9469.2011.00760.x}.
\Cref{fig:main_diff_plot_chunk_val,fig:main_diff_plot_chunk_p} show the respective plots of (significant) differences and probabilities for the chunk case.
Overall, we conclude that the influence from left and right word neighbors is significantly different.
\\[0.5em] 
\noindent\textbf{Chunk influence.}
We investigate the difference between neighbors that are within a chunk with the rated word vs. those that are not.
We find qualitative differences in \Cref{fig:left_and_right_neighbours} as well as statistically significant differences (\Cref{fig:main_diff_plot_left_val,fig:main_diff_plot_left_p}).
\\[0.5em] 
\noindent\textbf{Saliency moderates neighbor difference.}
\Cref{fig:left_and_right_neighbours} shows that the effect of a neighbor's saliency difference (x-axis) is moderated by the rated word's saliency (y-axis).
We confirm this observation statistically (\Cref{fig:main_diff_saliency_levels_left})  by comparing functions at a rated word saliency of 0.25 and 0.75, using unidimensional difference plots \cite{van2015itsadug}.
\\[0.5em] 
\noindent\textbf{Combined effects.}
We identify two general opposing effects: assimilation and contrast.\footnote{We borrow these terms from psychology (\Cref{sec:relation_psychology}).}

We refer to \textit{Assimilation} as situations where a word's perceived saliency is
perceived as more (or less) important based on whether its neighbor has a higher (or lower) saliency.
We find assimilation effects from \textit{left} neighbors that form a chunk with a moderate saliency (0.25--0.75) rated word. 

We refer to \textit{Contrast} as situations where a word's perceived saliency is perceived as less (or more) important based on whether its neighbor has a higher (or lower) saliency.
We find contrast effects from left and right neighbors that do not form a chunk with the rated word.\footnote{Note that although  \Cref{fig:right_neighbor_saliency_difference_saliency_lexical_chunk} suggests a contrast effect,
the color normalization inflates the minimal differences in this figure and the Wald tests did \textit{not} signal a significant effect.}

\subsection{SHAP-Value Explanations}
\paragraph{Shared results.}
Our SHAP-value experiment confirms our observation of (i) asymmetric influence of left/right neighbors (\Cref{fig:diff_plot_chunk_val_shap,fig:diff_plot_chunk_p_shap}), (ii) chunk influence (\Cref{fig:diff_plot_right_val_shap,fig:diff_plot_right_p_shap}), (iii) a moderating effect of saliency (\Cref{fig:diff_saliency_levels_right_shap}), and (iv) assimilation and contrast effects (\Cref{fig:right_neighbor_saliency_difference_saliency_lexical_chunk_shap}).

\paragraph{Variant results.}
Notably, our SHAP-value results differ from our randomized saliency results with respect to the effects left/right direction.
For the randomized saliency experiment, we observe assimilation effects from left neighbors within a chunk (\Cref{fig:left_neighbor_saliency_difference_saliency_lexical_chunk}) and contrast effects from left and right neighbors outside a chunk (\Cref{fig:left_neighbor_saliency_difference_saliency_no_lexical_chunk,fig:right_neighbor_saliency_difference_saliency_no_lexical_chunk}).
For our SHAP-value experiment, we observe assimilation (low rated word saliencies) and contrast effects (medium normalized rated word saliencies) from right neighbors within a chunk (\Cref{fig:right_neighbor_saliency_difference_saliency_lexical_chunk_shap}).
We hypothesize that this difference can be attributed to the inter-dependencies of SHAP values as indicated in \Cref{fig:distributions} in \Cref{app:technical-details}.

\pagebreak

\subsection{Takeaways}
\textit{Overall, we find that} (a) left/right influences are not the same, (b) strong bigram relationships can invert contrasts into assimilation for left neighbors, (c) extreme saliencies can inhibit assimilation, and (d) biasing effects can be observed for randomized explanations as well as SHAP-value explanations.

\section{Theoretical Grounds in Psychology}\label{sec:relation_psychology}
The assimilation effect is, of course, intuitive---it simply mean that neighbors' importance ``leaks'' from neighbor to the rated word for strong bigram relationships.
But is there precedence for the observed assimilation and contrast effects in the literature? How do they relate to each other?

Psychology investigates how a prime (e.g., being exposed to a specific word) influences human judgement, as part of two categories: \textit{assimilation} (the rating is ``pulled'' towards the prime) and \textit{contrast} (the rating is ``pushed'' away from the prime) effects \cite[i.a.,][]{BLESS201626}.

\citet{forster2008effect} demonstrate how \textit{global} processing (e.g. looking at the overall structure) vs. \textit{local} processing (e.g., looking at the details of a structure) leads to assimilation vs. contrast.
We argue that some of our observations can be explained with their model: Multi-word phrase neighbors may induce global processing that leads to assimilation (for example, in the randomized explanation experiments, left neighbors) while other neighbors (in the randomized explanation experiments, right neighbors and unrelated left neighbors) induce local processing that leads to contrast.
Future work may investigate the properties that induce global processing in specific contexts.

\section{Conclusions}
We conduct a user study in a setting of laypeople observing common word-importance explanations, as color-coded importance, in the English Wikipedia domain. In this setting, we find that when the explainee understands the attributed importance of a word, the importance of \textit{other words} can influence their understanding in unintended ways.

Common wisdom posits that when communicating the importance of a component in a feature-attribution explanation, the explainee will understand this importance as it is shown. We find that this is not the case: The explainee's contextualized understanding of the input portion---for us, a word as a part of a phrase---may influence their understanding of the explanation. 

\section*{Limitations}

The observed effects in this work, in principle, can only be applied to the setting of our user study (English text, English-speaking crowd-workers, color-coded word-level saliency, and so on, as described in the paper). Therefore this study serves only as a \textit{proof of existence}, for a reasonably plausible and common setting in NLP research, that laypeople can be influenced by context outside of the attributed part of the input when comprehending a feature-attribution explanation. Action taken on design and implementation of explanation technology for NLP systems in another setting, or other systems of similar nature, should either investigate the generalization of effects to the setting in practice (towards which we aim to release our full reproduction code), or take conservative action in anticipation that the effects will generalize without compromising the possibility that they will not. 

\section*{Acknowledgements}
We are grateful to Diego Frassinelli and the anonymous reviewers for valuable feedback and helpful comments.
A. Jacovi and Y. Goldberg received funding from the European Research Council (ERC) under the European Union’s Horizon 2020 research and innovation programme, grant agreement No. 802774 (iEXTRACT).
N.T. Vu is funded by Carl Zeiss Foundation.

\bibliography{anthology,custom}
\bibliographystyle{acl_natbib}


\appendix

\begin{figure*}[!ht]
\centering
\fbox{\includegraphics[width=0.8\linewidth]{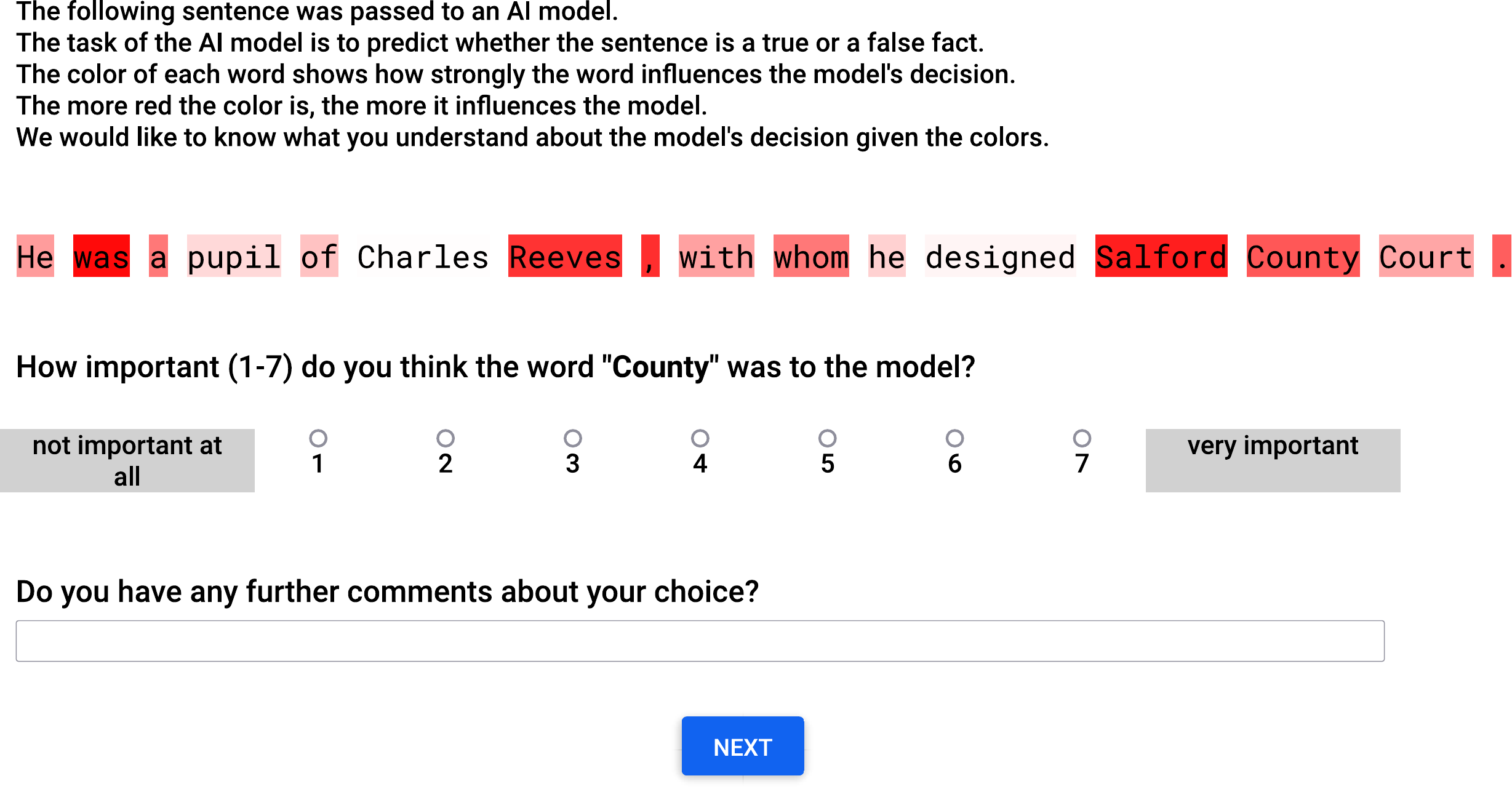}}
\caption{Screenshot of the rating interface.}
\label{fig:gui_2}
\end{figure*}

\section{User Study Details}
This section provides details on our user study setup.
\label{app:user-study-details}

\subsection{Interface}
\begin{figure*}[t]
\centering
\fbox{\includegraphics[width=0.7\linewidth]{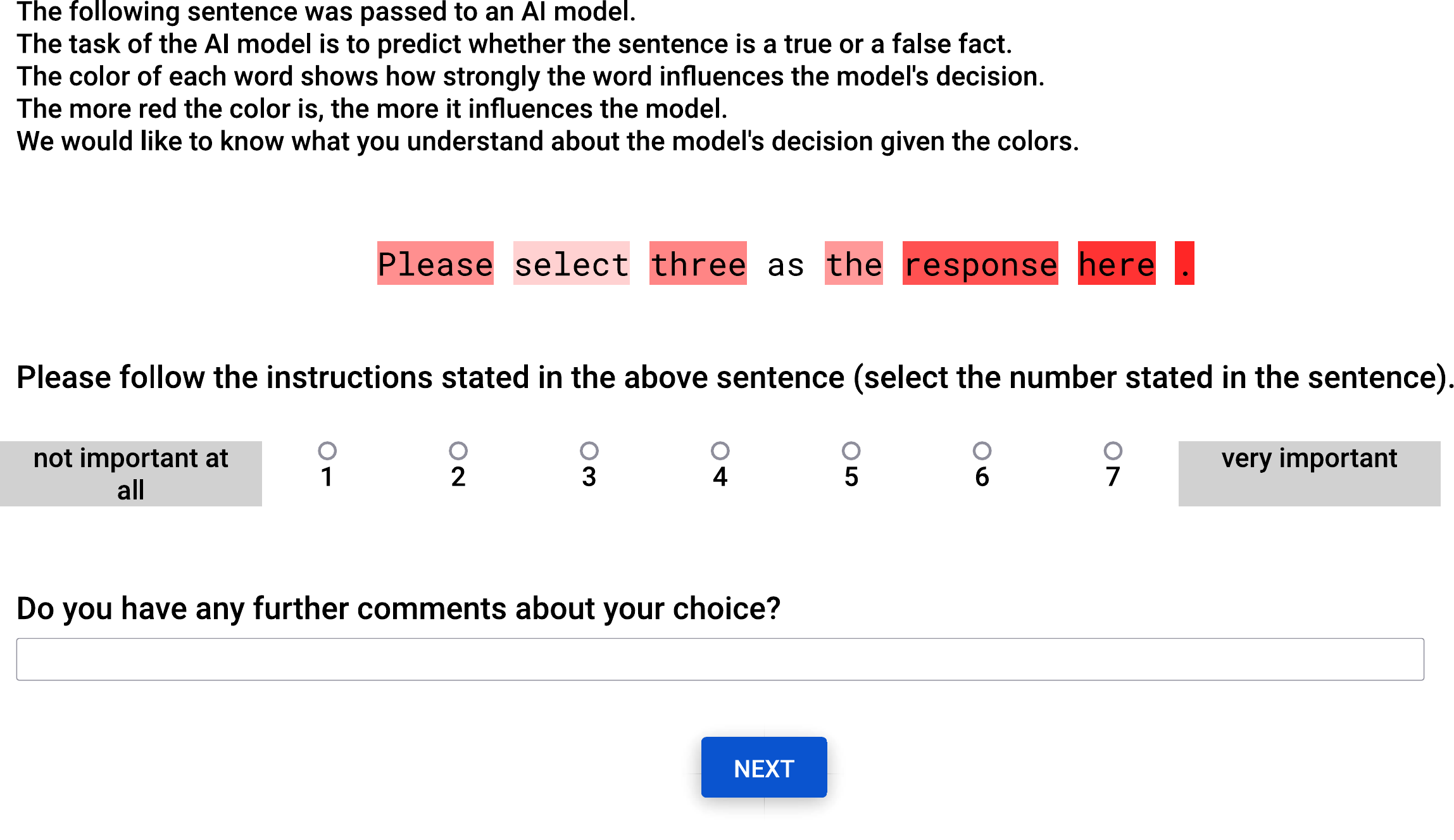}}
\caption{Screenshot of the rating interface for an attention check.}
\label{fig:gui_trap}
\end{figure*}

\Cref{fig:gui_2} shows a screenshot of our rating interface.
\Cref{fig:gui_trap} shows a screenshot of an attention check.

\subsection{Attention Checks}
We include three attention checks per participants which we randomly place within the last two thirds of the study following \citet{DBLP:conf/fat/SchuffJAGV22}.

\subsection{Participants}
In total, we recruit 76 crowd workers from English-speaking countries via Amazon Mechanical Turk for our randomized explanation study and 36 crowd workers for our SHAP-value explanation study.
We require workers to have at least 5,000 approved HITs and 95\% approval rate.
Raters are screened with three hidden attention checks that they must answer correctly to be included (but are paid fully regardless).
From the 76 workers, 64 workers passed the screening, i.e., we excluded 15.8\% of responses on a participant level.
From the 36 workers, all workers passed the screening.
On average, participants were compensated with an hourly wage of US\$8.95.
We do not collect any personally-identifiable data from participants.

\section{Statistical Model Details}
In this section, we give a brief general introduction to statistical model we used (i.e., GAMM) and provide additional results of our analysis.
\label{app:technical-details}

\subsection{Introduction to GAMM Models}
We refer to the very brief introduction to GAMMs in \citet{DBLP:conf/fat/SchuffJAGV22} (appendix).
Very briefly, an ordinal GAMM can be described as a generalized additive model that additionally accounts for random effects and models ordinal ratings via a continuous latent variable that is separated into the ordinal categories via estimated threshold values.
For further details, \citet{divjak_ordinal_2017} provide a practical introduction to ordinal GAMs in a linguistic context and \citet{wood_generalized_2017} offers a detailed textbook on GAM(M)s including implementation and analysis details.

\begin{table}[t]
\ra{1.5}
\centering
\resizebox{0.99\linewidth}{!}{
\begin{tabular}{p{10cm}}
  \toprule 
 \textbf{Examples} \\ 
  \midrule 
The Emerging Pathogens Institute is an interdisciplinary research institution associated with the University of Florida.\\
Luca Emanuel Meisl (born 4 March 1999) is an Austrian footballer currently playing for FC Liefering. \\
The black-throated toucanet (Aulacorhynchus atrogularis) is a near-passerine bird found in central Ecuador to western Bolivia.\\
Christopher Robert Coste (born February 4, 1973) is an author and former Major League Baseball catcher.\\
WGTA surrendered its license to the Federal Communications Commission (FCC) on November 3, 2014.\\
   \bottomrule
\end{tabular} }
\caption{Examples of Wikipedia sentences used in our study.
}
\label{tab:sentence-samples}
\end{table}

\subsection{Model Details in Our Analysis}
We control for all main effects (word length, sentence length etc.) as well as all random effects used by \citet{DBLP:conf/fat/SchuffJAGV22}.
We exclude the pairwise interactions due to model instability when including the interactions.

We additionally include four new novel bivariate smooth terms.
Each of these terms models a tensor product of saliency (i.e. the rated word's color intensity) and the neighboring (left or right) word's saliency difference to the rated word.
For each side (left and right), we model the smooths for neighbors that (i) are within a lexical chunk to the rated word and (ii) are not.
\Cref{fig:left_and_right_neighbours} shows the estimated four (bivariate) functions.

\subsection{Data Preprocessing}
Following \citet{DBLP:conf/fat/SchuffJAGV22}, we exclude ratings with a completion time of less than a minute (implausibly fast completion) and exclude words with a length over 20 characters.
We effectively exclude 1.8\% of ratings.

In order to analyze left as well as right neighbors, we additionally have to ensure that we only include ratings for which both---left and right--- neighbors exist.
Therefore, we additionally exclude rating for which the leftmost or rightmost word in the sentence was rated.
This excludes 11.7\% of ratings.
In total, we thus use 9489 ratings to fit our model.

\subsection{Chunk Measures}
We explore and combine two approaches of identifying multi-word phrases (or ``chunks)''.
\paragraph{Syntactic measures (constituents).}
We first apply binary chunk measures based on the sentences' parse trees.
We use Stanza \cite{qi2020stanza} (version 1.4.2) to generate parse tree for each sentence.
We assess whether the rated word and its neighbor (left/right) share a constituent at the lowest possible level.
Concretely, we (a) start at the rated word and move up one level in the parse tree and (b) start at the neighboring word and move up one level in the parse tree.
If we now arrived at the same node in the parse tree, we the rated word and its neighbor share a first-order constituent.
If we arrived at different nodes, they do not.
Restricting the type of first-level shared constituents to noun phrases yields a further category.
We provide respective examples for shared first-level constituents and the respective noun phrase constituents extracted from our data in \Cref{tab:all-measures} (upper part).

\paragraph{Statistical measures (cooccurrence scores).}
We additionally explore numeric association measures and calculate all available bigram collocation measures available in NLTK's \textit{BigramAssocMeasures} module\footnote{\url{https://www.nltk.org/_modules/nltk/metrics/association.html}}.
The calculation is based on the 7 million Wikipedia-2018 sentences in \textit{Wikipedia Sentences} (\Cref{footnote:wikisent}).
A description of each metric as well as top-scored examples on our data is provided in \Cref{tab:all-measures} (lower part).
We separate examples into examples that form a constituent vs. do not form a constituent to highlight the necessity to apply a constituent filter in order to get meaningful categorization into chunks vs. no chunks.

\begin{table*}[htpb]
\ra{1.2}
\centering
\resizebox{0.99\linewidth}{!}{
\begin{tabular}{>{\raggedright}p{3.5cm}>{\raggedright}p{5cm}>{\raggedright}p{5cm}p{6cm}}
  \toprule 
\textbf{Measure} & \textbf{Constituent Examples} & \textbf{No Constituent Examples} & \textbf{Description} \\ 
  \midrule 
\textit{First-order constituent} & highly developed, more than, such as, DVD combo, 4 million & --- & Smallest multi-word constituent sub-trees in the constituency tree. \\
\textit{Noun phrase}  & Tokyo Marathon, ski racer, the UK, a retired, the city & --- & Multi-word first-order noun phrase in the constituency tree. \\
\midrule
\textit{Mutual information} & as well, more than, ice hockey, United Kingdom, a species & is a, of the, in the, is an, it was & Bigram mutual information variant (per NLTK implementation).  \\
\textit{Frequency} & the United, the family, a species, an American, such as & of the, in the, is a, to the, on the & Raw, unnormalized frequency. \\
\textit{Poisson Stirling}  & an American, such as, a species, as well, the family & is a, of the, in the, is an, it was, has been & Poisson Stirling bigram score. \\
\textit{Jaccard}  & Massar Egbari, ice hockey, Air Force, more than, Udo Dirkschneider & teachers/students teaching/studying, is a, has been, it was, of the &  Bigram Jaccard index.\\ 
\textit{$\upvarphi^2$}  & Massar Egbari, ice hockey, Udo Dirkschneider, Air Force, New Zealand & teachers/students teaching/studying, is a, has been, footballer who, is an & Square of the Pearson correlation coefficient. \\ 
   \bottomrule
\end{tabular} }
\caption{The list of phrase measures we tested for. Examples for numeric measures are chosen based on highest cooccurrence scores whereas the (boolean) noun phrase and constituent examples are chosen arbitrarily. For the numeric measures, we provide examples that (a) form a constituent with their neighbor and (b) do not. The examples underline the necessity to combine numeric scores with a constituent filter.
\label{tab:all-measures}
}
\end{table*}

\subsection{Detailed Results}
As described in \Cref{sec:analysis}, we observe different influences of left/right neighbors, chunk/no chunk neighbors as well as rated word saliency levels in our randomized explanation experiment.

\begin{figure*}[h!]
    \centering
    \begin{subfigure}[t]{.24\textwidth}
        \centering
    \includegraphics[width=\textwidth]{figures/phi_sq_875/diff_plot_chunk_val.pdf}
    \caption{Difference between right - left (chunk). The contour line marks zero.}\label{fig:diff_plot_chunk_val}
    \end{subfigure}%
    \hfill
    \begin{subfigure}[t]{.24\textwidth}
        \centering
    \includegraphics[width=\textwidth]{figures/phi_sq_875/diff_plot_chunk_p.pdf}
    \caption{Difference (chunk) $p$ values. The contour line marks 0.05.}\label{fig:diff_plot_chunk_p}
    \end{subfigure}%
    \hfill
    \begin{subfigure}[t]{.24\textwidth}
        \centering
    \includegraphics[width=\textwidth]{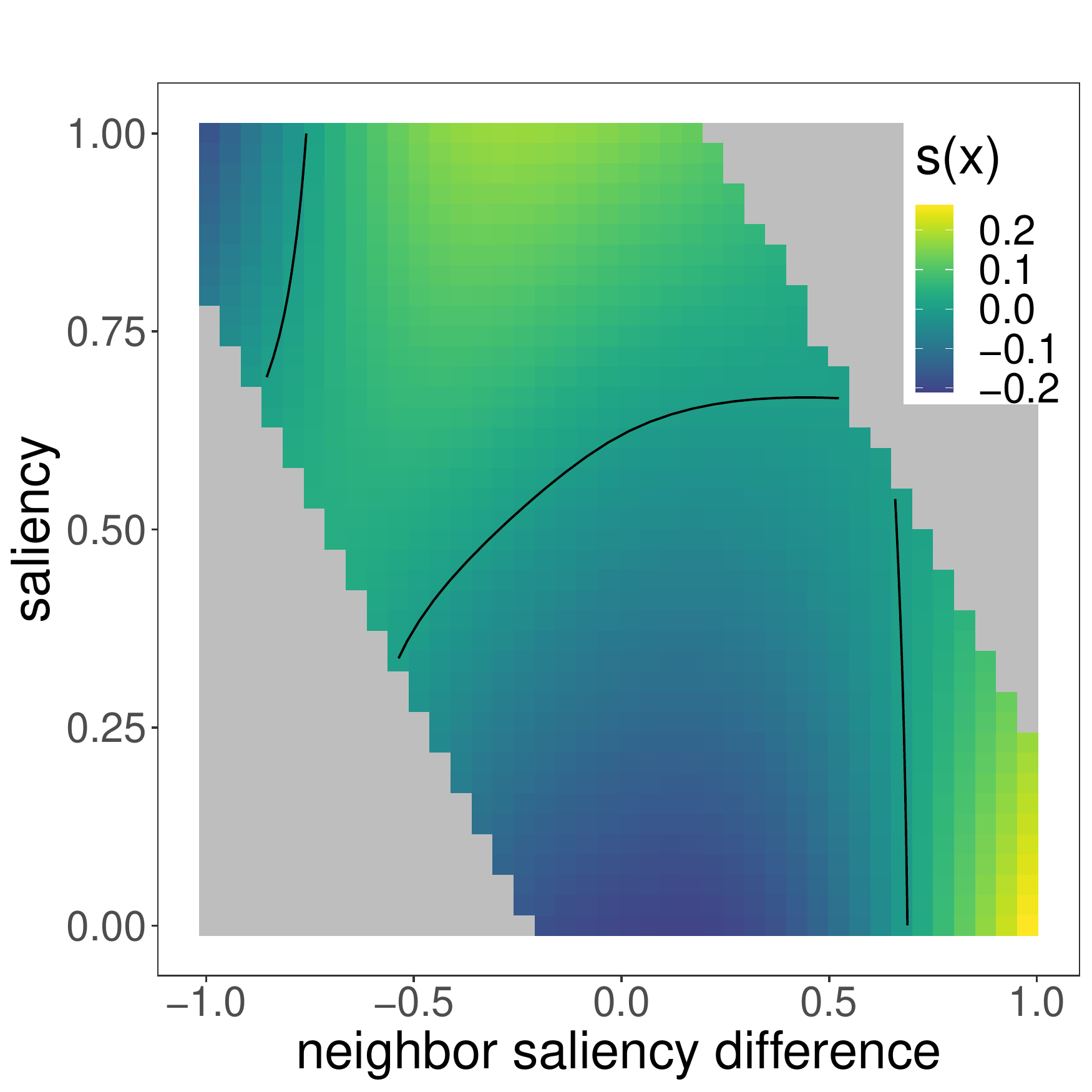}
    \caption{Difference between right - left (no chunk). The contour line marks zero.}\label{fig:diff_plot_no_chunk_val}
    \end{subfigure}%
    \hfill
    \begin{subfigure}[t]{.24\textwidth}
        \centering
    \includegraphics[width=\textwidth]{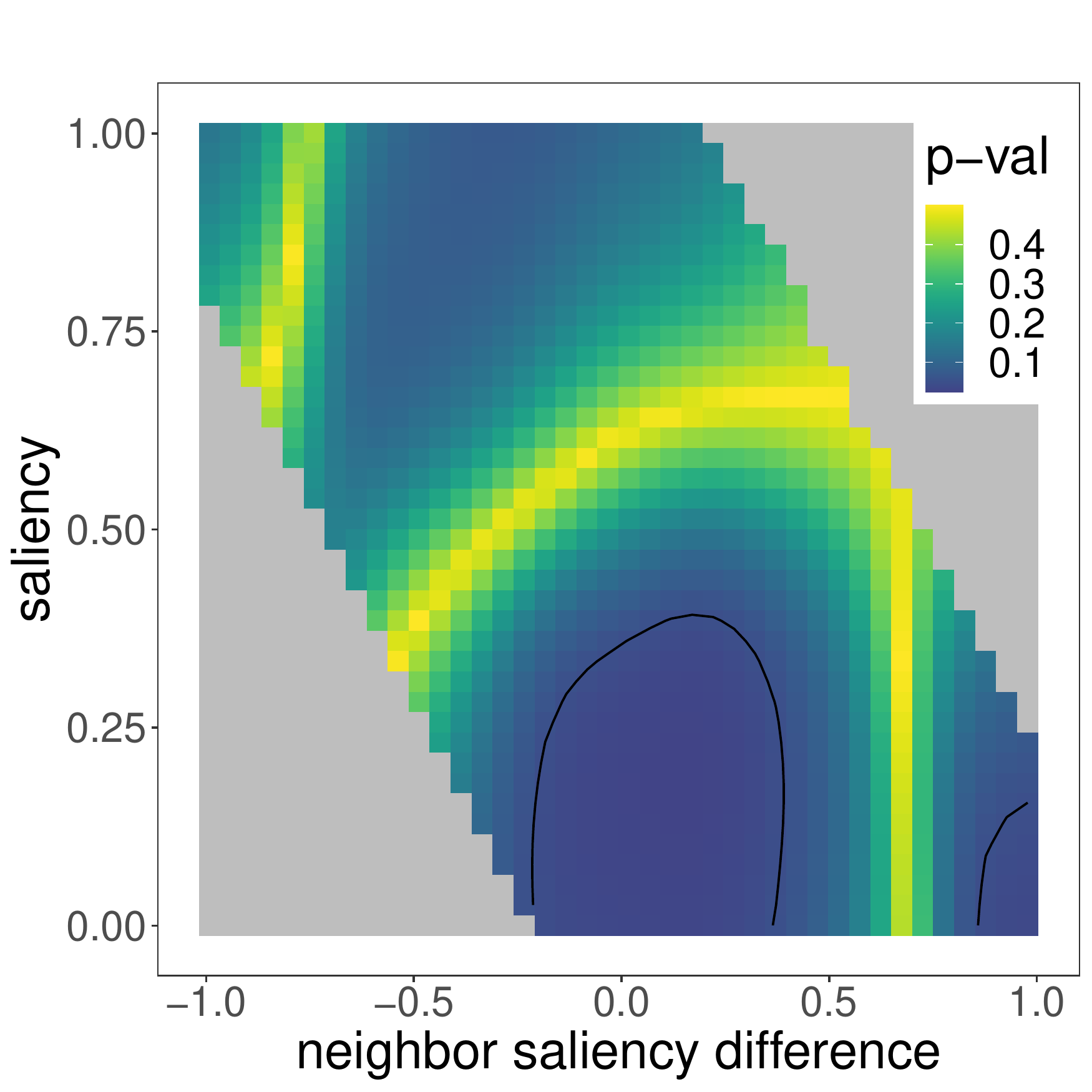}
    \caption{Difference (no chunk) $p$ values. The contour line marks 0.05.}\label{fig:diff_plot_no_chunk_p}
    \end{subfigure}
    \caption{Differences and $p$ values for (no) lexical chunk neighbors for our randomized explanation experiment.}\label{fig:difference_plots_chunk_no_chunk}
\end{figure*}

\paragraph{Left vs. right neighbors.}
\Cref{fig:difference_plots_chunk_no_chunk} shows difference plots (and respective p values) between left and right neighbors for chunk neighbors (\Cref{fig:diff_plot_chunk_val,fig:diff_plot_chunk_p}) and no chunk neighbors  (\Cref{fig:diff_plot_no_chunk_val,fig:diff_plot_no_chunk_p}).

\begin{figure*}[h!]
    \centering
    \begin{subfigure}[t]{.24\textwidth}
        \centering
    \includegraphics[width=\textwidth]{figures/phi_sq_875/diff_plot_left_val.pdf}
    \caption{Difference between (left neighbors) chunk - no chunk. The contour line marks zero.}\label{fig:diff_plot_left_val}
    \end{subfigure}%
    \hfill
    \begin{subfigure}[t]{.24\textwidth}
        \centering
    \includegraphics[width=\textwidth]{figures/phi_sq_875/diff_plot_left_p.pdf}
    \caption{Difference (left neighbors) $p$ values. The contour line marks 0.05.}\label{fig:diff_plot_left_p}
    \end{subfigure}%
    \hfill
    \begin{subfigure}[t]{.24\textwidth}
        \centering
    \includegraphics[width=\textwidth]{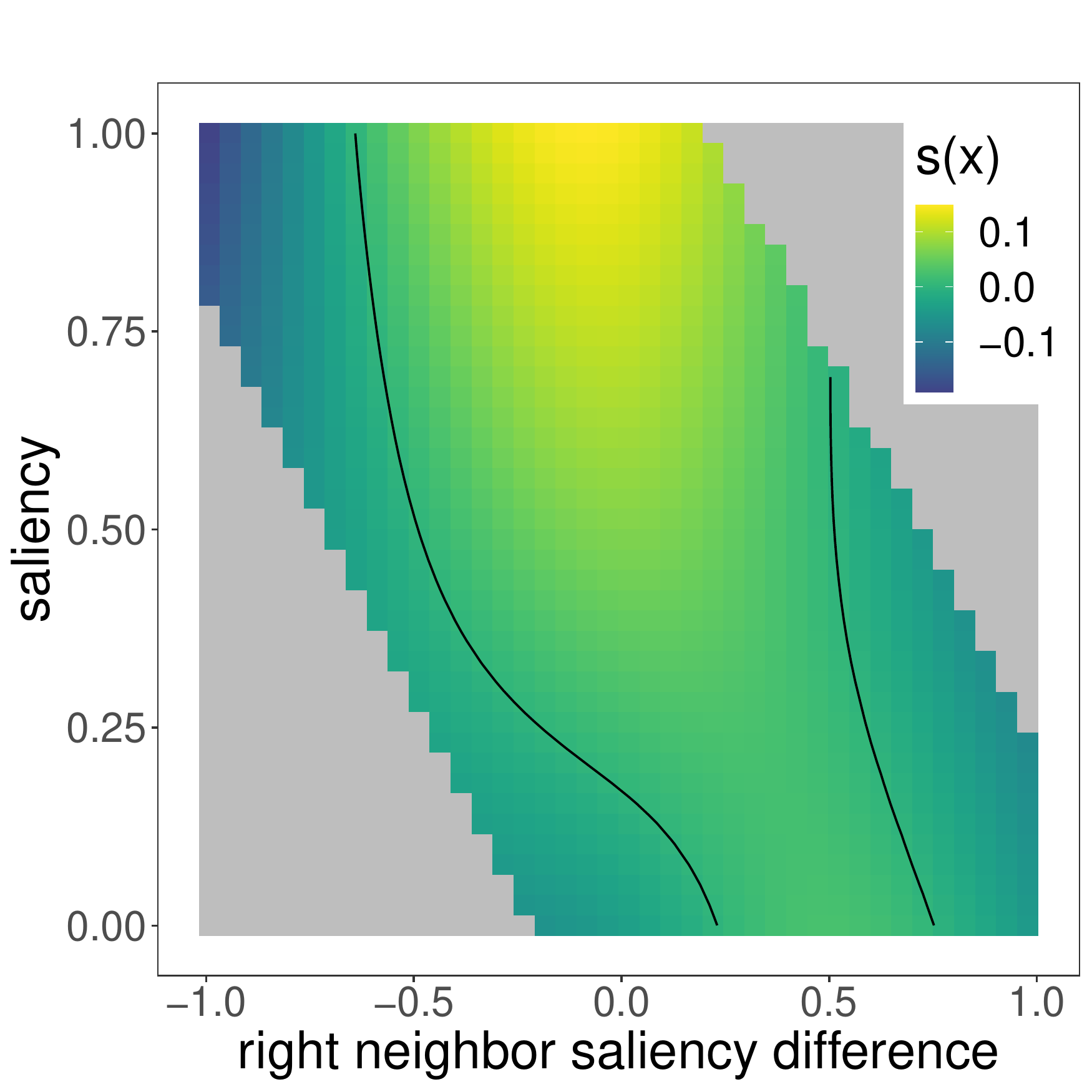}
    \caption{Difference between (right neighbors) chunk - no chunk. The contour line marks zero.}\label{fig:diff_plot_right_val}
    \end{subfigure}%
    \hfill
    \begin{subfigure}[t]{.24\textwidth}
        \centering
    \includegraphics[width=\textwidth]{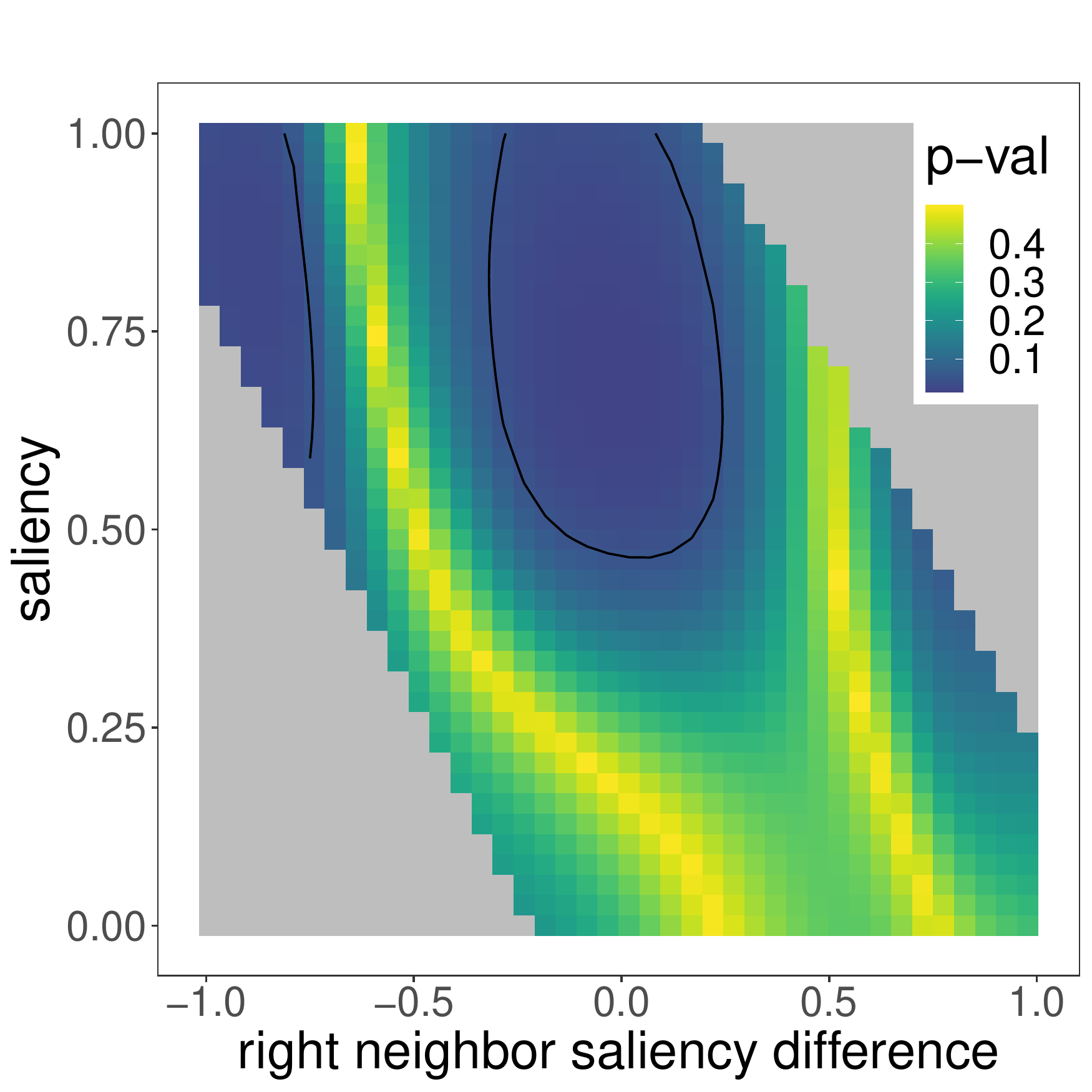}
    \caption{Difference (right neighbors) $p$ values. The contour line marks 0.05.}\label{fig:diff_plot_right_p}
    \end{subfigure}
    \caption{Differences and $p$ values for left and right neighbors for our randomized explanation experiment.}\label{fig:difference_plots_left_and_right}
\end{figure*}

\paragraph{Chunk vs. no chunk.}
Respectively, \Cref{fig:difference_plots_left_and_right} shows difference plots (and respective p values) between chunk and no chunk neighbors for left neighbors (\Cref{fig:diff_plot_left_val,fig:diff_plot_left_p}) and right neighbors  (\Cref{fig:diff_plot_right_val,fig:diff_plot_right_p}).

\begin{figure*}[h!]
    \centering
    \begin{subfigure}[t]{.24\textwidth}
        \centering
    \includegraphics[width=\textwidth]{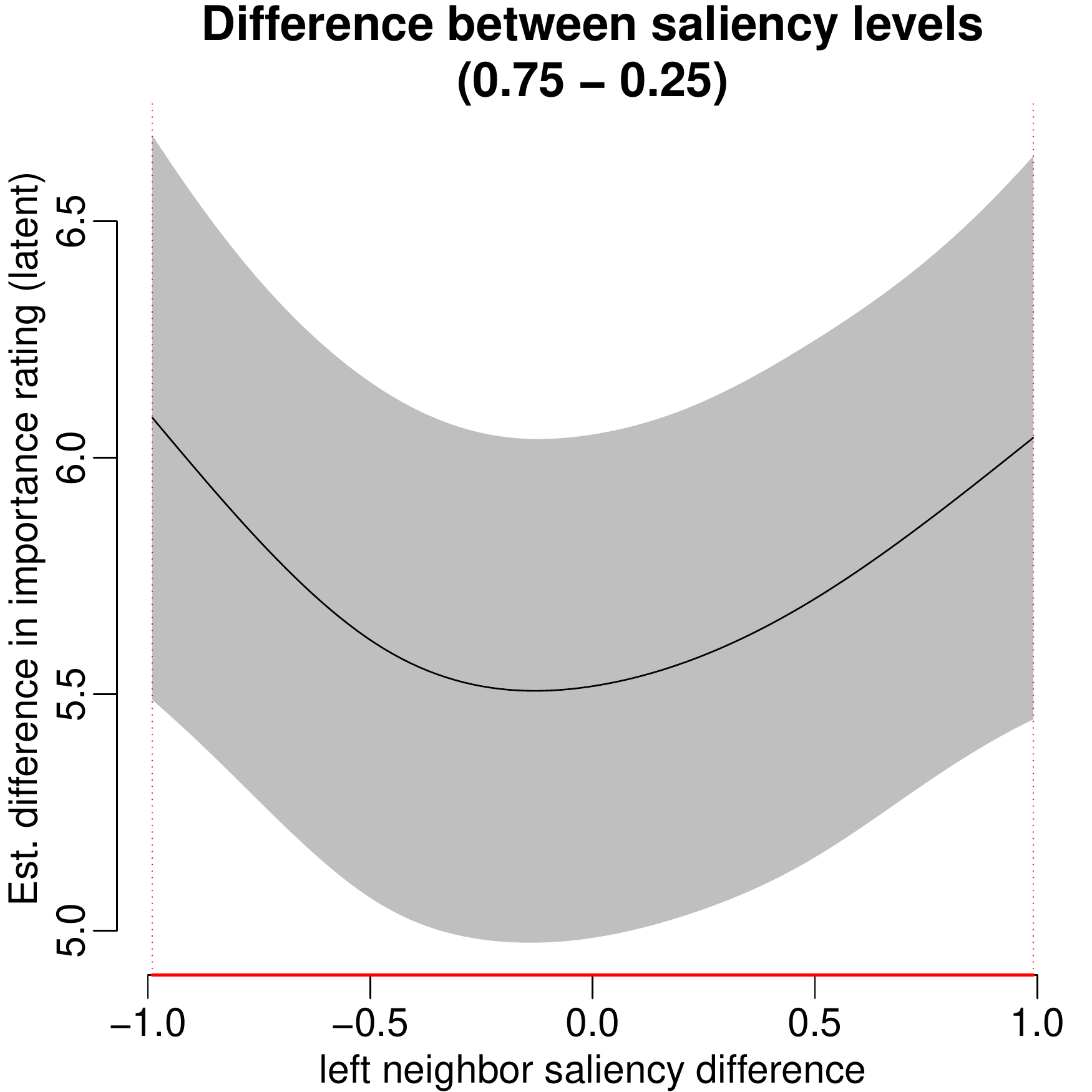}
    \caption{Left.}\label{fig:diff_saliency_levels_left}
    \end{subfigure}%
    \hspace{2cm}%
    \begin{subfigure}[t]{.24\textwidth}
        \centering
    \includegraphics[width=\textwidth]{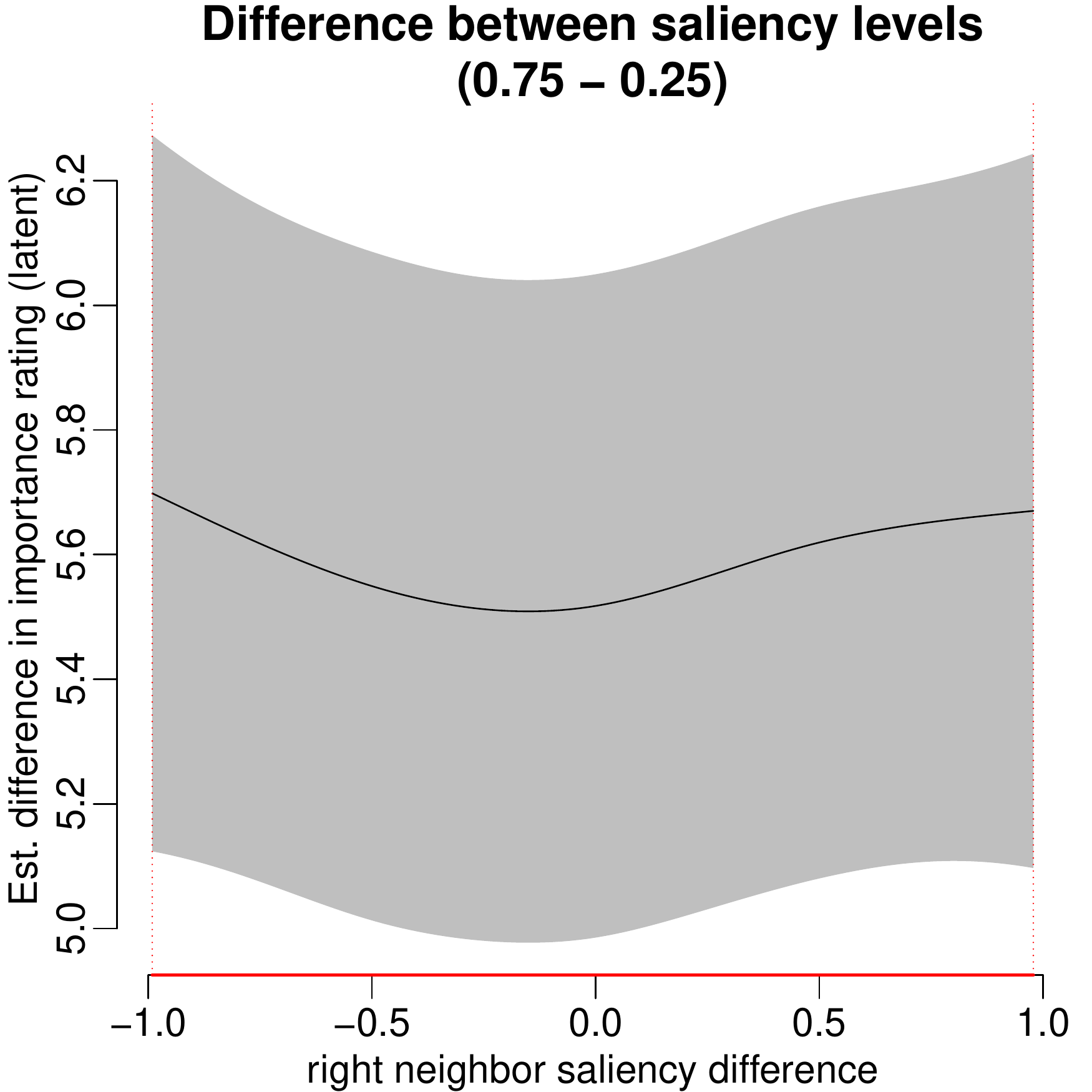}
    \caption{Right.}\label{fig:diff_saliency_levels_right}
    \end{subfigure}
    \caption{Difference plots between the influence of saliency differences between exemplary high (0.75) and low (0.25) rated word saliency levels. Red x-axis areas indicate significant differences.}\label{fig:difference_saliency_levels}
\end{figure*}

\paragraph{Differences across saliency levels.}
\Cref{fig:difference_saliency_levels} shows that the effects of saliency difference are significantly different between different levels of the rated word's saliency (0.25 and 0.75) for left neighbors (\Cref{fig:diff_saliency_levels_left}) as well as right neighbors (\Cref{fig:diff_saliency_levels_right}).

We report the detailed Wald test statistics for our randomized explanation experiment in \Cref{tab:stats_full}.

\subsection{SHAP-value Results}
\begin{figure*}[h!]
    \centering
    \begin{subfigure}[t]{.19\textwidth}
        \centering
        \includegraphics[width=\textwidth]{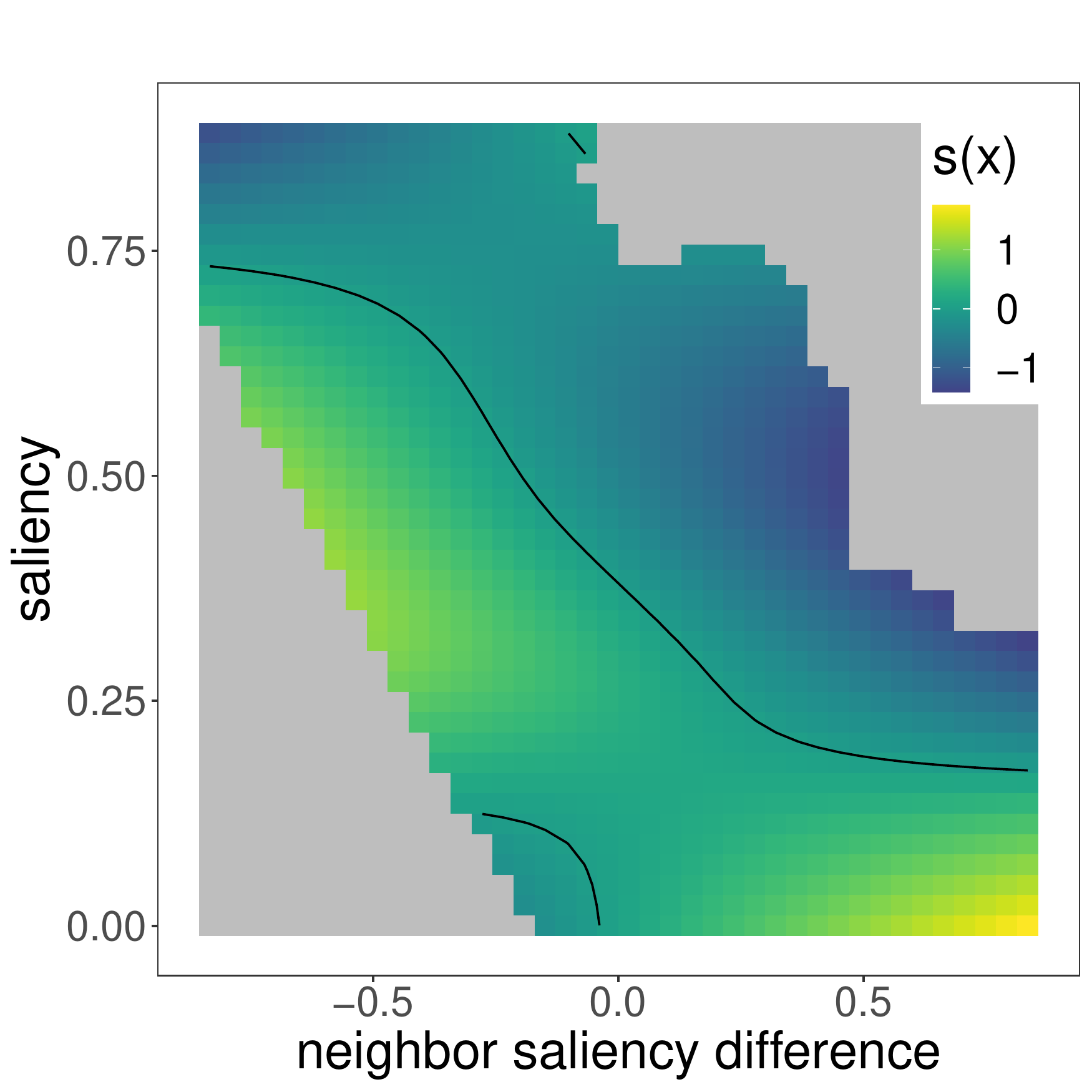}
        \caption{Right/left difference for chunks (contour marks 0).}\label{fig:diff_plot_chunk_val_shap}
    \end{subfigure}%
    \hfill
    \begin{subfigure}[t]{.19\textwidth}
        \centering
        \includegraphics[width=\textwidth]{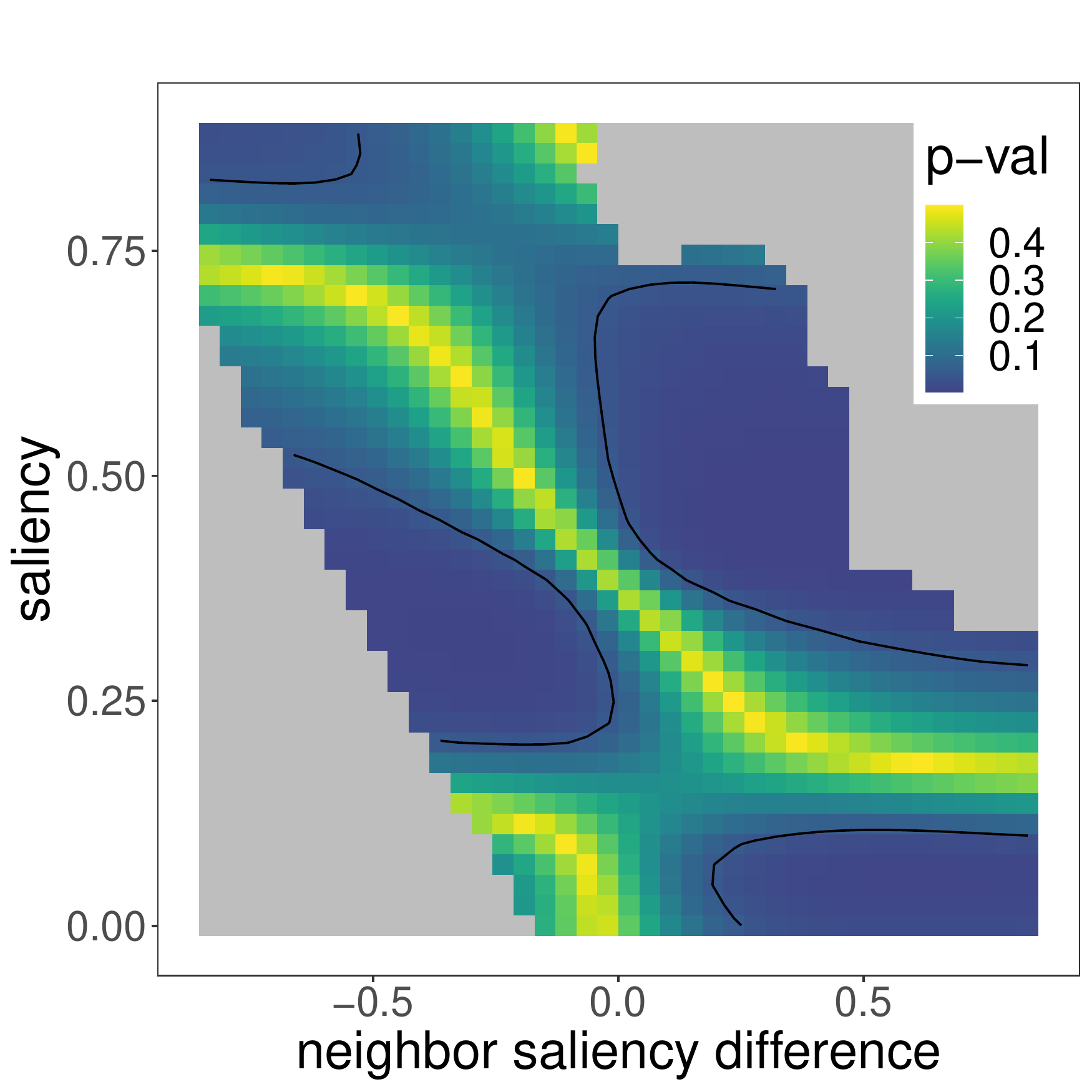}
        \caption{Right/left difference (chunk) $p$ values (contour marks 0.05).}\label{fig:diff_plot_chunk_p_shap}
    \end{subfigure}%
    \hfill
    \begin{subfigure}[t]{.19\textwidth}
        \centering
        \includegraphics[width=\textwidth]{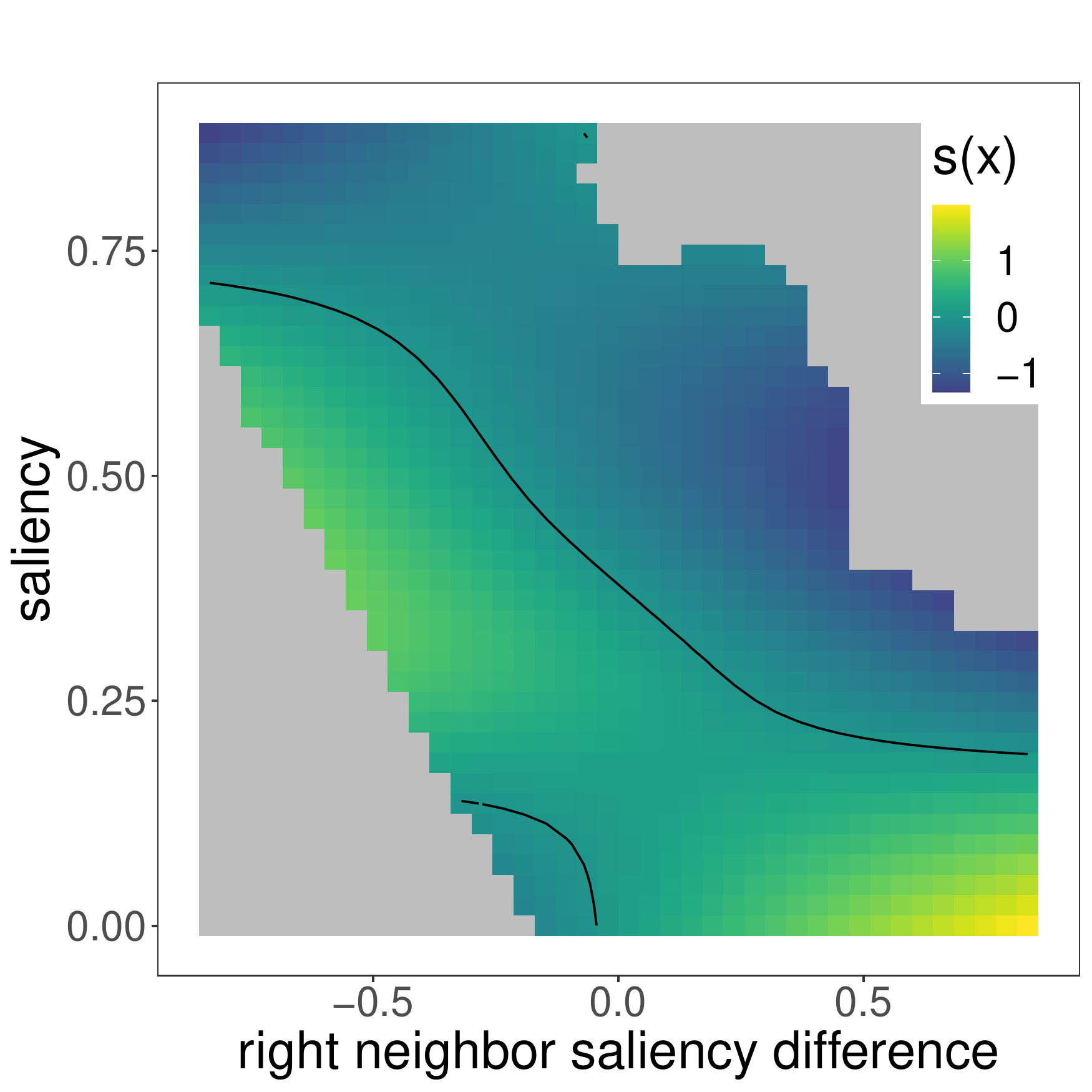}
        \caption{Chunk/no chunk difference for right neighbor (contour marks 0).}\label{fig:diff_plot_right_val_shap}
    \end{subfigure}%
    \hfill
    \begin{subfigure}[t]{.19\textwidth}
        \centering
        \includegraphics[width=\textwidth]{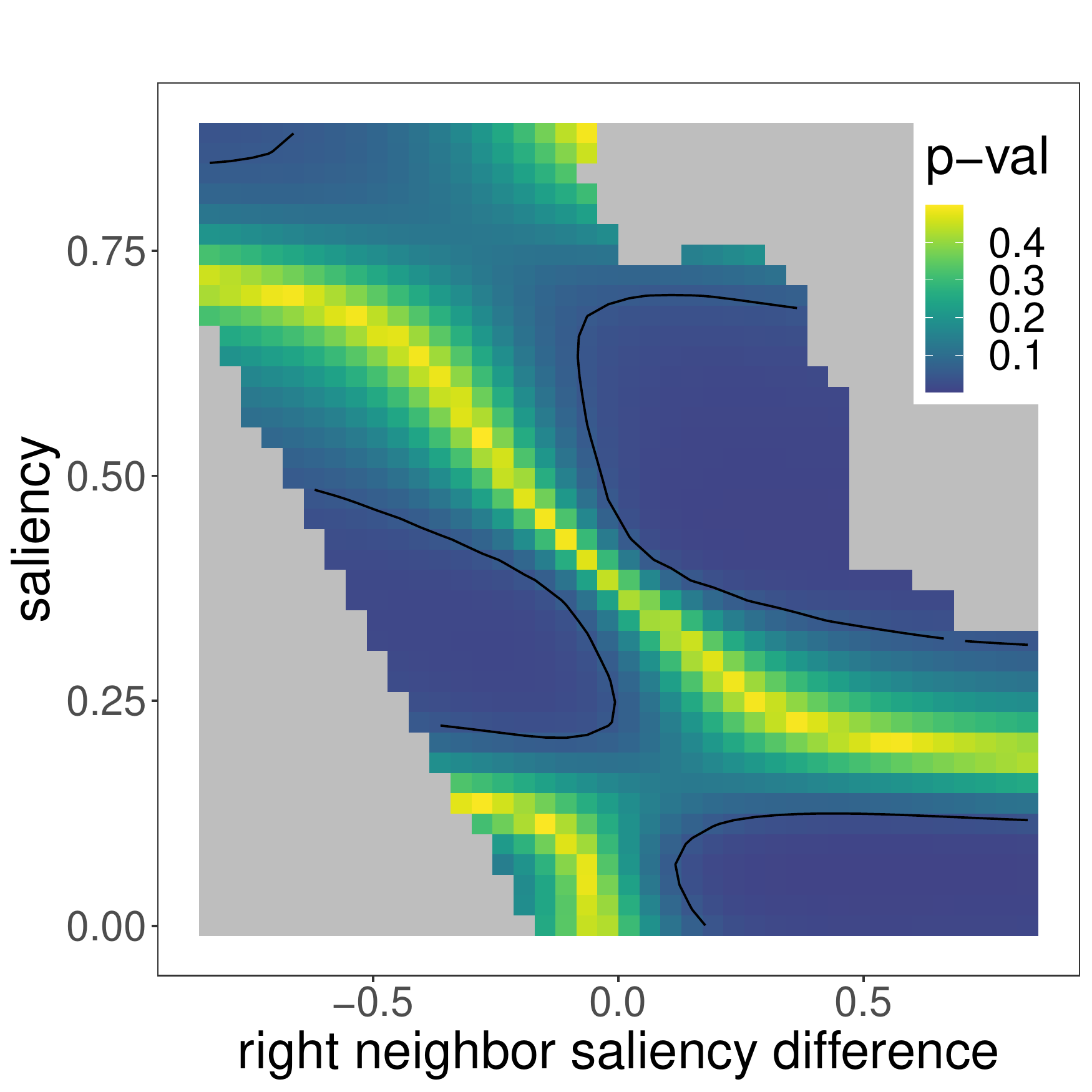}
        \caption{Chunk/no chunk difference (right) $p$ values (contour marks 0.05).}\label{fig:diff_plot_right_p_shap}
    \end{subfigure}%
    \hfill
    \begin{subfigure}[t]{.19\textwidth}
        \centering
        \includegraphics[width=\textwidth]{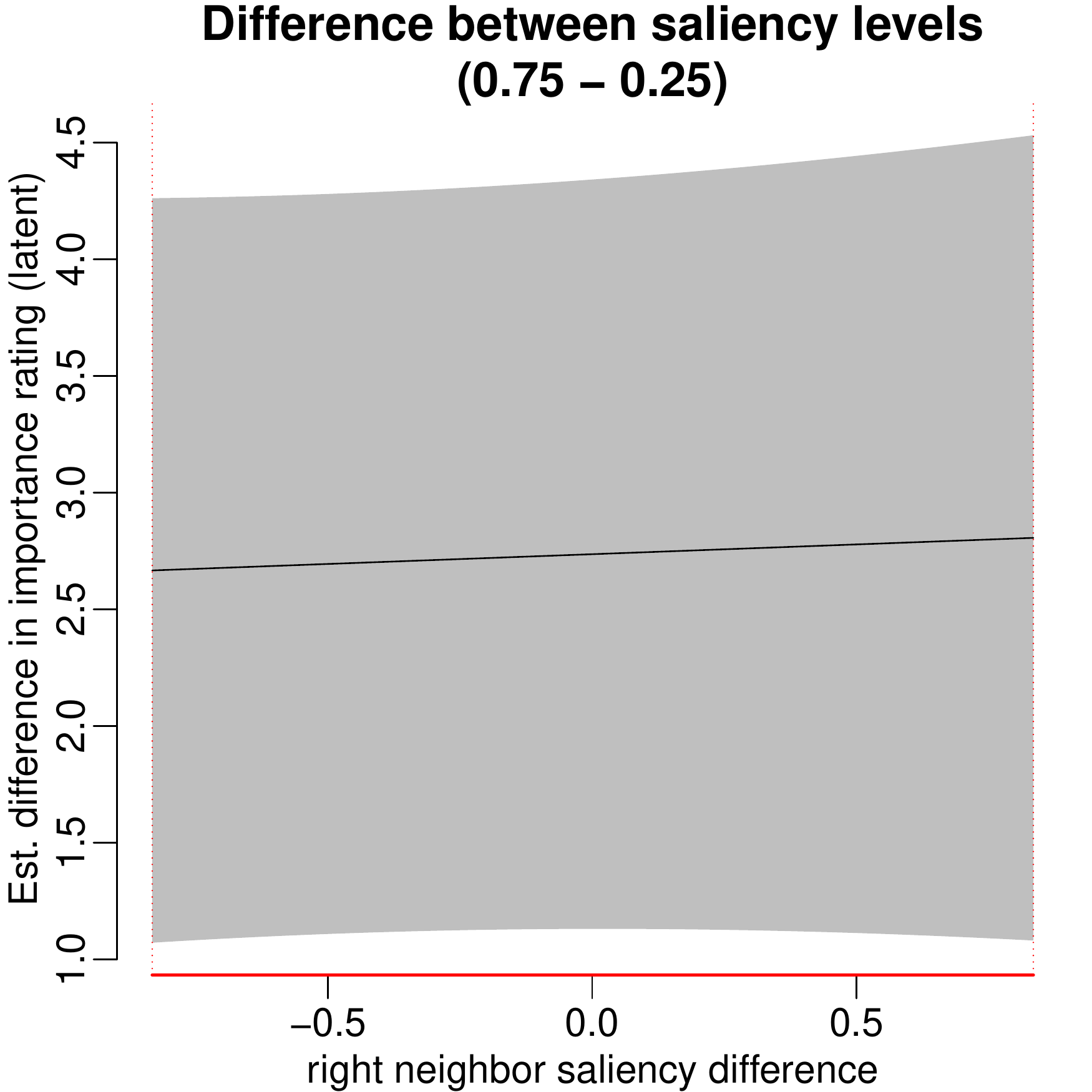}
        \caption{Difference between rated saliency and right neighbor saliency.
        }\label{fig:diff_saliency_levels_right_shap}
    \end{subfigure}%
    \caption{Difference plots of our SHAP-value experiment results. Contour refers to the contour line. Red x-axis in (e) marks significant differences.}\label{fig:difference_plots_main_paper_shap}
\end{figure*}

\begin{figure*}[h!]
    \centering
    \begin{subfigure}[t]{.25\textwidth}
        \centering
        \includegraphics[width=\textwidth]{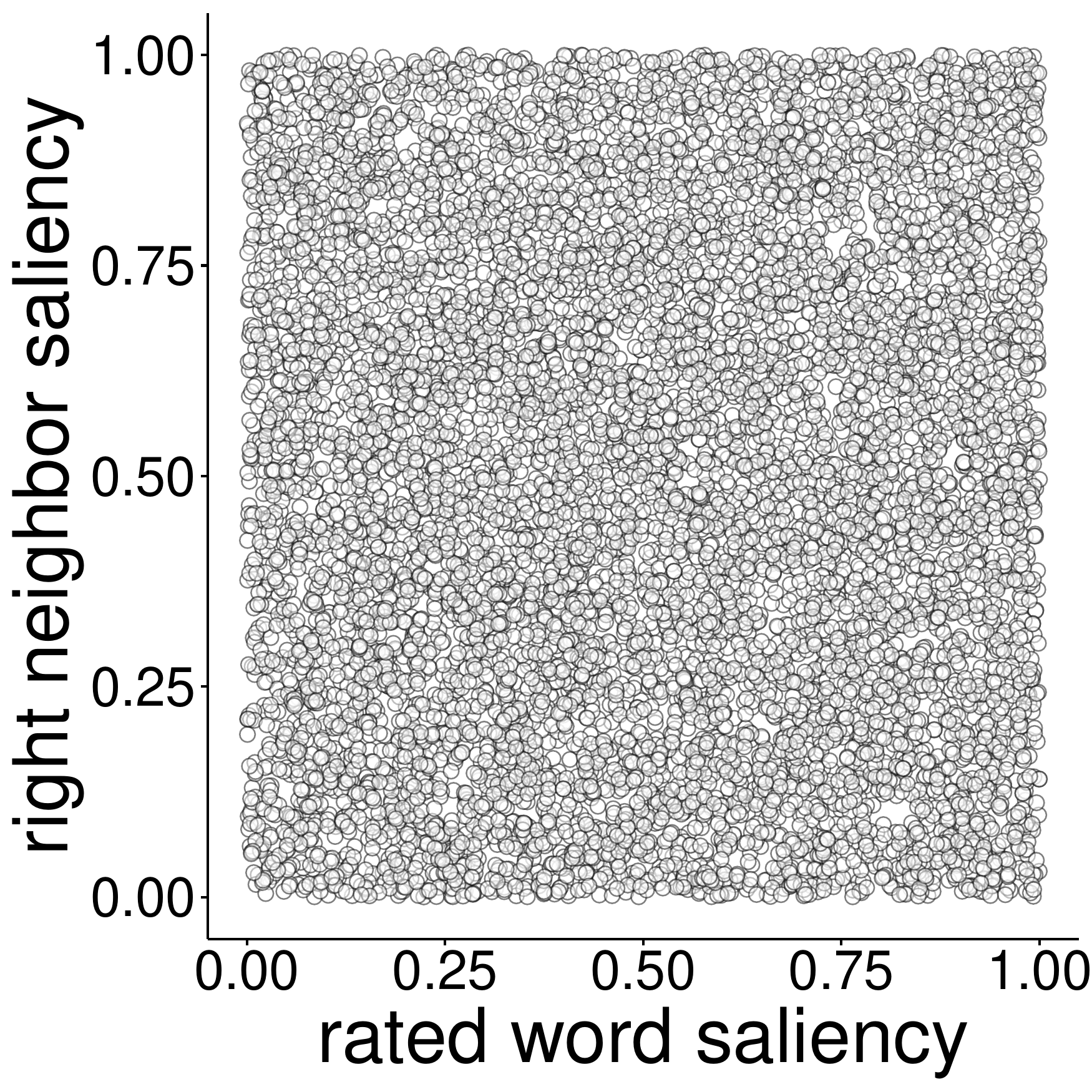}
        \caption{Randomized saliency.}\label{fig:middle_right_distribution_random_saliency}
    \end{subfigure}%
    \hspace{1.5cm}
    \begin{subfigure}[t]{.25\textwidth}
        \centering
        \includegraphics[width=\textwidth]{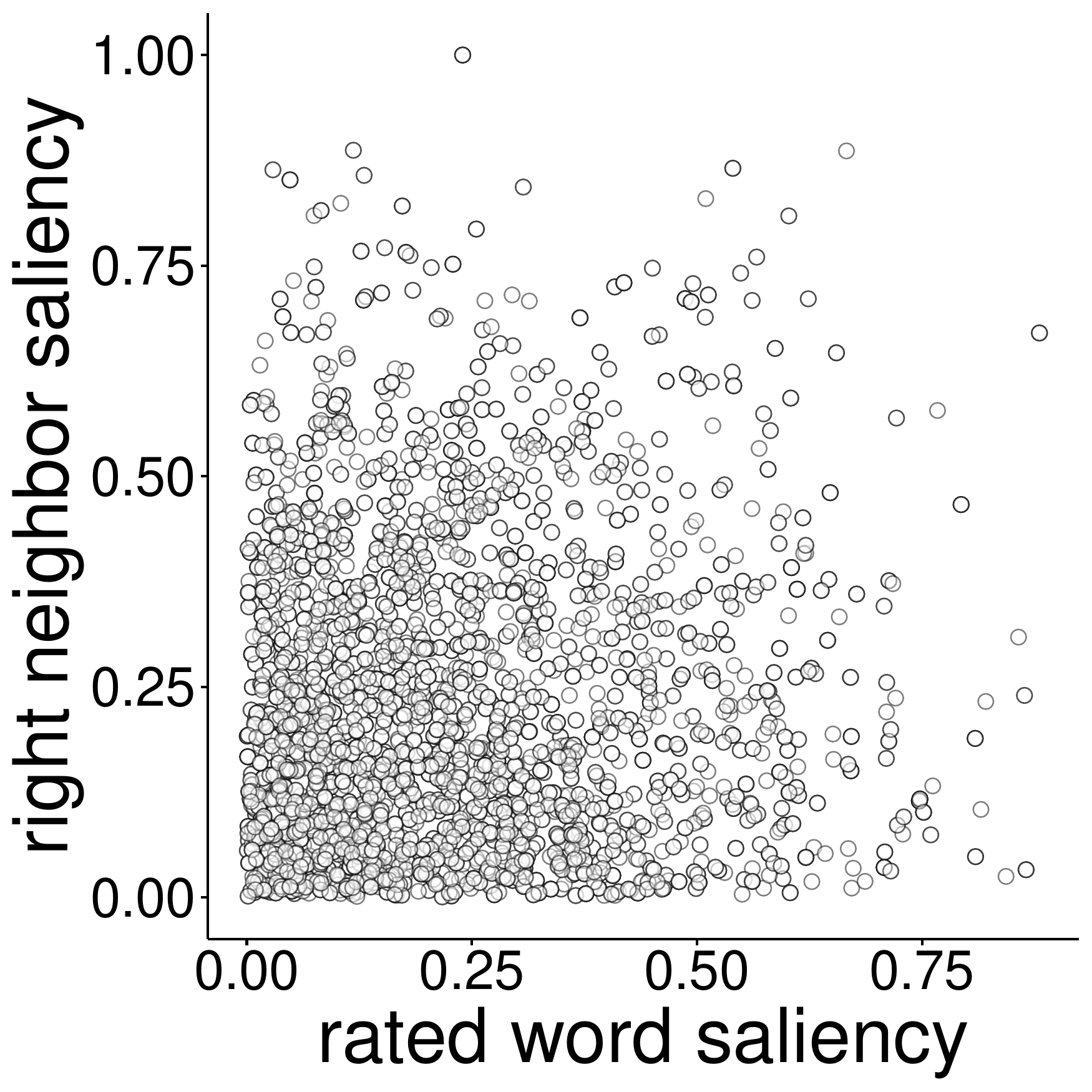}
        \caption{SHAP-value saliency.}\label{fig:middle_right_distribution_shap}
    \end{subfigure}%
    \caption{Comparison of the distributions of rated word saliency and right neighbor saliency across our randomized explanations (left) and our SHAP-value experiments (right).}\label{fig:distributions}
\end{figure*}

We additionally report details regarding our SHAP-value experiment results.
\Cref{fig:difference_plots_main_paper_shap} displays left/right, chunk/no chunk, and rated word saliency level difference plots.
We report the detailed Wald test statistics for our SHAP-value explanation experiment in \Cref{tab:stats_full_shap}.
\Cref{fig:distributions} illustrates how the distribution of saliency scores is uniforlmy random for our randomized explanations in contrast to the distributions of SHAP values.

\begin{figure*}[h!]
    \centering
    \begin{subfigure}[t]{.24\textwidth}
        \centering
    \includegraphics[width=\textwidth]{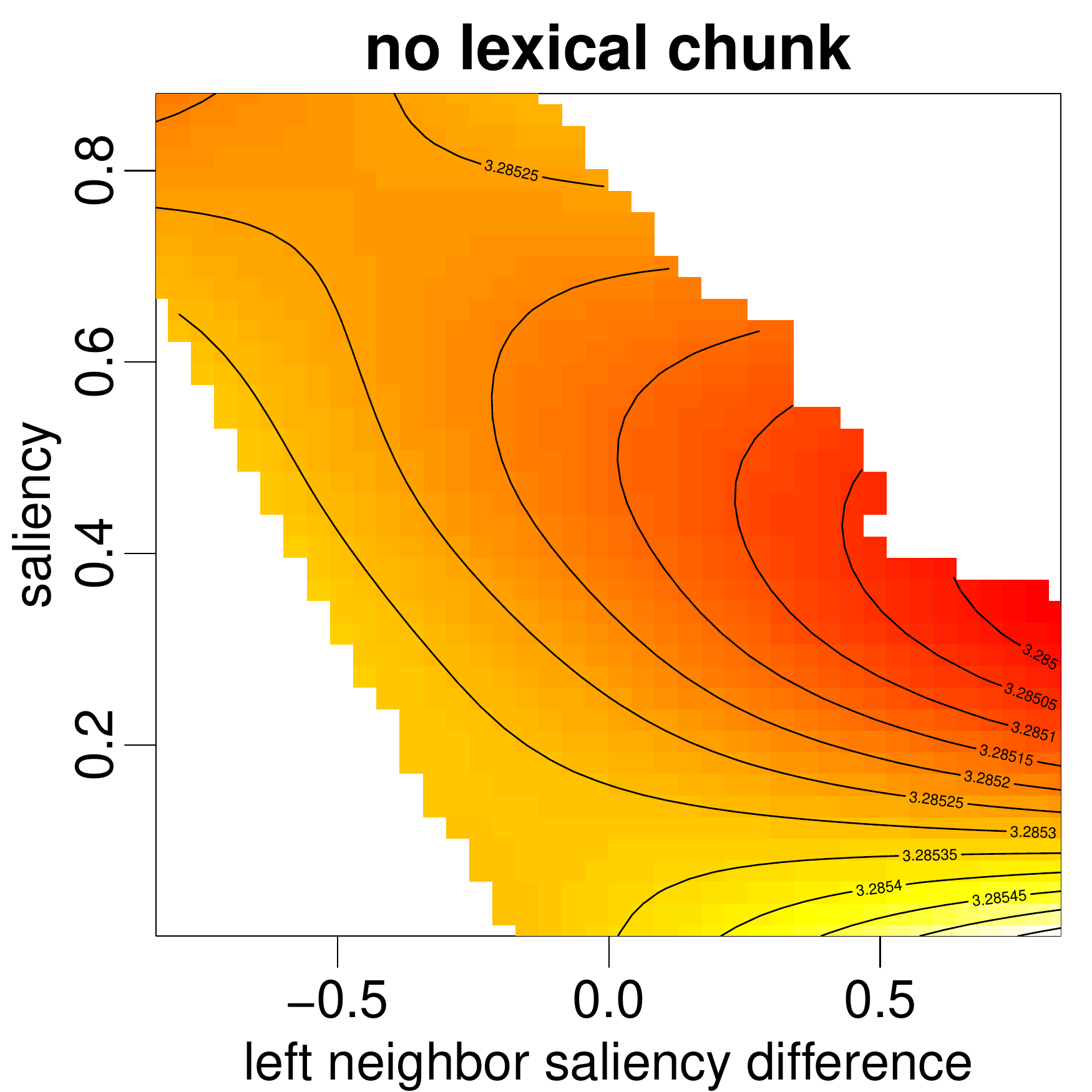}
    \caption{Left, no chunk.}\label{fig:left_neighbor_saliency_difference_saliency_no_lexical_chunk_shap}
    \end{subfigure}%
    \hfill
    \begin{subfigure}[t]{.24\textwidth}
        \centering
    \includegraphics[width=\textwidth]{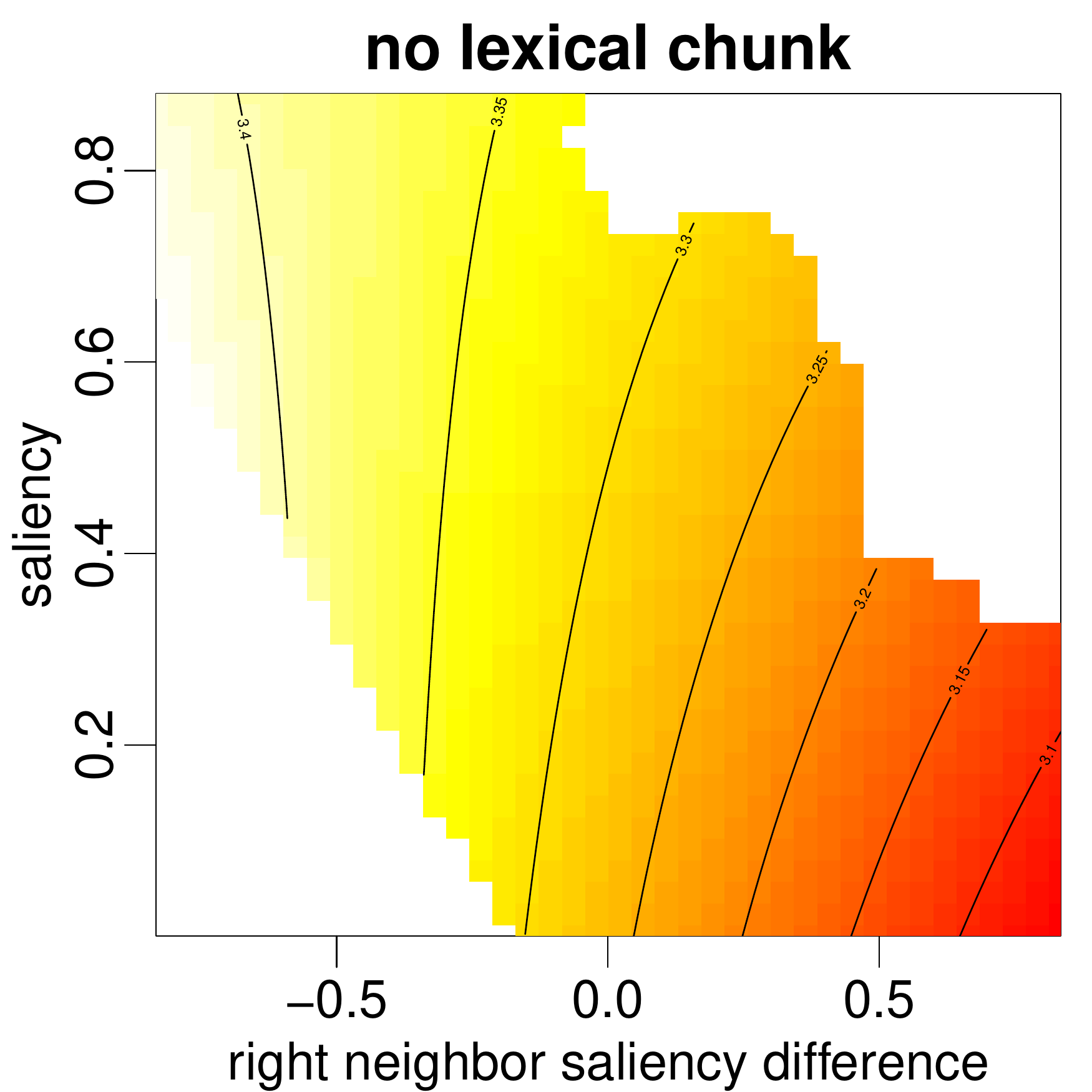}
    \caption{Right, no chunk.}\label{fig:right_neighbor_saliency_difference_saliency_no_lexical_chunk_shap}
    \end{subfigure}%
        \hfill
    \begin{subfigure}[t]{.24\textwidth}
        \centering
    \includegraphics[width=\textwidth]{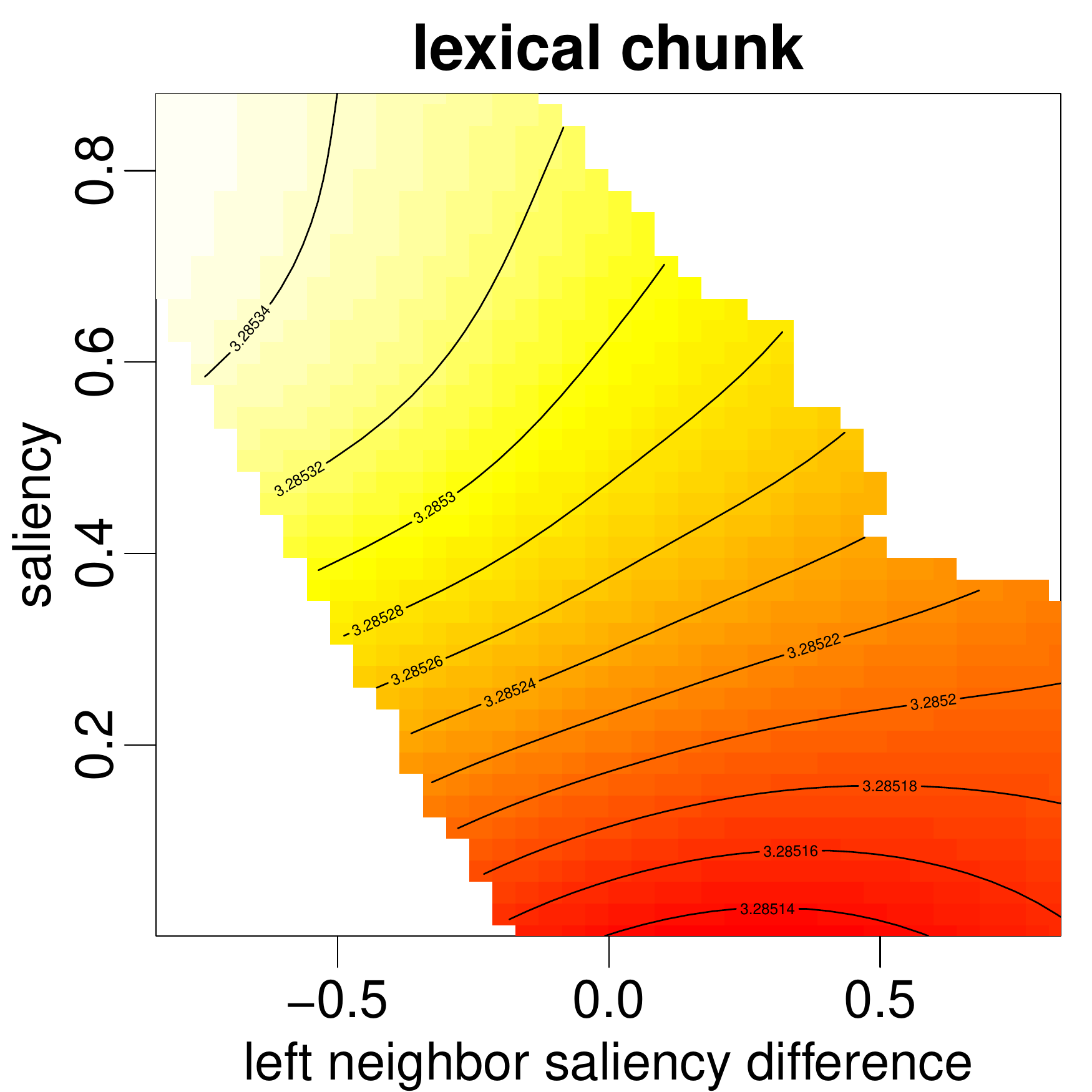}
    \caption{Left, chunk.}\label{fig:left_neighbor_saliency_difference_saliency_lexical_chunk_shap}
    \end{subfigure}%
    \hfill
    \begin{subfigure}[t]{.24\textwidth}
        \centering
    \includegraphics[width=\textwidth]{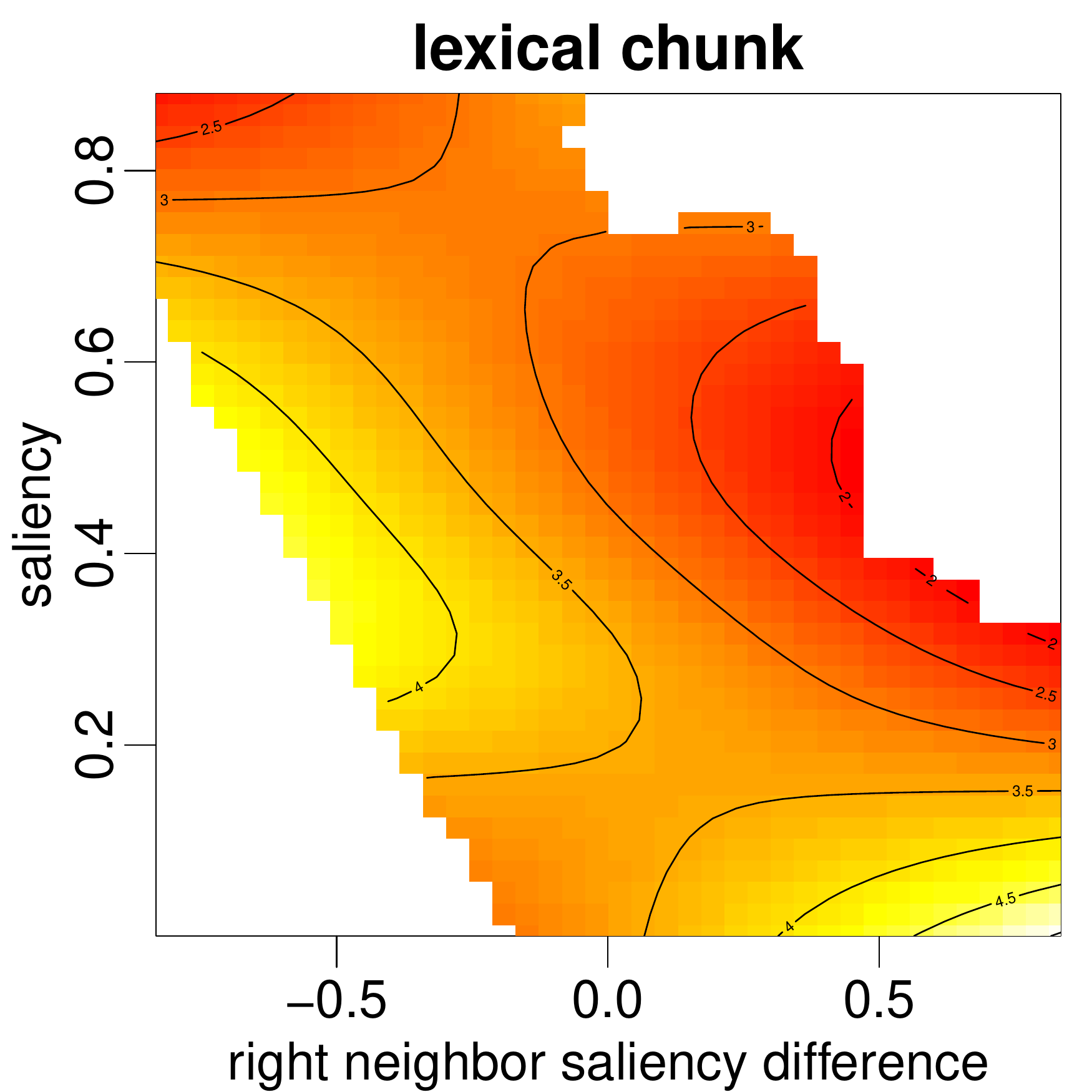}
    \caption{Right, chunk. ($\ast$)}\label{fig:right_neighbor_saliency_difference_saliency_lexical_chunk_shap}
    \end{subfigure}
    \caption{Left and right neighbours in our SHAP-value experiment. ($\ast$) marks statistically significant smooths. Colors are normalized per figure. Note that the first three plots correspond to non-significant effects and their respective color mappings covers a small value range.}\label{fig:left_and_right_neighbours_shap}
\end{figure*}

\subsection{Reproduction of \citet{DBLP:conf/fat/SchuffJAGV22}}
\label{app:reproduction-details}
We confirm previous results from \citet{DBLP:conf/fat/SchuffJAGV22} and find significant effects of \textbf{word length}, \textbf{display index}, \textbf{capitalization}, and \textbf{dependency relation}.
We report detailed statistics of our randomized saliency experiment in \Cref{tab:stats_full} and our SHAP experiment in  \Cref{tab:stats_full_shap}.

\begin{table*}[t]
\centering
\resizebox{0.65\linewidth}{!}{
\begin{tabular}{p{4.9cm}rrrr}
  \toprule
\textbf{Term} & \textbf{(e)df} & \textbf{Ref.df} & \textbf{F} & $p$ \\ 
  \midrule
s(saliency) & 11.22 & 19.00 & 580.89 & \textbf{$<$0.0001} \\ 
  s(display index) & 3.04 & 9.00 & 22.02 & \textbf{$<$0.0001} \\ 
  s(word length) & 1.64 & 9.00 & 16.44 & \textbf{$<$0.0001} \\ 
  s(sentence length) & 0.00 & 4.00 & 0.00 & 0.425 \\ 
  s(relative word frequency) & 0.00 & 9.00 & 0.00 & 0.844 \\ 
  s(normalized saliency rank) & 0.59 & 9.00 & 0.37 & 0.115 \\ 
  s(word position) & 0.58 & 9.00 & 0.18 & 0.177 \\ 
  te(left diff.,saliency): no chunk & 3.12 & 24.00 & 1.50 & \textbf{0.002} \\ 
  te(left diff.,saliency): chunk & 2.24 & 24.00 & 0.51 & \textbf{0.038} \\ 
  te(right diff.,saliency): no chunk & 2.43 & 24.00 & 0.47 & \textbf{0.049} \\ 
  te(right diff.,saliency): chunk & 0.00 & 24.00 & 0.00 & 0.578 \\ 
  \midrule
  s(sentence ID) & 0.00 & 149.00 & 0.00 & 0.616 \\ 
  s(saliency,sentence ID) & 16.13 & 150.00 & 0.14 & 0.191 \\ 
  s(worker ID) & 62.19 & 63.00 & 30911.89 & \textbf{$<$0.0001}\\ 
  s(saliency,worker ID) & 62.11 & 64.00 & 16760.88 & \textbf{$<$0.0001} \\ 
  \midrule
  capitalization &2.00  & & 3.15 & \textbf{0.042} \\ 
  dependency relation & 35.00  & & 2.92 & \textbf{$<$0.0001} \\
   \bottomrule
\end{tabular}}
\caption{Random saliency experiment results details. (Effective) degrees of freedom, reference degrees of freedom and Wald test statistics for the univariate smooth terms (top), random effects terms (middle) and parametric fixed terms (bottom) using $t=87.5\%$ and $\upvarphi^2$ measure.}
\label{tab:stats_full}
\end{table*}

\begin{table*}[t]
\centering
\resizebox{0.65\linewidth}{!}{
\begin{tabular}{p{4.9cm}rrrr}
  \toprule
\textbf{Term} & \textbf{(e)df} & \textbf{Ref.df} & \textbf{F} & $p$ \\ 
  \midrule
  s(saliency) & 6.71 & 19.00 & 18.85 & \textbf{$<$0.0001} \\ 
  s(display index) & 1.88 & 9.00 & 6.45 & \textbf{$<$0.0001} \\ 
  s(word length) & 2.04 & 9.00 & 4.43 & \textbf{$<$0.0001} \\ 
  s(sentence length) & 0.00 & 4.00 & 0.00 & 0.98 \\
  s(relative word frequency) & 0.00 & 9.00 & 0.00 & 0.64 \\ 
  s(normalized saliency rank) & 0.89 & 9.00 & 1.99 & \textbf{0.002} \\ 
  s(word position) & 0.42 & 9.00 & 0.12 & 0.19 \\ 
  te(left diff.,saliency): no chunk & 0.00 & 24.00 & 0.00 & 0.37 \\ 
  te(left diff.,saliency): chunk & 0.00 & 24.00 & 0.00 & 0.49 \\ 
  te(right diff.,saliency): no chunk & 0.99 & 24.00 & 0.20 & 0.06 \\ 
  te(right diff.,saliency): chunk & 3.24 & 24.00 & 1.09 & 0.01 \\ 
  \midrule
  s(sentence ID) & 0.00 & 149.00 & 0.00 & 0.52 \\ 
  s(saliency,sentence ID) & 11.31 & 150.00 & 0.10 & 0.14 \\ 
  s(worker ID) & 34.77 & 35.00 & 14185.28 & \textbf{$<$0.0001}\\ 
  s(saliency,worker ID) & 62.11 & 64.00 & 16760.88 & \textbf{$<$0.0001} \\ 
  \midrule
  capitalization & 2.00 & 0.35 & 0.71 \\ 
  dependency relation & 34.59 & 36.00 & 8468.22 & \textbf{$<$0.0001} \\ 
   \bottomrule
\end{tabular}}
\caption{SHAP experiment results details. (Effective) degrees of freedom, reference degrees of freedom and Wald test statistics for the univariate smooth terms (top), random effects terms (middle) and parametric fixed terms (bottom) using $t=87.5\%$ and $\upvarphi^2$ measure.}
\label{tab:stats_full_shap}
\end{table*}

\section{Robustness to Evaluation Parameters.}\label{app:robustness}
\begin{figure*}[h!]
    \centering
    \begin{subfigure}[t]{.25\textwidth}
        \centering
    \includegraphics[width=\textwidth]{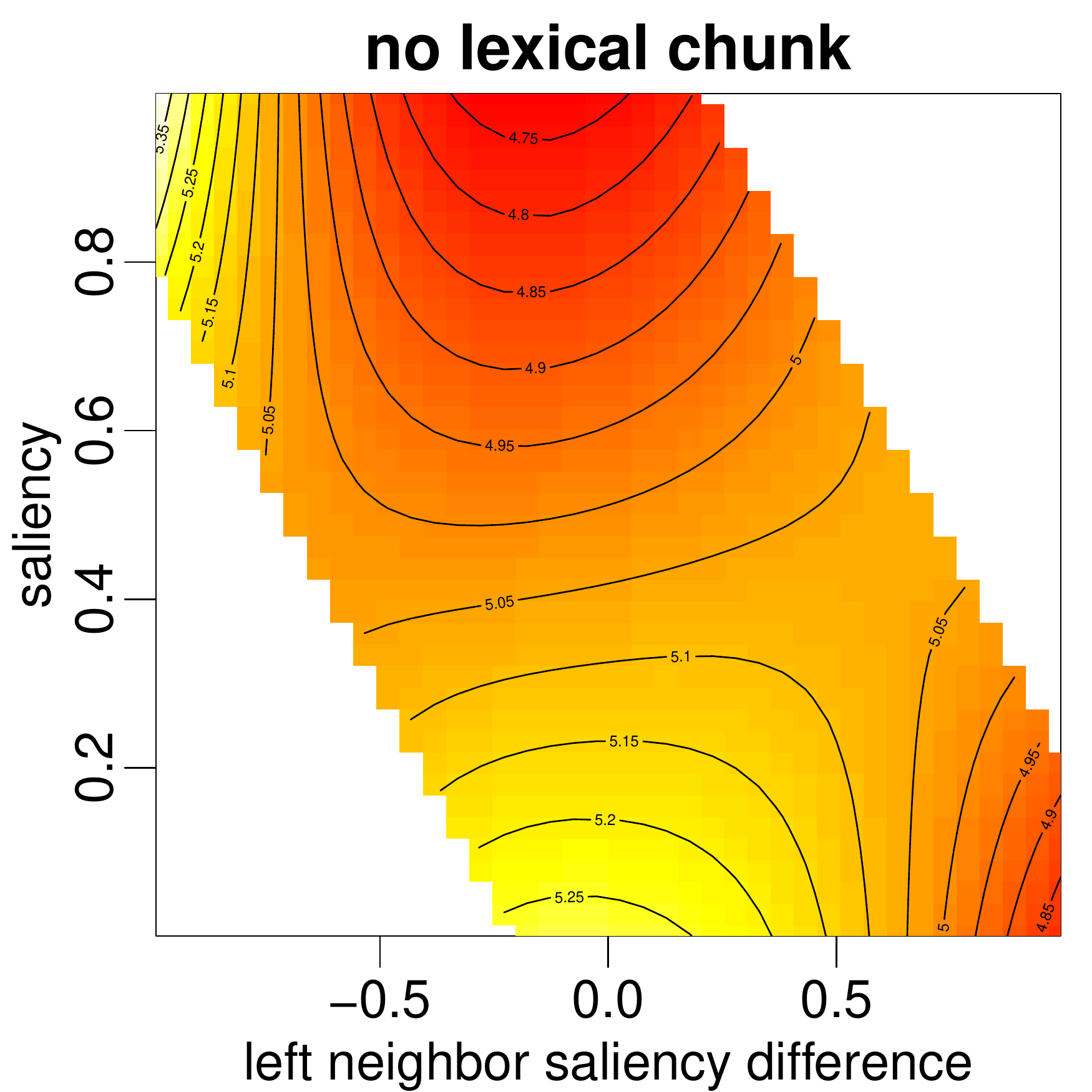}
    \caption{$\upvarphi^2$.}
    \end{subfigure}%
    \hfill
    \begin{subfigure}[t]{.25\textwidth}
        \centering
    \includegraphics[width=\textwidth]{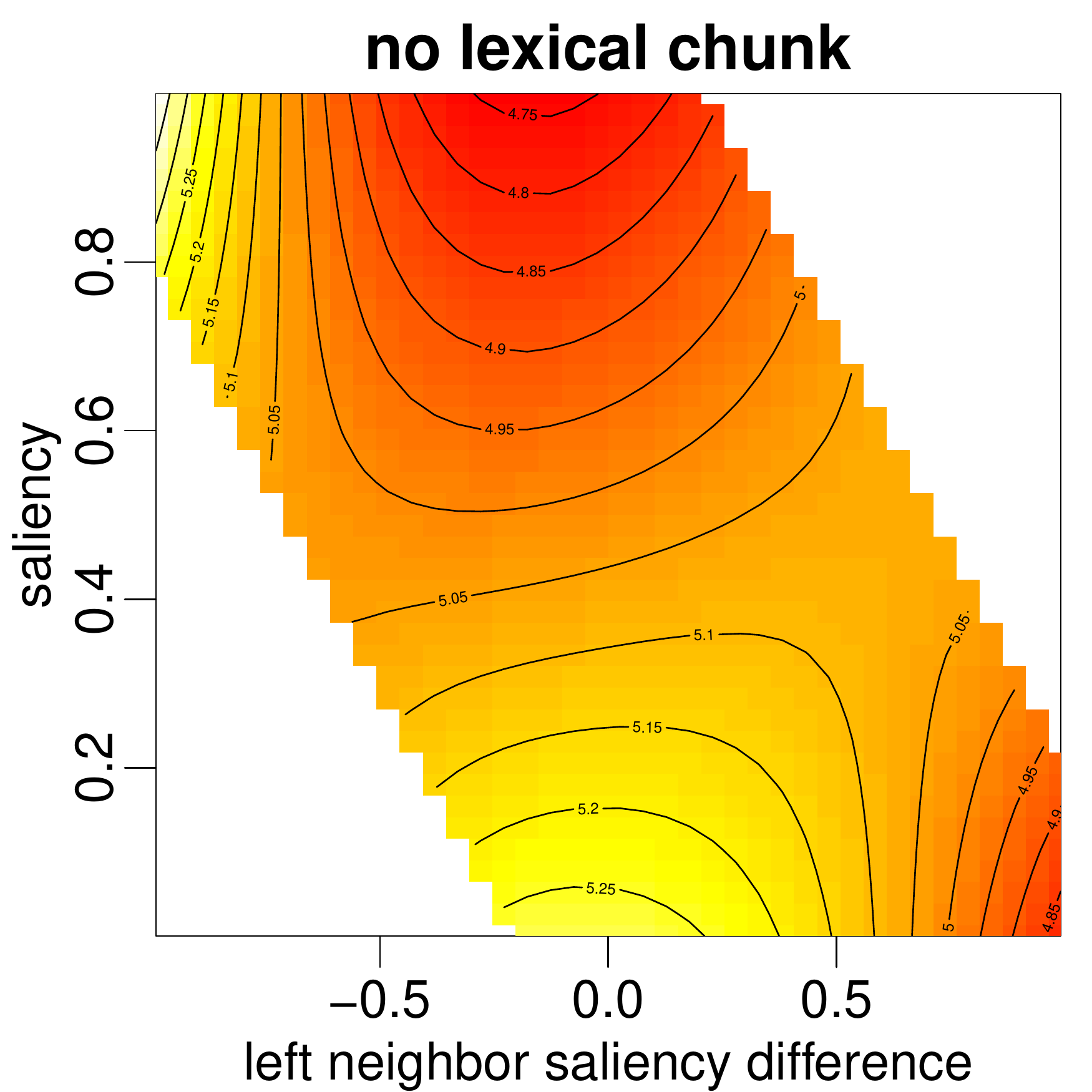}
    \caption{Jaccard.}
    \end{subfigure}%
    \begin{subfigure}[t]{.25\textwidth}
        \centering
    \includegraphics[width=\textwidth]{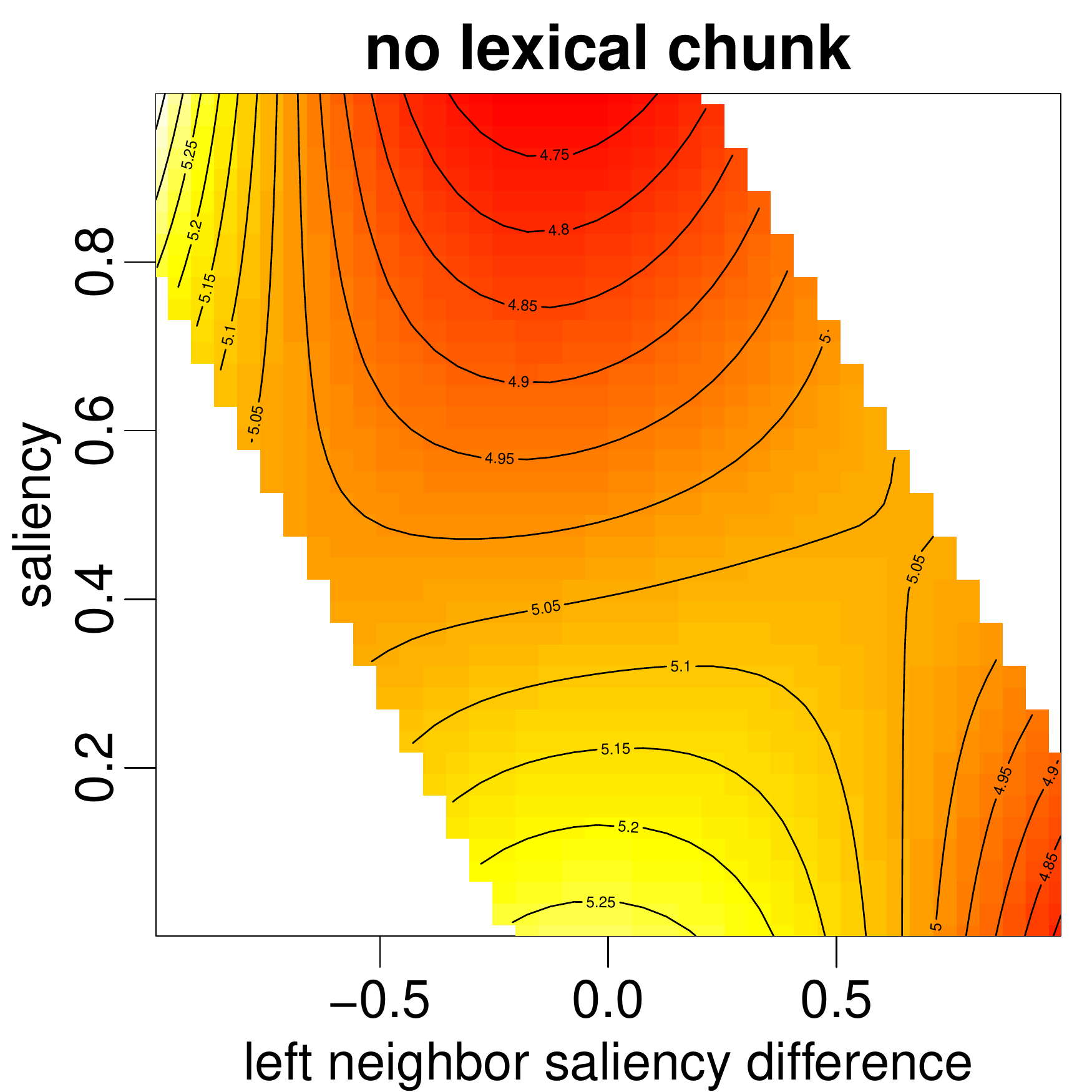}
    \caption{MI-like.}
    \end{subfigure}%
    \begin{subfigure}[t]{.25\textwidth}
        \centering
    \includegraphics[width=\textwidth]{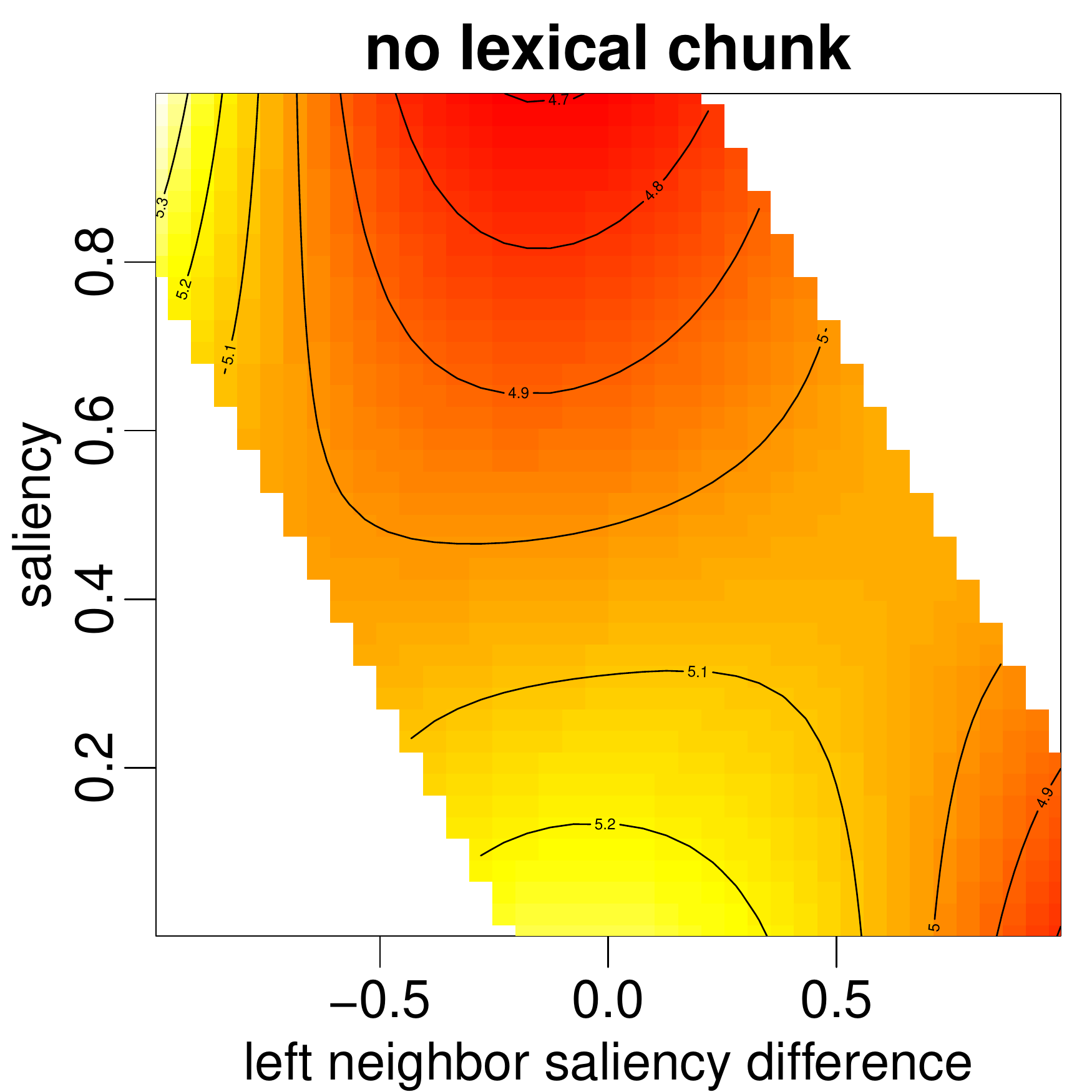}
    \caption{Poisson-Stirling.}
    \end{subfigure}
    \caption{Tensor product interactions for left saliency difference in the outside chunk setting across different choices of cooccurrence measures for our randomized explanation experiment. We find similar patterns across all settings. $t=87.5$ is consistent for all plots.}\label{fig:eval_robustness_measure_left_no_chunk}
\end{figure*}

\begin{figure*}[h!]
    \centering
    \begin{subfigure}[t]{.25\textwidth}
        \centering
    \includegraphics[width=\textwidth]{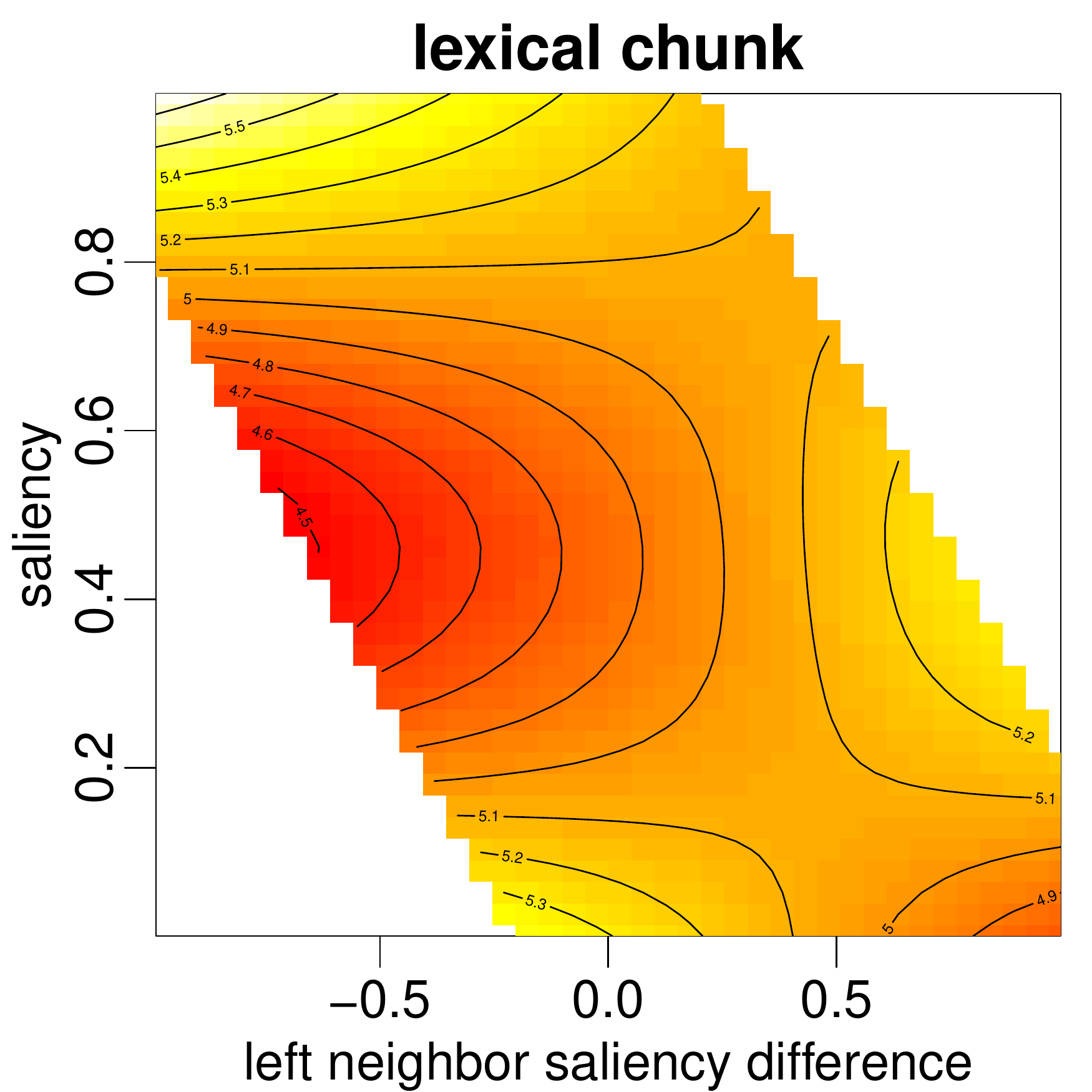}
    \caption{$\upvarphi^2$.}
    \end{subfigure}%
    \hfill
    \begin{subfigure}[t]{.25\textwidth}
        \centering
    \includegraphics[width=\textwidth]{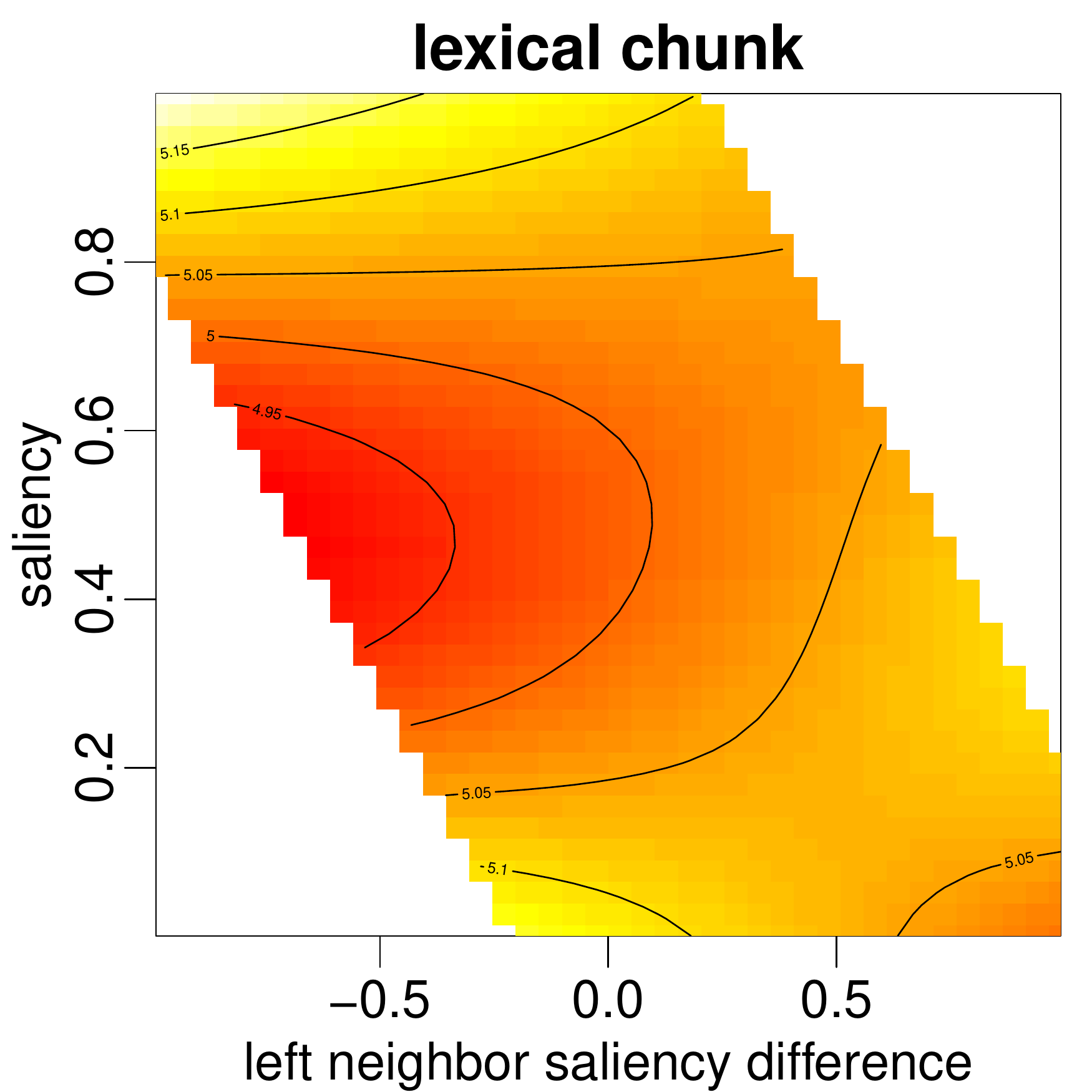}
    \caption{Jaccard.}
    \end{subfigure}%
    \begin{subfigure}[t]{.25\textwidth}
        \centering
    \includegraphics[width=\textwidth]{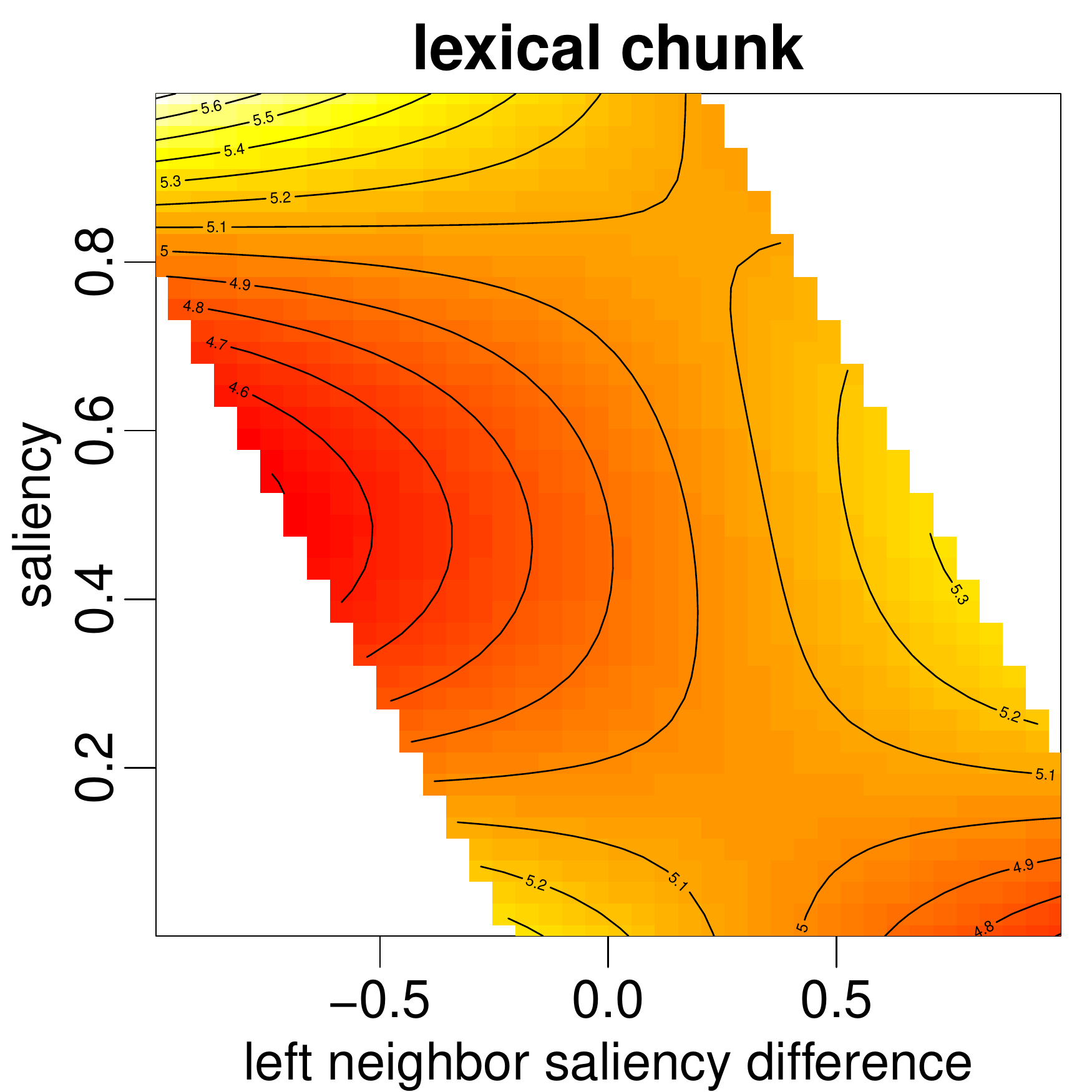}
    \caption{MI-like.}
    \end{subfigure}%
    \begin{subfigure}[t]{.25\textwidth}
        \centering
    \includegraphics[width=\textwidth]{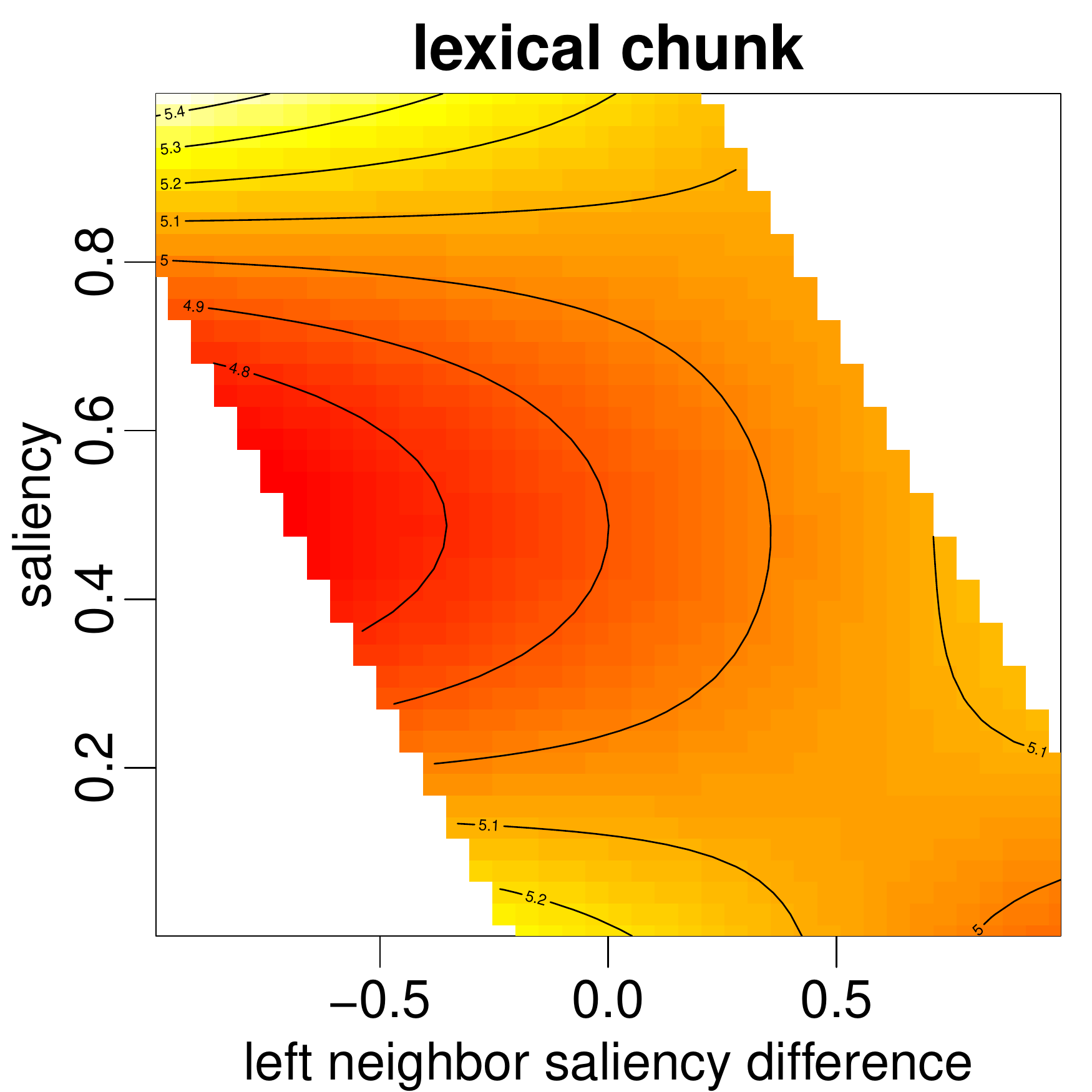}
    \caption{Poisson-Stirling.}
    \end{subfigure}
    \caption{Tensor product interactions for left saliency difference in the within chunk setting across different choices of cooccurrence measures for our randomized explanation experiment. We find similar patterns across all settings. $t=87.5$ is consistent for all plots.}\label{fig:eval_robustness_measure_left_chunk}
\end{figure*}

\begin{figure*}[h!]
    \centering
    \begin{subfigure}[t]{.25\textwidth}
        \centering
    \includegraphics[width=\textwidth]{figures/phi_sq_875/diff_plot_no_chunk_p.pdf}
    \caption{$\upvarphi^2$.}
    \end{subfigure}%
    \hfill
    \begin{subfigure}[t]{.25\textwidth}
        \centering
    \includegraphics[width=\textwidth]{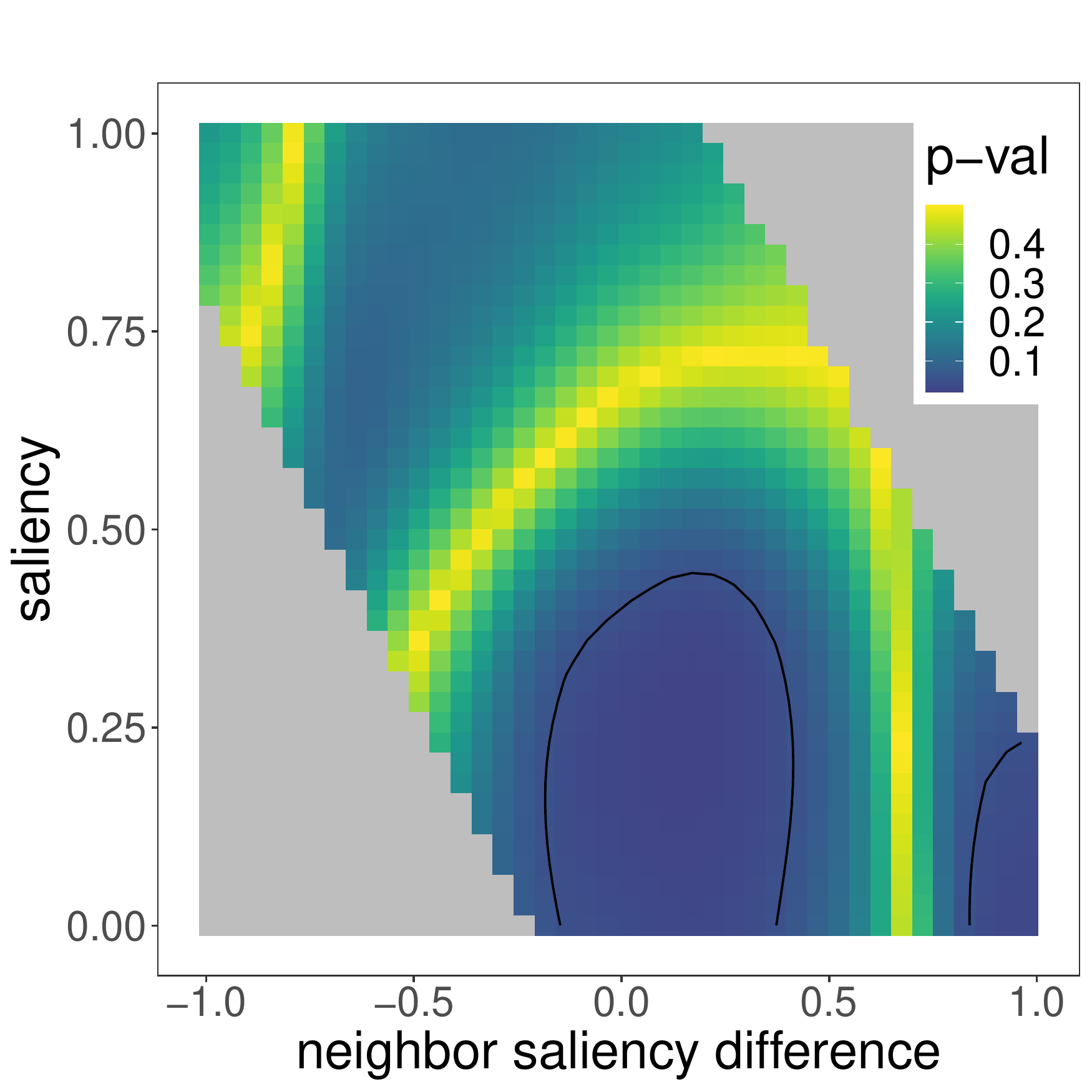}
    \caption{Jaccard.}
    \end{subfigure}%
    \begin{subfigure}[t]{.25\textwidth}
        \centering
    \includegraphics[width=\textwidth]{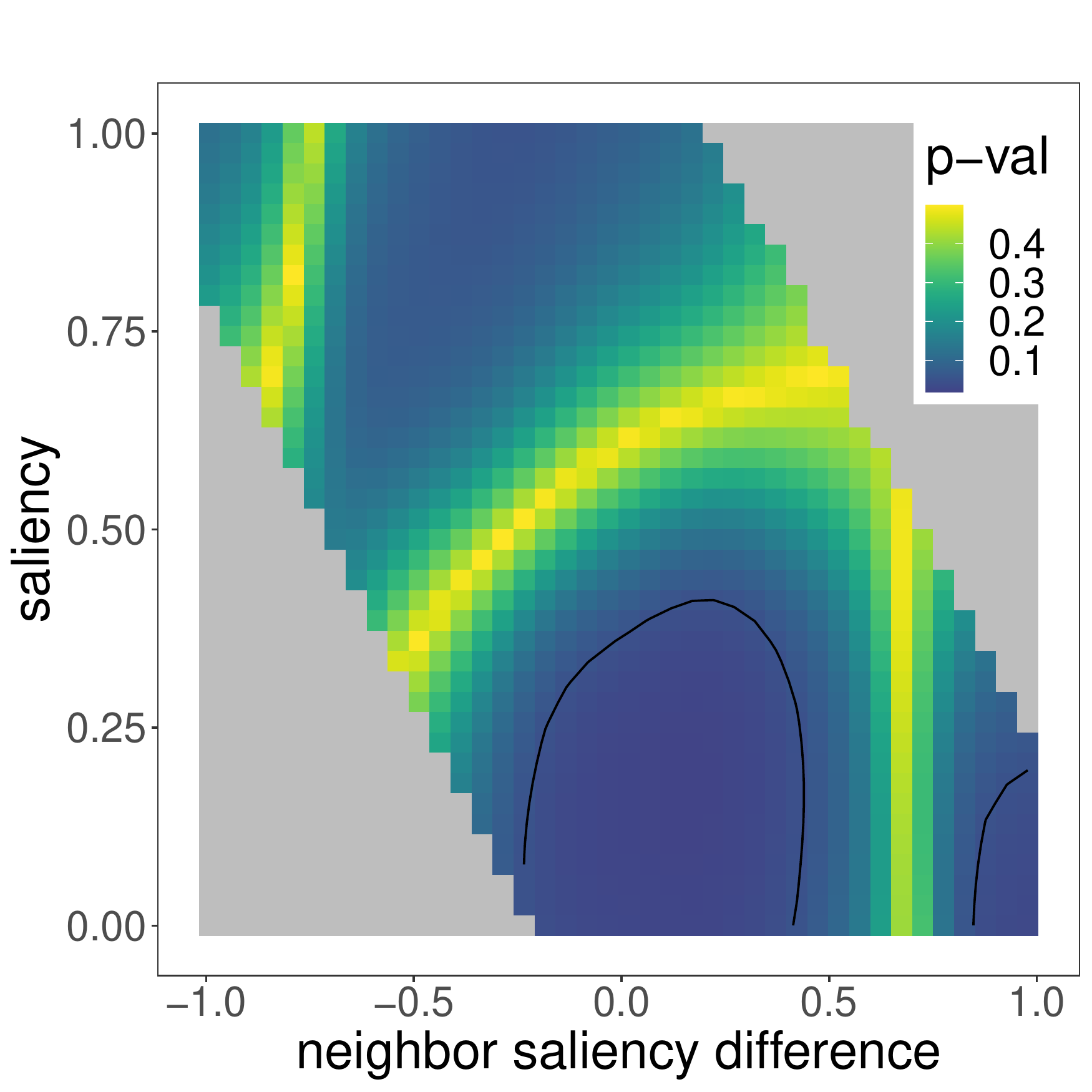}
    \caption{MI-like.}
    \end{subfigure}%
    \begin{subfigure}[t]{.25\textwidth}
        \centering
    \includegraphics[width=\textwidth]{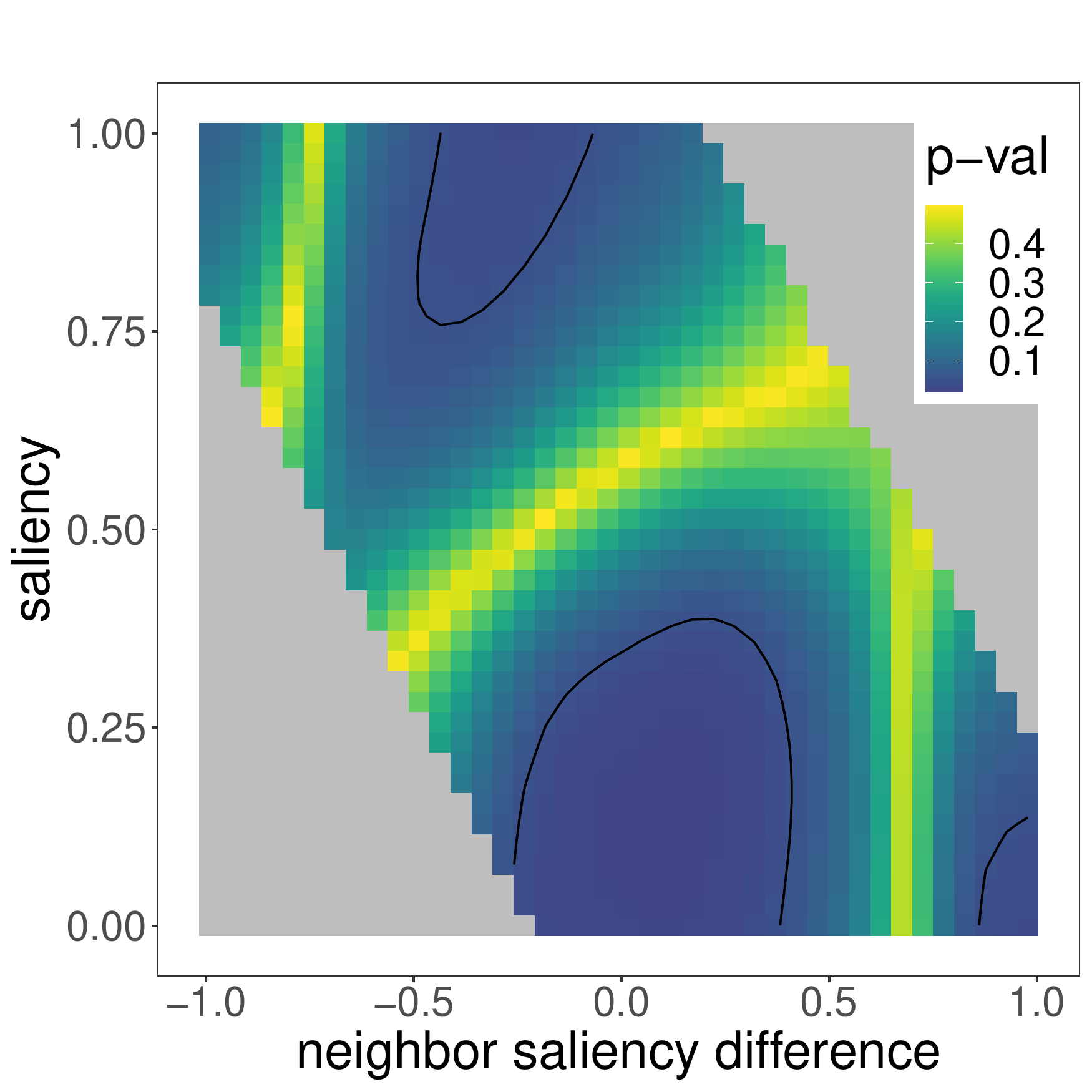}
    \caption{Poisson-Stirling.}
    \end{subfigure}
    \caption{$p$ values for between right - left for no lexical chunk neighbors across different choices of cooccurrence measures for our randomized explanation experiment. We find similar patterns across all settings. $t=87.5$ is consistent for all plots.}\label{fig:eval_robustness_measure_chunk_diff_p}
\end{figure*}

\begin{figure*}[h!]
    \centering
    \begin{subfigure}[t]{.25\textwidth}
        \centering
    \includegraphics[width=\textwidth]{figures/phi_sq_875/diff_plot_no_chunk_p.pdf}
    \caption{$t=87.5\%$}
    \end{subfigure}%
    \hfill
    \begin{subfigure}[t]{.25\textwidth}
        \centering
    \includegraphics[width=\textwidth]{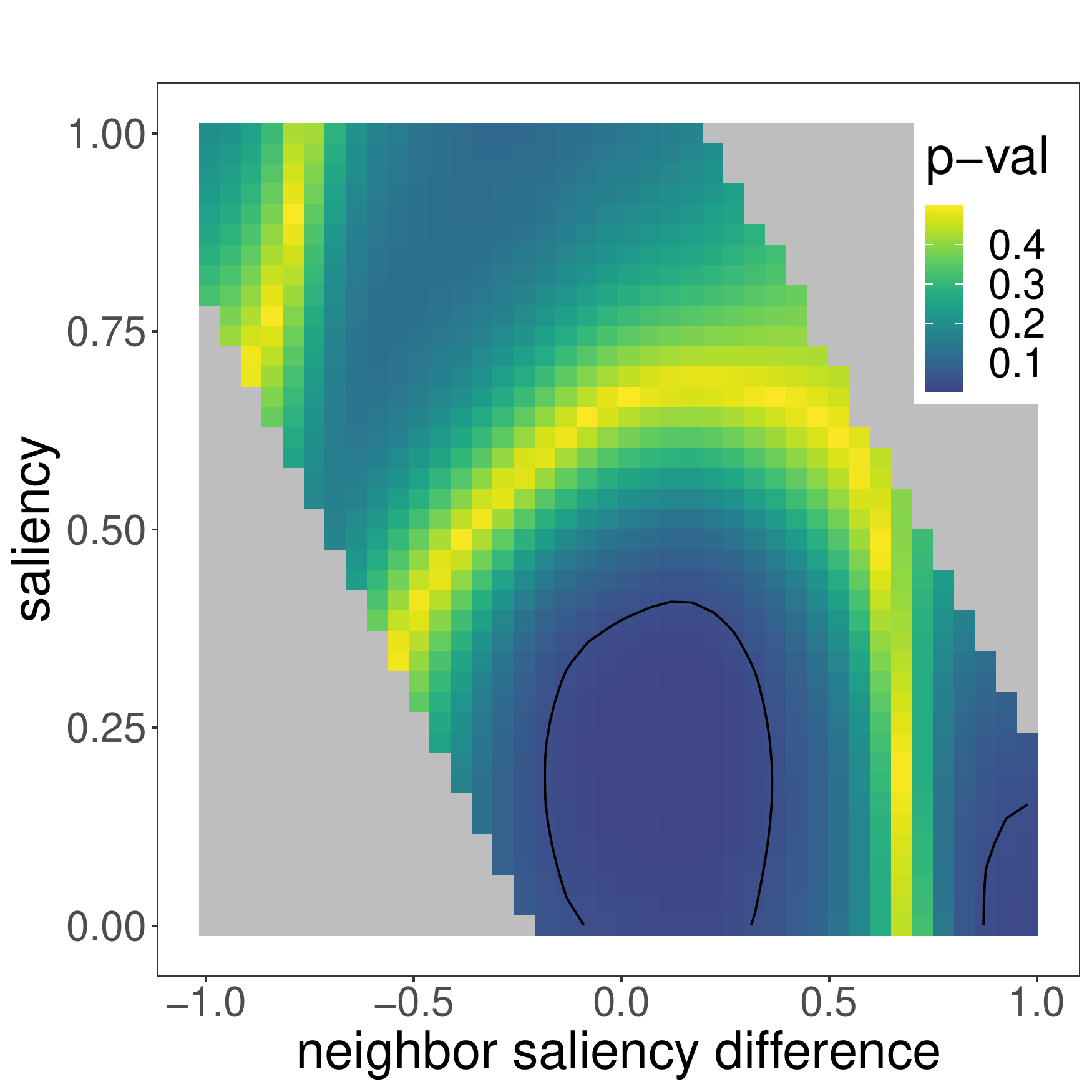}
    \caption{$t=75\%$}
    \end{subfigure}%
    \begin{subfigure}[t]{.25\textwidth}
        \centering
    \includegraphics[width=\textwidth]{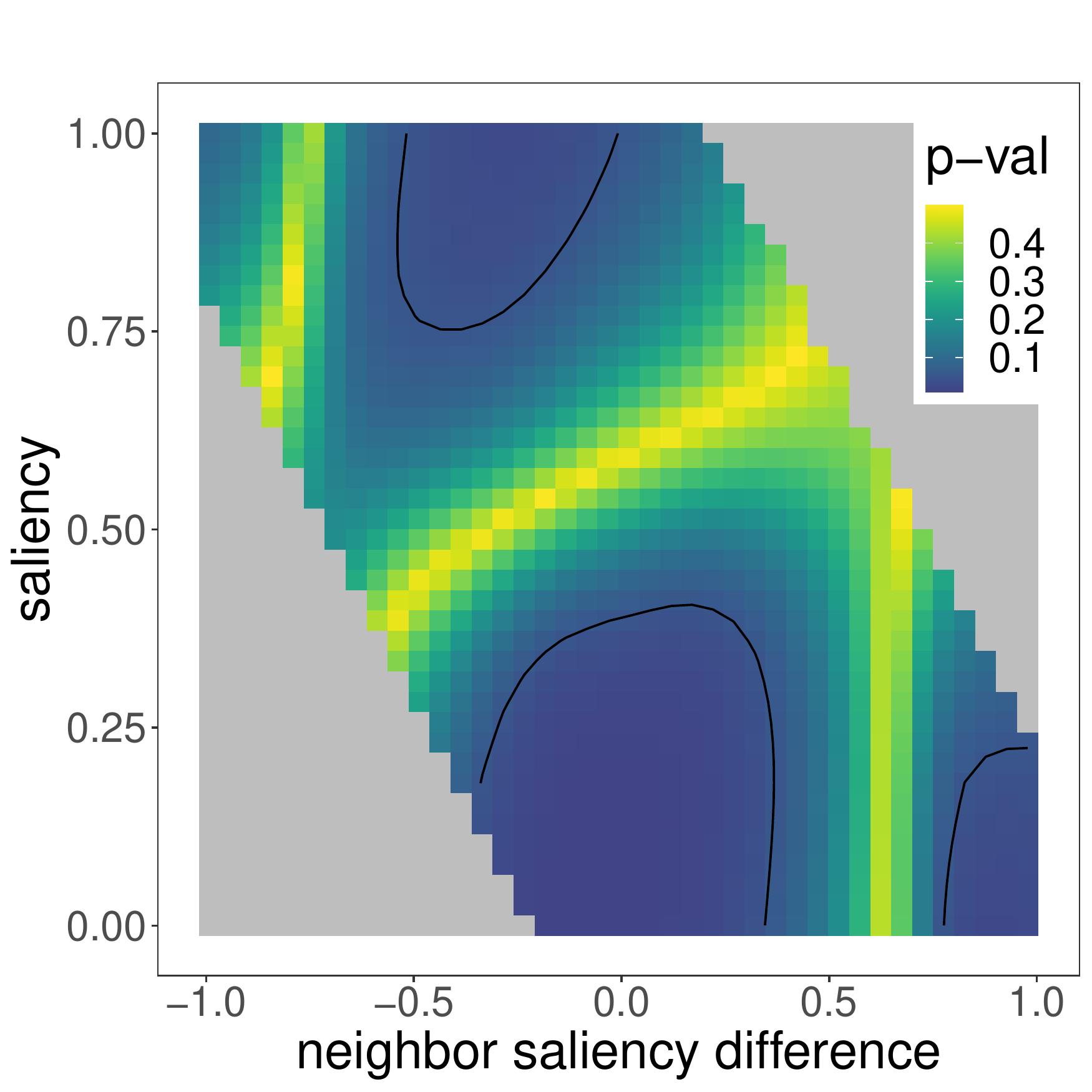}
    \caption{$t=50\%$}
    \end{subfigure}%
    \begin{subfigure}[t]{.25\textwidth}
        \centering
    \includegraphics[width=\textwidth]{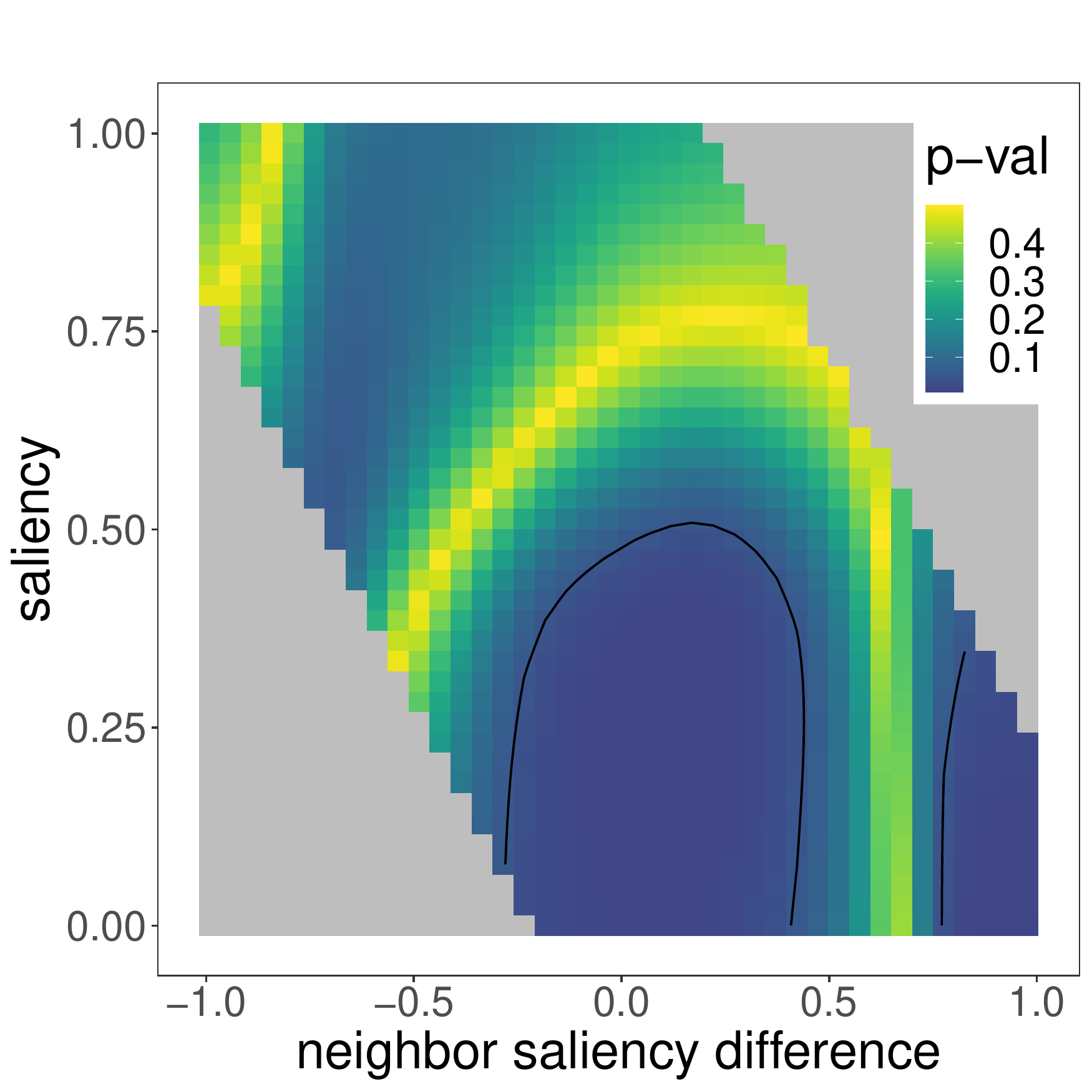}
    \caption{$t=25\%$}
    \end{subfigure}
    \caption{$p$ values for differences between right - left for no lexical chunk neighbors across different choices of thresholds for our randomized explanation experiment. We find similar patterns across all settings. The $\upvarphi^2$ measure is used across all plots.}\label{fig:eval_robustness_threshold_no_chunk_diff_p}
\end{figure*}

\begin{figure*}[h!]
    \centering
    \begin{subfigure}[t]{.25\textwidth}
        \centering
    \includegraphics[width=\textwidth]{figures/phi_sq_875/diff_saliency_levels_left.pdf}
    \caption{$t=87.5\%$}
    \end{subfigure}%
    \hfill
    \begin{subfigure}[t]{.25\textwidth}
        \centering
    \includegraphics[width=\textwidth]{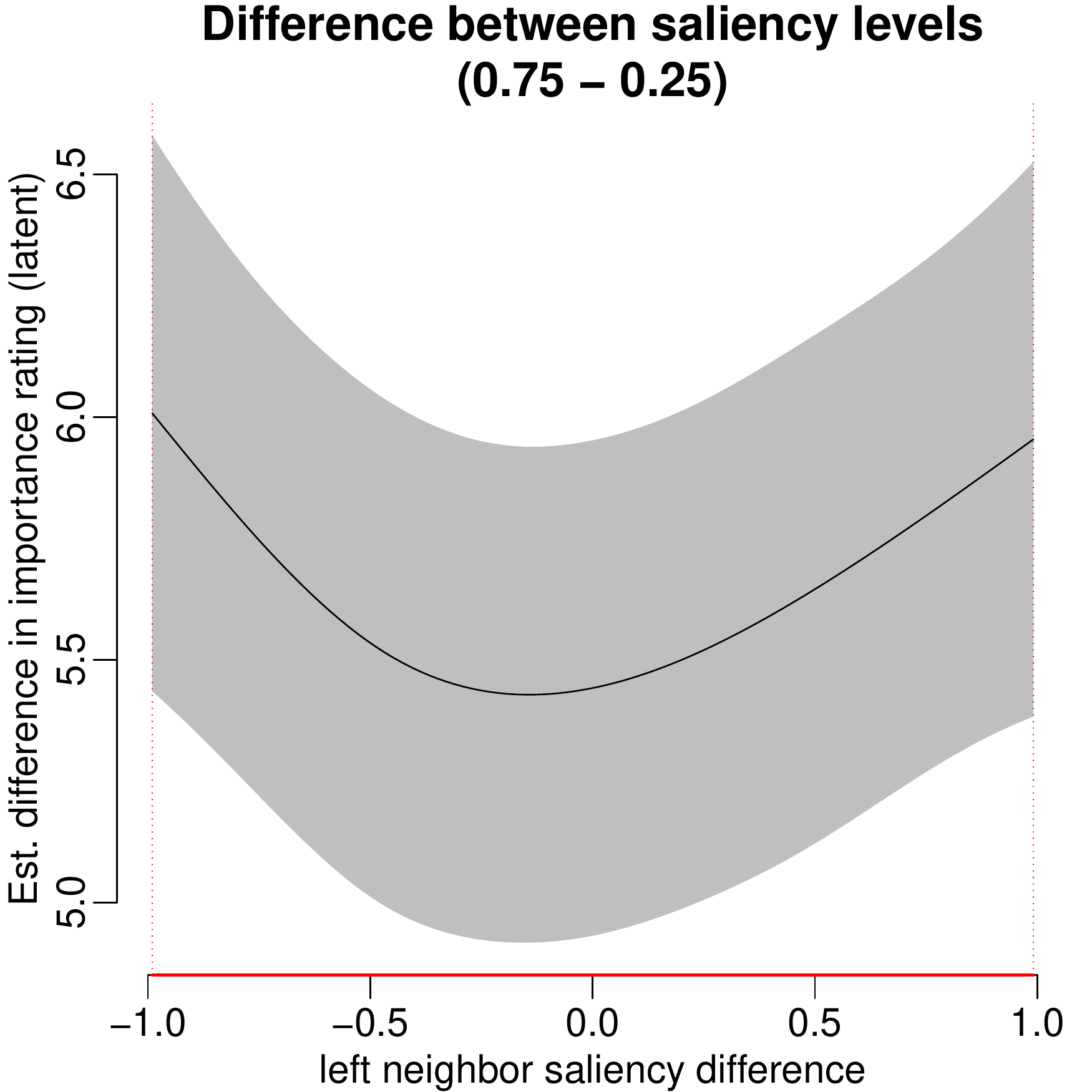}
    \caption{$t=75\%$}
    \end{subfigure}%
    \begin{subfigure}[t]{.25\textwidth}
        \centering
    \includegraphics[width=\textwidth]{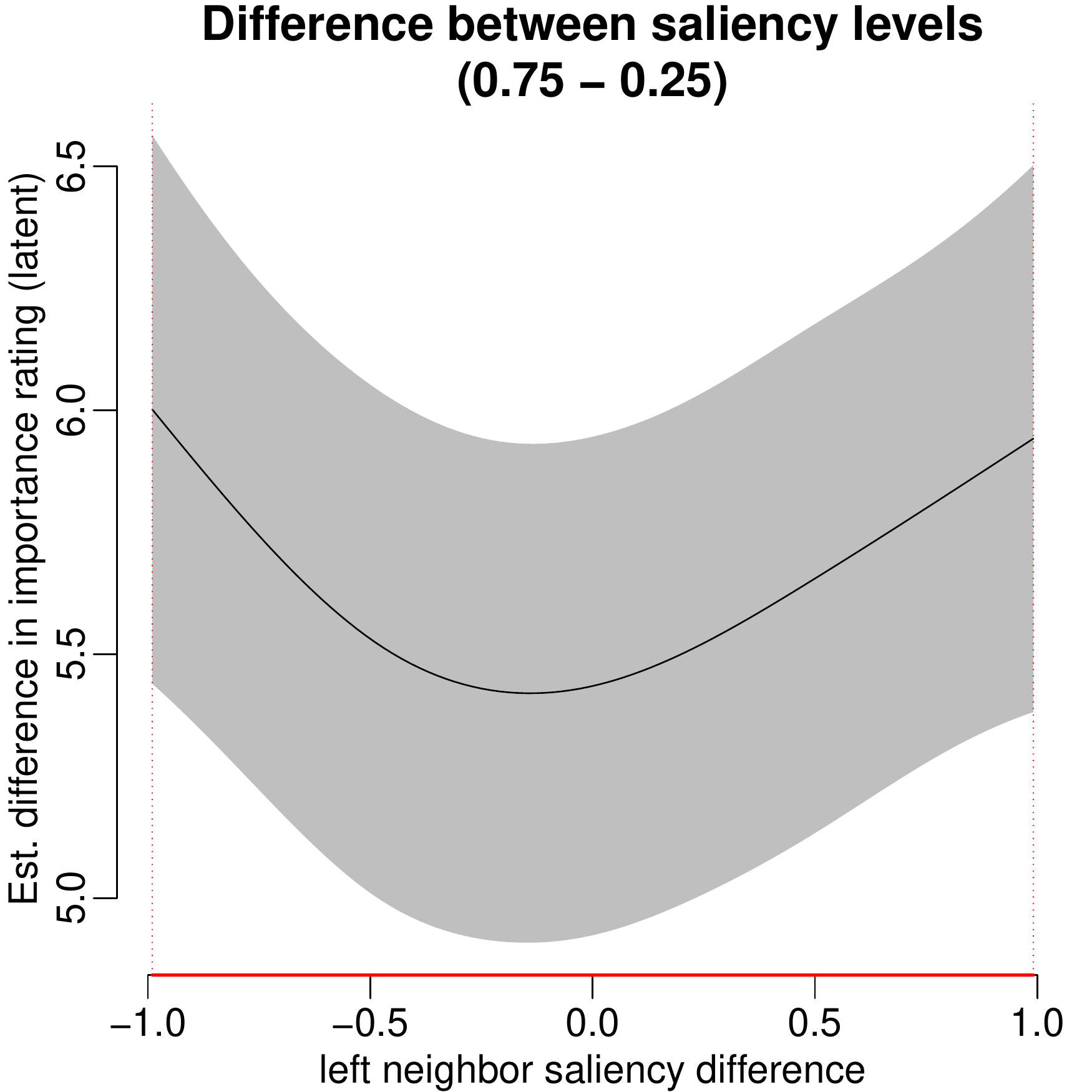}
    \caption{$t=50\%$}
    \end{subfigure}%
    \begin{subfigure}[t]{.25\textwidth}
        \centering
    \includegraphics[width=\textwidth]{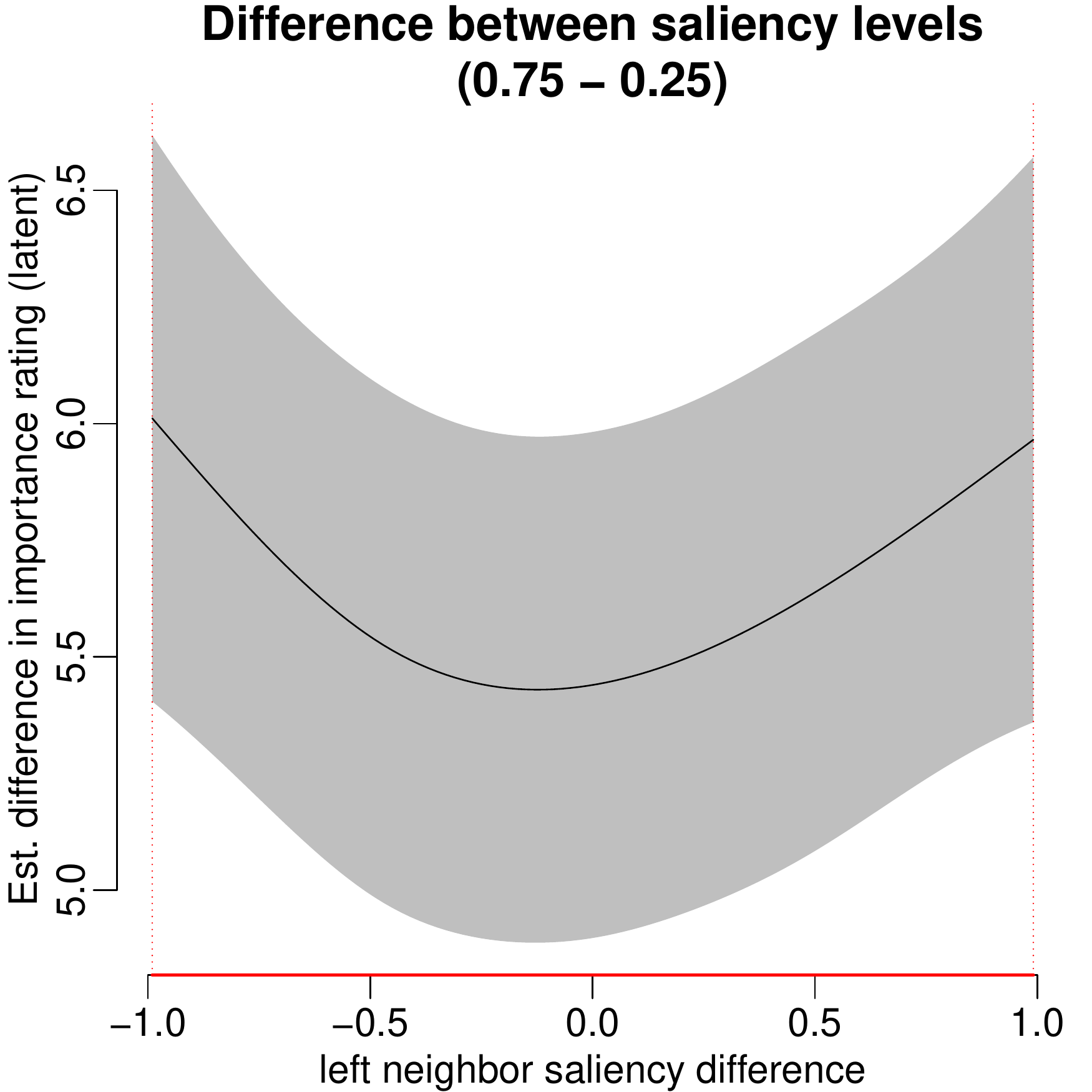}
    \caption{$t=25\%$}
    \end{subfigure}
    \caption{Difference plots between the influence of left saliency differences between exemplary high (0.75) and low (0.25) rated word saliency levels across different choices of thresholds for our randomized explanation experiment. We find similar patterns across all settings. The $\upvarphi^2$ measure is used across all plots.}\label{fig:eval_robustness_threshold_saliency_level_left}
\end{figure*}

\begin{figure*}[h!]
    \centering
    \begin{subfigure}[t]{.25\textwidth}
        \centering
    \includegraphics[width=\textwidth]{figures/phi_sq_875/diff_saliency_levels_left.pdf}
    \caption{$\upvarphi^2$.}
    \end{subfigure}%
    \hfill
    \begin{subfigure}[t]{.25\textwidth}
        \centering
    \includegraphics[width=\textwidth]{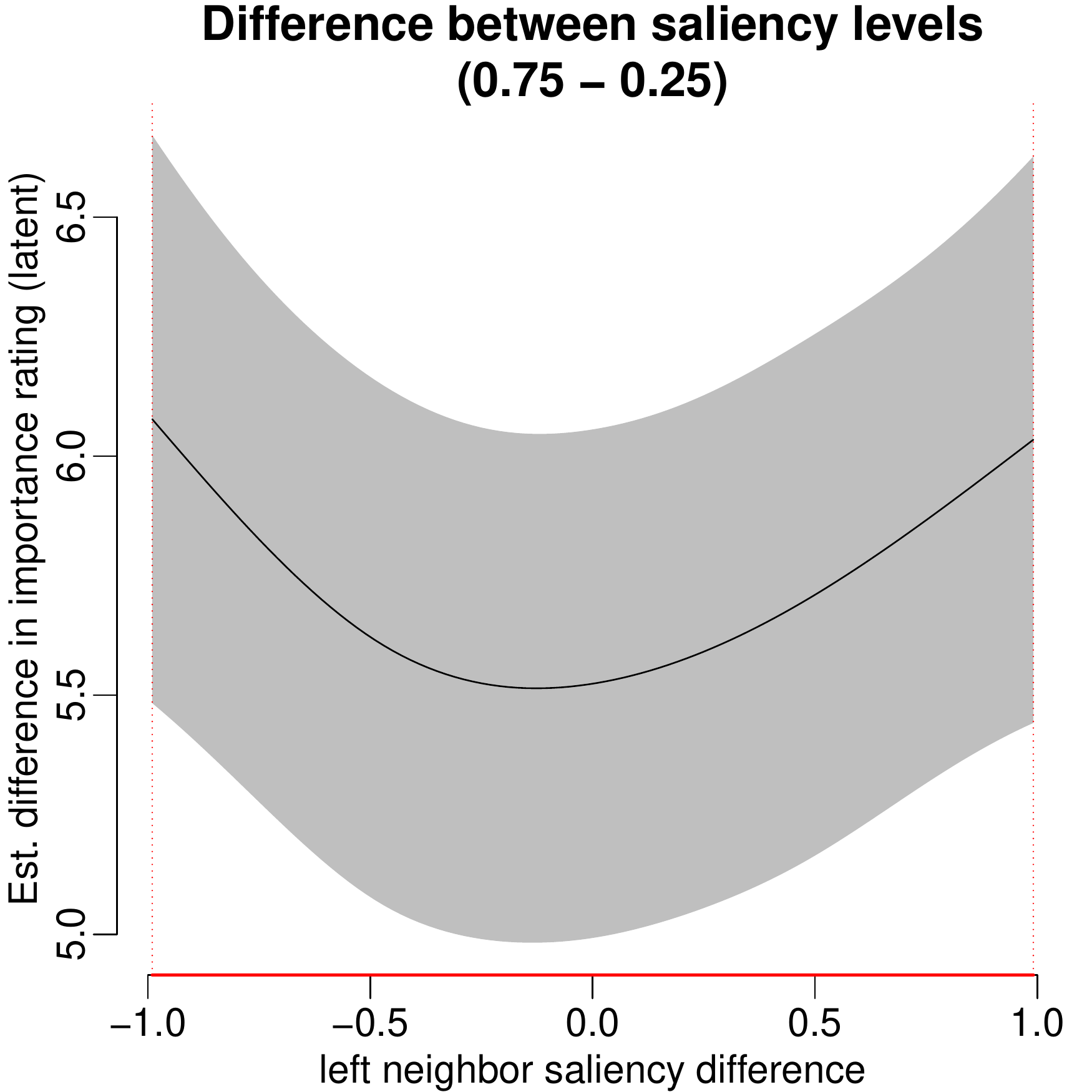}
    \caption{Jaccard.}
    \end{subfigure}%
    \begin{subfigure}[t]{.25\textwidth}
        \centering
    \includegraphics[width=\textwidth]{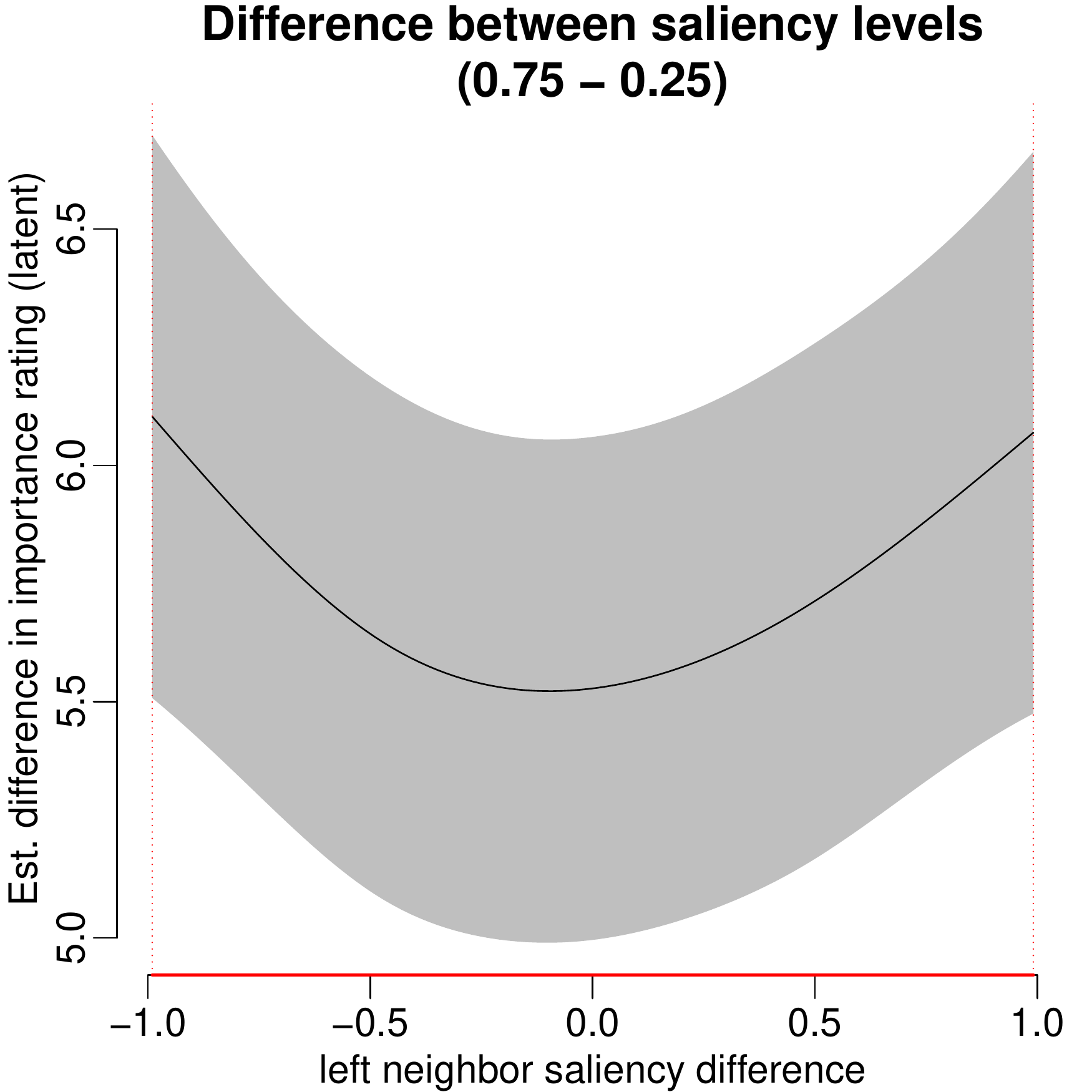}
    \caption{MI-like.}
    \end{subfigure}%
    \begin{subfigure}[t]{.25\textwidth}
        \centering
    \includegraphics[width=\textwidth]{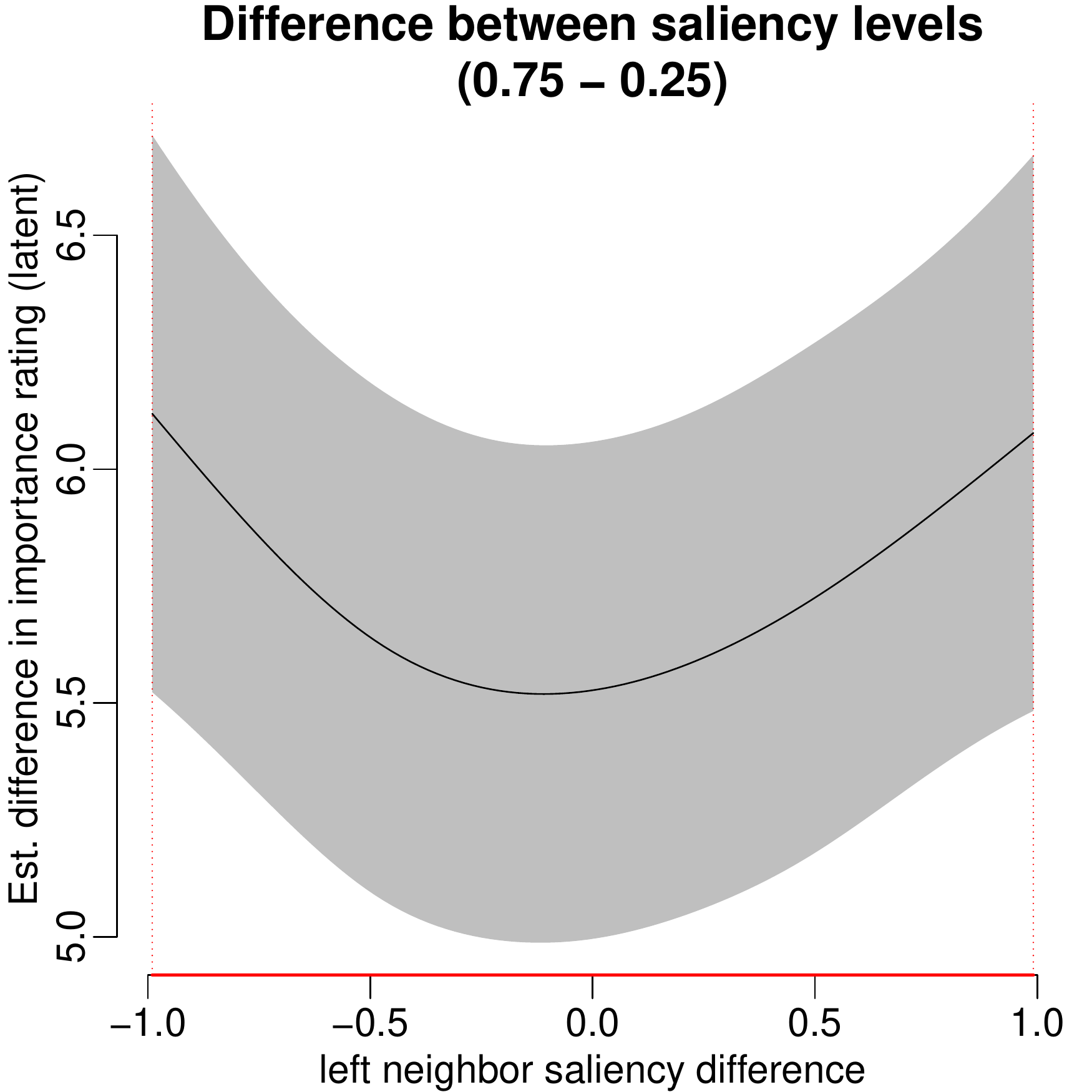}
    \caption{Poisson-Stirling.}
    \end{subfigure}
    \caption{Difference plots between the influence of left saliency differences between exemplary high (0.75) and low (0.25) rated word saliency levels across across different choices of cooccurrence measures for our randomized explanation experiment. We find similar patterns across all settings. $t=87.5$ is consistent for all plots.}\label{fig:eval_robustness_measure_saliency_level_left}
\end{figure*}

To ensure our results are not an artifact of the particular combination of threshold and cooccurrence measure, we investigate how our results change if we (i) vary the threshold within $\{0.5, 0.75, 0.875\}$ and (ii) vary the cooccurence measure within \{Jaccard, MI-like, $\upvarphi^2$, Poisson-Stirling\}.
We find significant interactions and observe similar interaction patterns as well as areas of significant differences (left/right, chunk/no chink as well as saliency levels) across all settings.
We provide a representative selection of plots in \Cref{fig:eval_robustness_measure_left_no_chunk,fig:eval_robustness_measure_left_chunk,fig:eval_robustness_measure_chunk_diff_p,fig:eval_robustness_threshold_no_chunk_diff_p,fig:eval_robustness_threshold_saliency_level_left,fig:eval_robustness_measure_saliency_level_left}.
Additionally, \Cref{tab:stats_full_phi_sq_25,tab:stats_full_mi_like_875} demonstrate that changing the threshold or cooccurrence measure leads to model statistics that are largely consistent with the results reported in \Cref{tab:stats_full}.
We choose the $\upvarphi^2$ and a 87.5\% threshold as no other model reaches a higher deviance explained and a comparison of randomly-sampled chunk/no chunk examples across measures and thresholds yields the best results for this setting.

\FloatBarrier

\begin{table*}[t]
\centering

\resizebox{0.65\linewidth}{!}{
\begin{tabular}{p{5cm}rrrr}
  \toprule
 \textbf{Term} & \textbf{(e)df} & \textbf{Ref.df} & \textbf{F} & $p$ \\ 
  \midrule
s(saliency) & 11.23 & 19.00 & 547.16 & \textbf{$<$ 0.0001} \\ 
  s(display\_index) & 3.10 & 9.00 & 20.93 & \textbf{$<$ 0.0001} \\ 
  s(word\_length) & 1.61 & 9.00 & 16.47 & \textbf{$<$ 0.0001} \\ 
  s(sentence\_length) & 0.00 & 4.00 & 0.00 & 0.436 \\ 
  s(relative\_word\_frequency) & 0.00 & 9.00 & 0.00 & 0.814 \\ 
  s(normalized\_saliency\_rank) & 0.58 & 9.00 & 0.36 & 0.120 \\ 
  s(word\_position) & 0.59 & 9.00 & 0.18 & 0.173 \\ 
  te(left diff.,saliency): no chunk & 2.90 & 24.00 & 1.21 & \textbf{0.003} \\ 
  te(left diff.,saliency): chunk & 3.34 & 24.00 & 0.92 & \textbf{0.015} \\ 
  te(right diff.,saliency): no chunk & 2.50 & 24.00 & 0.67 & \textbf{0.021} \\ 
  te(right diff.,saliency): chunk & 0.00 & 24.00 & 0.00 & 0.836 \\ 
  \midrule
  s(sentence\_id) & 0.00 & 149.00 & 0.00 & 0.601 \\ 
  s(saliency,sentence\_id) & 17.35 & 150.00 & 0.15 & 0.178 \\ 
  s(worker\_id) & 62.19 & 63.00 & 30421.05 & \textbf{$<$ 0.0001} \\ 
  s(saliency,worker\_id) & 62.11 & 64.00 & 17591.01 & \textbf{$<$ 0.0001}\\ 
  \midrule
  capitalization & 2.00 & & 3.01 & \textbf{0.049} \\ 
  dependency\_relation & 35.00&  & 2.93 & \textbf{$<$ 0.0001} \\ 
   \bottomrule
\end{tabular}}
\caption{(Effective) degrees of freedom, reference degrees of freedom and Wald test statistics for the univariate smooth terms (top), random effects terms (middle) and parametric fixed terms (bottom) using $t=25\%$ and $\upvarphi^2$ measure for our randomized explanation experiment.}
\label{tab:stats_full_phi_sq_25}
\end{table*}

\begin{table*}[ht]
\centering

\resizebox{0.65\linewidth}{!}{
\begin{tabular}{p{5cm}rrrr}
  \midrule
\textbf{Term} & \textbf{(e)df} & \textbf{Ref.df} & \textbf{F} & $p$ \\ 
  \midrule
s(saliency) & 11.21 & 19.00 & 584.57 & \textbf{$<$ 0.0001} \\ 
  s(display\_index) & 3.04 & 9.00 & 21.63 & \textbf{$<$ 0.0001} \\ 
  s(word\_length) & 1.63 & 9.00 & 16.66 & \textbf{$<$ 0.0001} \\ 
  s(sentence\_length) & 0.00 & 4.00 & 0.00 & 0.407 \\ 
  s(relative\_word\_frequency) & 0.00 & 9.00 & 0.00 & 0.813 \\ 
  s(normalized\_saliency\_rank) & 0.56 & 9.00 & 0.32 & 0.130 \\ 
  s(word\_position) & 0.65 & 9.00 & 0.22 & 0.159 \\ 
  te(left diff.,saliency): no chunk & 3.10 & 24.00 & 1.57 & \textbf{0.0010} \\ 
  te(left diff.,saliency): chunk & 1.79 & 24.00 & 0.34 & 0.082 \\ 
  te(right diff.,saliency): no chunk & 2.37 & 24.00 & 0.47 & \textbf{0.048} \\ 
  te(right diff.,saliency): chunk & 0.64 & 24.00 & 0.05 & 0.249 \\ 
  \midrule
  s(sentence ID) & 0.00 & 149.00 & 0.00 & 0.638 \\ 
  s(saliency,sentence ID) & 17.14 & 150.00 & 0.15 & 0.164 \\ 
  s(worker ID) & 62.19 & 63.00 & 30521.95 & \textbf{$<$ 0.0001} \\ 
  s(saliency,worker ID) & 62.11 & 64.00 & 16749.25 & \textbf{$<$ 0.0001}\\ 
  \midrule
  capitalization & 2.00 & & 3.23 & \textbf{0.039} \\ 
  dependency relation & 35.00 & & 2.94 & \textbf{$<$ 0.0001} \\ 
   \bottomrule
\end{tabular}}
\caption{(Effective) degrees of freedom, reference degrees of freedom and Wald test statistics for the univariate smooth terms (top), random effects terms (middle) and parametric fixed terms (bottom) using $t=87.5\%$ and MI-like measure for our randomized explanation experiment.}
\label{tab:stats_full_mi_like_875}
\end{table*}

\end{document}